\documentclass[pdflatex,sn-mathphys-ay]{sn-jnl}% Math and Physical Sciences Numbered Reference Style
\usepackage{graphicx}%
\usepackage{multirow}%
\usepackage{amsmath,amssymb,amsfonts}%
\usepackage{amsthm}%
\usepackage{mathrsfs}%
\usepackage[title]{appendix}%
\usepackage{xcolor}%
\usepackage{textcomp}%
\usepackage{manyfoot}%
\usepackage{booktabs}%
\usepackage{url}
\usepackage{algorithm}%
\usepackage{algorithmicx}%
\usepackage{algpseudocode}%
\usepackage{listings}%

\usepackage{array}%
\usepackage{tikz}
\usepackage[T1]{fontenc}
\usepackage{newtxtext,newtxmath}
\usepackage{microtype}
\usepackage{longtable}
\usepackage{needspace}
\usepackage{ragged2e}
\hypersetup{
  hidelinks,
  pdftitle={Adaptation of Foundation Models in Medical Image Analysis: Mechanisms, Evidence, and Deployment Considerations},
  pdfauthor={Karma Phuntsho, Abdullah, Kyungmi Lee, Ickjai Lee, Euijoon Ahn},
  pdfsubject={Preprint}
}
\newcolumntype{P}[1]{>{\RaggedRight\arraybackslash}p{#1}}
\newcolumntype{Q}[1]{>{\RaggedLeft\arraybackslash}p{#1}}

\theoremstyle{thmstyleone}%
\theoremstyle{thmstyletwo}%

\theoremstyle{thmstylethree}%

\begin{document}

\title[Article Title]{Adaptation of Foundation Models in Medical Image Analysis: Mechanisms, Evidence, and Deployment Considerations}


\author*[1]{\fnm{Karma} \sur{Phuntsho}}\email{karma.phuntsho@my.jcu.edu.au}

\author[1]{\fnm{} \sur{Abdullah}}\email{abdullah@my.jcu.edu.au}

\author[1]{\fnm{Kyungmi} \sur{Lee}}\email{joanne.lee@jcu.edu.au}

\author[1]{\fnm{Ickjai} \sur{Lee}}\email{ickjai.lee@jcu.edu.au}

\author[1]{\fnm{Euijoon} \sur{Ahn}}\email{euijoon.ahn@jcu.edu.au}

\affil*[1]{\orgdiv{College of Science and Engineering}, \orgname{James Cook University}, \orgaddress{\street{McGregor Road}, \city{Smithfield}, \state{QLD}, \postcode{4878}, \country{Australia}}}


\abstract{

Foundation models, pretrained on diverse data, objectives, or modalities for reuse across downstream tasks, are increasingly applied to medical image analysis. Their performance depends not only on pretraining but also on how they are adapted to specific tasks, data regimes, and deployment settings. Whereas previous reviews have considered adaptation within broader syntheses, this review focuses specifically on post-pretraining interventions applied to existing FMs. Using a structured narrative review with a systematic, auditable workflow, we identified 719 eligible reports representing 705 independent studies. Of these, 473 provided controlled comparative evidence: 322 formally published studies formed the primary synthesis and 151 preprint-only studies a sensitivity layer. Adaptation was organized into five non-mutually exclusive families: parameter-level, representation-level, objective-level, data-centric, and architectural/sequence-level adaptation, with model use and update regime analyzed separately. Controlled comparisons showed no consistent ranking among frozen, parameter-efficient, partial, or full updating, or across broader adaptation families; outcomes varied by task, data regime, comparator, and evaluation setting. Clinical-readiness evidence remained limited: 173/705 studies (24.5\%) had confirmed independent external validation, 10 (1.4\%) reported calibration or uncertainty evidence, and 97/322 formally published comparative studies (30.1\%) included a qualifying externally validated arm. Overall, FM adaptation is a context-dependent design space rather than a hierarchy of preferred strategies.
}




\maketitle

\section{Introduction}\label{sec1:intro}

Foundation models (FMs), broadly defined as models pretrained on sufficiently broad and diverse data, objectives, or modalities to support reuse across multiple downstream tasks or domains, have emerged as an important paradigm in medical
image analysis (MIA) \citep{ref10_fm,ref36_vfm_mia}. Their capacity for downstream reuse is particularly relevant in medical imaging, where expert annotation is costly and data distributions vary across modalities, institutions, scanners, acquisition protocols, and patient populations. Post-pretraining adaptation of existing FMs has been investigated across classification and diagnosis \citep{ref126_tfa-lt,ref136_melo,ref96_peft_mia}, segmentation \citep{ref31_medsam,ref53_medsa,ref127_masam,ref130_3dsam_adapter}, and detection or anomaly localization \citep{ref145_mvfa,ref146_mediclip}. In parallel, a distinct body of work has focused on developing or pretraining new medical FMs for segmentation, localization, multimodal reasoning, and report generation \citep{ref157_medsam2,ref119_sammed3d,ref121_vista3d,ref122_medlsam,ref123_BiomedParse,ref34_generalist_fm}. Although this broader literature establishes the expanding capability landscape of medical FMs, evidence from FM development or pretraining does not, by itself, establish the effects of post-pretraining adaptation.

Adaptation warrants separate analysis because medical imaging introduces constraints that pretraining alone may not resolve. Medical datasets are often smaller, imbalanced, modality-specific, and subject to variation across acquisition protocols and institutions \citep{ref34_generalist_fm,ref1_3d_ssl_methods_medical}. Broader medical-AI evidence also identifies distribution shift, calibration, rare findings, and task-specific errors as concerns that may remain obscured by internal benchmark performance \citep{ref12_challenges_per_fm,tejani2024claim,ref_decide_ai2022}. These considerations motivate empirical evaluation of adaptation, but they do not establish that any particular adaptation mechanism confers greater robustness or clinical reliability. An adapted FM may improve a development endpoint without demonstrating external transportability, calibration, subgroup performance, or clinical utility.

Existing surveys address FMs at healthcare-wide, radiology-specific, and medical-imaging-specific levels \citep{ref33_comprehensive_survey_fm,ref35_fm_advancing_healthcare,ref9_fm_medical_comprehensive_survey,jiao2025foundation,ryu2025vision,rajendran2025foundation,d2026foundation,paschali2025foundation}. Many provide substantial accounts of adaptation, robustness, evaluation, and deployment, and the relevant distinction is therefore not the mere presence or absence of these topics. Instead, the comparison below focuses on how adaptation is delimited as an analytical object, how evidence is organized and compared, and how failure, validation, and lifecycle evidence are incorporated. Table~\ref{tab:positioning} is consequently a descriptive audit of review architecture rather than a coverage score or assessment of quality, novelty, completeness, or overall value. The present review is deliberately not included as a comparison row because its positioning is stated separately after the source-level audit.

\begingroup
\fontsize{8.4}{10.5}\selectfont
\renewcommand{\arraystretch}{1.15}
\setlength{\tabcolsep}{2pt}
\setlength{\LTleft}{0pt}
\setlength{\LTright}{0pt}
\begin{longtable}{@{}P{2.30cm}P{2.70cm}P{2.75cm}P{2.55cm}P{2.55cm}P{3.05cm}@{}}
\caption{Scope and evidence architecture of selected recent surveys of foundation models (FMs) in healthcare and medical image analysis.}\label{tab:positioning}\\
\toprule

\textbf{Survey reference} &
\textbf{Adaptation evidence boundary} &
\textbf{Adaptation framework} &
\textbf{Comparative synthesis design} &
\textbf{Failure-analysis structure} &
\textbf{Validation and lifecycle evidence} \\
\midrule
\endfirsthead
\multicolumn{6}{@{}l}{\textit{Table \thetable\ (continued)}}\\[3pt]
\toprule

\textbf{Survey reference} &
\textbf{Adaptation evidence boundary} &
\textbf{Adaptation framework} &
\textbf{Comparative synthesis design} &
\textbf{Failure-analysis structure} &
\textbf{Validation and lifecycle evidence} \\
\midrule
\endhead
\endfoot
\endlastfoot

Khan et al. \citeyearpar{ref33_comprehensive_survey_fm} &
Conceptual distinction between pretraining and downstream reuse &
Model/application taxonomy; adaptation dispersed &
Narrative/tabular synthesis &
General robustness, reliability, and security challenges &
Narrative validation, verification, and adoption \\

He et al. \citeyearpar{ref35_fm_advancing_healthcare} &
Explicit pretraining-to-adaptation stage framework &
Fine-tuning, adapter tuning, and prompting &
Narrative/tabular synthesis &
Structured responsibility, reliability, capability, and adaptability &
Narrative real-world, human--AI, trust, and continual updating \\

\addlinespace[2pt]

D'Antonoli et al. \citeyearpar{d2026foundation} &
Conceptual pretraining/adaptation distinction &
Probing, partial/full fine-tuning, and parameter-efficient update pathways &
Narrative/application synthesis &
Structured general FM-risk analysis &
Structured validation, monitoring, regulation, and governance \\

Paschali et al. \citeyearpar{paschali2025foundation} &
Explicit pretraining-to-task-adaptation stages &
Zero/few-shot, probing, fine-tuning, instruction, and alignment pathways &
Narrative synthesis &
Structured FM failure and drift analysis &
Structured clinical evaluation, monitoring, regulation, and continual learning \\

\addlinespace[2pt]

Jiao et al. \citeyearpar{jiao2025foundation} &
Conceptual distinction between pretraining and downstream modification &
Data/model/training-strategy taxonomy &
Narrative/tabular synthesis &
Negative transfer and interference within model limitations &
Narrative evaluation, deployment, and regulation \\

Ryu et al. \citeyearpar{ryu2025vision} &
Post-pretraining evidence boundary not substantively developed &
Taxonomy by modality, architecture, and task &
Descriptive model/task/metric tables &
Bias, variance, and evaluation limitations &
No substantive validation-maturity or lifecycle synthesis identified \\

Rajendran et al. \citeyearpar{rajendran2025foundation} &
Pretraining and downstream adaptation treated in separate sections &
Vision FM/Vision \& Language FM, model/task/strategy categories &
Systematic quantitative cross-study synthesis &
Structured robustness evaluation &
Structured evaluation, fairness, governance, and monitoring \\

Azad et al. \citeyearpar{ref9_fm_medical_comprehensive_survey} &
Conceptual distinction between pretraining and downstream adaptation &
Taxonomy by prompt, model, and training strategy &
Narrative/tabular synthesis &
Structured bias, adversarial-risk, and adaptation challenges &
Narrative clinical applicability and benchmarking \\

\botrule

\end{longtable}

\vspace{0.4em}

\footnotesize

\textit{Note.} Entries summarize the reported organization and evidence-synthesis design of each survey after examination of its full text, section structure, tables, and figures. ``Adaptation evidence boundary'' describes whether FM development or pretraining is distinguished analytically from subsequent task adaptation; it does not imply that this distinction was used as a study-eligibility rule. ``Comparative synthesis design'' describes the unit and structure of comparison and does not imply that one synthesis design is intrinsically superior to another. ``No substantive \ldots{} identified'' denotes a bounded audit finding rather than a claim of absolute absence. The table does not assess survey quality, rigor, importance, completeness, novelty, or superiority.

\endgroup

The descriptors reflect different review purposes rather than a quality ordering. D'Antonoli et al. organize radiology FM adaptation around probing, fine-tuning, and parameter-efficient updating and provide sustained treatment of risk, validation, monitoring, and governance \citep{d2026foundation}. Khan et al. survey healthcare FM applications and challenges, with downstream adaptation and validation distributed across model- and application-oriented sections \citep{ref33_comprehensive_survey_fm}. He et al. explicitly distinguish pretraining from fine-tuning, adapter tuning, and prompt engineering and structure general FM limitations around responsibility, reliability, capability, and adaptability \citep{ref35_fm_advancing_healthcare}. Jiao et al. organize medical FMs by data, model family, training strategy, and application, with evaluation, failure, and regulatory themes treated across the broader survey rather than as dedicated post-pretraining evidence frameworks \citep{jiao2025foundation}. Ryu et al. provide a modality-, architecture-, and task-oriented synthesis of medical VLMs, including bias and metric limitations, without a substantive post-pretraining evidence boundary or validation-maturity framework identified in the audited report \citep{ryu2025vision}. Rajendran et al. provide systematic quantitative synthesis across tasks and applications together with structured robustness, fairness, evaluation, and lifecycle discussions. Their quantitative synthesis operates across heterogeneous studies rather than through a prespecified within-base attribution rule \citep{rajendran2025foundation}. Azad et al. organize prompted and visually pretrained FMs and give structured attention to bias, adversarial risk, and downstream-adaptation challenges \citep{ref9_fm_medical_comprehensive_survey}. Paschali et al. explicitly distinguish pretraining from task adaptation and provide structured clinical-evaluation, failure, drift, monitoring, and regulatory analyses \citep{paschali2025foundation}.

Accordingly, the distinction of the review lies in the \emph{integration of analytical operations} rather than exclusive coverage of any individual topic. Specifically, it applies an arm-level boundary for direct post-pretraining adaptation; separates model-side locus, learning-control modifier, and model use/update regime; introduces a comparative layer with source-level traceability restricted to attributable within-base contrasts; interprets failure evidence according to claim-specific provenance; and characterizes validation and reporting maturity at study level. This structure distinguishes three questions that are often conflated: whether an adapted system performs well, what can be attributed to the adaptation itself, and what evidence is available to support broader claims of robustness or clinical readiness.

\begin{figure}[t]
    \centering
    \includegraphics[
        width=\linewidth,
        height=0.7\textheight,
        keepaspectratio
    ]{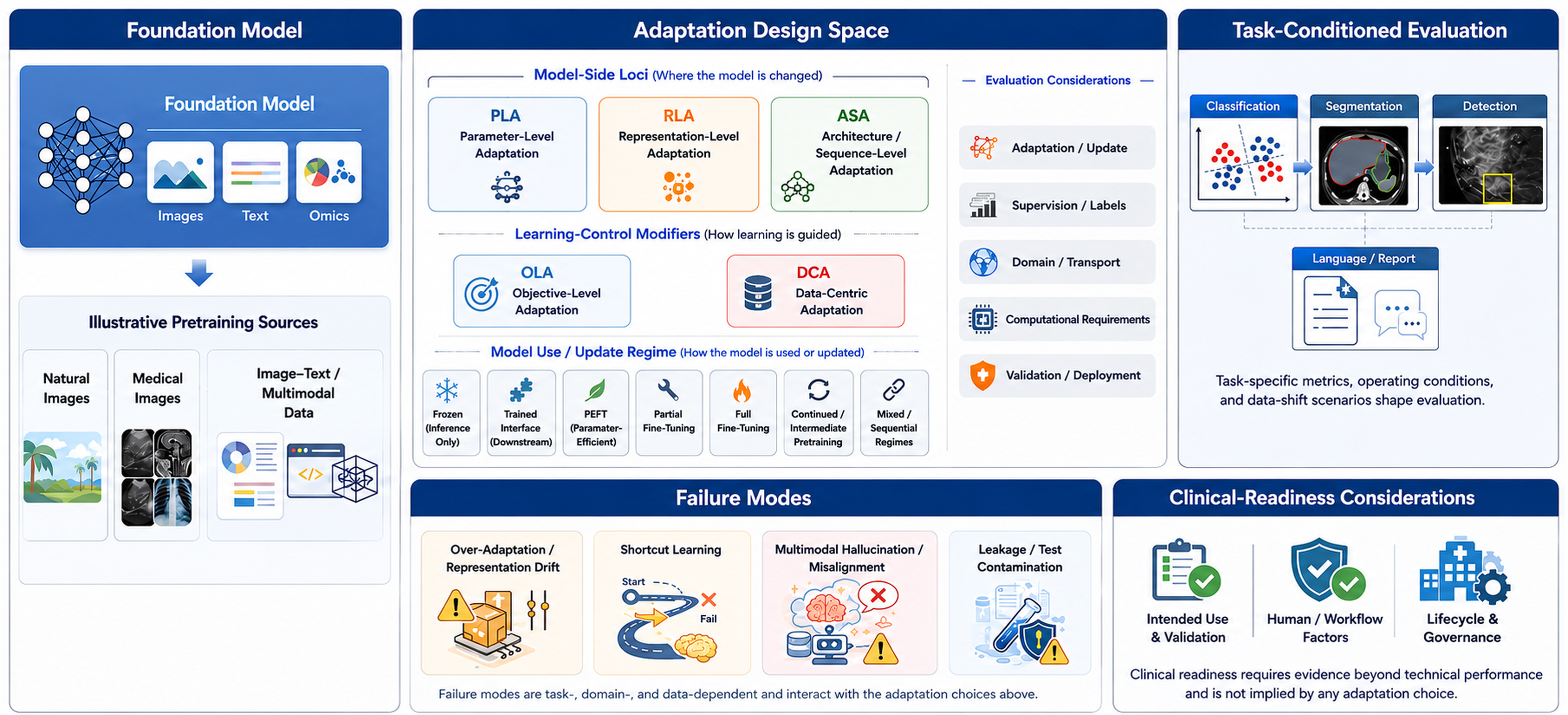}
    \caption{
    Overview of the mechanism-based framework for foundation-model adaptation in medical image analysis.
    Post-pretraining adaptation is characterized using five non-mutually exclusive mechanism families:
    parameter-level adaptation (PLA), representation-level adaptation (RLA), objective-level adaptation (OLA),
    data-centric adaptation (DCA), and architectural/sequence-level adaptation (ASA).
    These mechanisms may occur individually or in combination and are considered separately from the model
    use or update regime.
    }
    \label{fig:overview}
\end{figure}

This review therefore treats adaptation as an identifiable post-pretraining intervention rather than as synonymous with fine-tuning. The framework separates adaptation into two intervention dimensions. Model-side mechanisms comprise parameter-level, representation-level, and architectural/sequence-level adaptation, whereas learning-control mechanisms comprise objective-level and data-centric adaptation. Model use and update regime are coded separately, distinguishing frozen use, downstream interfaces, parameter-efficient updating, partial updating, full updating, continued pretraining, and mixed or sequential pathways. These mechanism families are non-mutually exclusive and descriptive. They do not represent stages in a temporal sequence, an ordinal scale of adaptation depth, or evidence of robustness, safety, or clinical readiness. Figure~\ref{fig:overview} summarizes the framework and the relationships among these dimensions.

Performance outcomes and evidentiary role are treated separately from mechanism labels. The review first maps the contemporary literature descriptively, then applies a nested comparative criterion to studies containing attributable within-study contrasts based on a shared pretrained model or explicit within-base component ablations. This distinction separates evidence that a complete adapted system performs well from evidence that supports attribution of an observed difference to a particular update regime, component, or adaptation intervention. Because tasks, datasets, comparators, and outcome measures are heterogeneous, results are not pooled into a common effect estimate, and no comparative ranking of adaptation families is inferred.

Adaptation is also examined in relation to task-specific failure and evaluation characteristics. Classification raises questions of long-tailed performance, calibration, and clinically relevant operating thresholds. Segmentation introduces boundary, surface, volumetric, and interaction-dependent errors, whereas detection or localization emphasizes rare-finding sensitivity, localization validity, and false-positive burden \citep{ref126_tfa-lt,ref47_metrics,muller2022towards,ref146_mediclip}. These characteristics are used as task-conditioned evaluation dimensions rather than as assumptions that any single failure mode is universally dominant or clinically decisive. Technical robustness is considered separately from clinical readiness, which additionally depends on factors such as intended use, external and prospective evaluation, clinical utility, human factors, workflow integration, equity, cybersecurity, and lifecycle monitoring.

The review makes three main contributions. First, it provides a mechanism-centered taxonomy that distinguishes model-side interventions, learning-control mechanisms, and update regimes, allowing overlapping adaptation pathways to be characterized without imposing a single hierarchy. Second, it maps adaptation across imaging modalities and major medical-imaging tasks and synthesizes task- and mechanism-relevant failure patterns, including negative transfer, shortcut reliance, data leakage, and multimodal semantic failure. Third, it distinguishes descriptive evidence mapping from controlled comparative synthesis and examines external validation, calibration, reporting, and clinical-readiness evidence at the level supported by the underlying studies. Together, these contributions provide an evidence map and interpretive framework for evaluating adaptation claims rather than a prescription for selecting a universally preferred adaptation strategy.

% ============================================================
% SECTION 2
% ============================================================

\section{Scope, Definitions, and Review Methods}
\label{sec:scope}

The rapid adoption of FMs in MIA has been accompanied by broad and sometimes inconsistent use of the term \emph{adaptation}. To support consistent study selection and comparison, we classified adaptation mechanisms along two complementary dimensions: the \emph{model-side locus} at which an intervention acts and the \emph{learning-control modifier} through which the learning process is altered. The \emph{update regime}, describing whether and how the pretrained model is used or updated during downstream adaptation, was coded separately as a distinct property of the adaptation pathway. These dimensions were treated as descriptive rather than ordinal as neither the extent of parameter updating nor membership in a particular mechanism family was assumed to imply greater robustness, clinical utility, safety, or deployment readiness.

\subsection{Review Design and Search Strategy}
\label{subsec:review_methodology}

We conducted a structured narrative review using a systematic and auditable workflow for evidence identification, study selection, data extraction, characterization, and synthesis. Searches were conducted on 1 August 2026 in PubMed, Scopus, Web of Science, IEEE Xplore, and arXiv for reports published or posted between 1 January 2020 and 1 August 2026. The search strategy combined three concept blocks covering foundation models, medical or biomedical imaging, and downstream use or adaptation. No document-type restriction was applied. English-language eligibility was implemented using native database filters where available and during screening otherwise.

Exact database-specific search strings, search fields, filters, search yields, and detailed record-flow information are provided in the Supplementary Reproducibility Appendix. Additional records identified through citation tracing and proceedings reconciliation were evaluated using the same eligibility criteria and did not extend the 1 August 2026 eligibility cutoff. Amendment, study-linkage, and correction provenance is documented in the accompanying audit materials.

\subsection{Eligibility, FM-Adaptation Definition, and Coding Framework}
\label{subsec:definition_scope}

For this review, a FM was defined as a model pretrained on sufficiently broad and diverse data, objectives, or modalities to support reuse across multiple downstream tasks or domains through prompting, probing, or adaptation. Model scale alone was insufficient for eligibility. Accordingly, ordinary use of an ImageNet-pretrained convolutional network, a task-specific transformer, or a model trained from scratch for a single downstream task did not automatically qualify as FM use. Both general-domain and medically pretrained FMs were eligible when a qualifying post-pretraining intervention was evaluated. Representative examples included RETFound \citep{zhou2023foundation}, PanDerm \citep{yan2025multimodal}, UNI \citep{chen2024towards}, Prov-GigaPath \citep{xu2024provgigapath}, MONET \citep{kim2024monet}, LLaVA-Med \citep{li2023llava}, and Med-PaLM M \citep{tu2024towards}.

We defined \emph{adaptation} as an identifiable post-pretraining intervention applied to a pretrained FM, its representation, supervision, optimization process, or downstream computational pathway to align it with a medical-imaging domain, task, or clinical context without retraining the base model from scratch. An evaluated arm contributed direct FM-adaptation evidence only when it began from an eligible pretrained FM, applied an attributable post-pretraining intervention, and retained the eligible FM, a component carrying its pretrained FM weights, or an FM-derived representation of the evaluated input within the inference pathway. FM used only during training as a teacher, pseudo-labeler, or synthetic-data generator did not qualify. A distilled student qualified only if it independently met the FM definition before the evaluated intervention or retained an eligible pretrained FM component. Eligibility was assessed at the arm level and did not imply that the intervention was beneficial or that the supporting evidence was comparatively strong.

Qualifying core tasks were classification or diagnosis, segmentation, and detection or localization. Studies entering the review through at least one eligible core-task arm could additionally contribute evidence on retrieval, image--text matching, report generation, visual question answering, explanation, medical-image or image-to-image generation, or image-grounded prognostic and biomarker tasks. These secondary tasks did not independently establish corpus eligibility. Large-scale pretraining alone, training from scratch, routine zero-shot use, and ordinary fixed-feature probing or task-head fitting did not independently confer direct FM-adaptation status. Conventional comparators reported within an eligible study likewise remained contextual unless they separately satisfied the adaptation criteria.

Adaptation mechanisms were coded using five non-mutually exclusive families. Three described the \emph{model-side locus} of intervention. Parameter-Level Adaptation (PLA) captured modification of pretrained or added trainable parameters; Representation-Level Adaptation (RLA) captured post-pretraining transformation or realignment of an existing FM representation; and Architectural/Sequence-Level Adaptation (ASA) captured bounded structural, fusion, projection, volumetric, or sequence-processing changes around a retained FM. Two additional families described \emph{learning-control modifiers}. Objective-Level Adaptation (OLA) captured changes to the post-pretraining optimization objective, whereas Data-Centric Adaptation (DCA) captured changes to the effective learning signal through sampling, augmentation, synthetic data, weak supervision, retrieval examples, or related controls. Mechanism families were coded using a multi-label scheme and were not treated as mutually exclusive or ordinal.

Model use and update regime were coded separately as U0, inference-only frozen use; U1, frozen FM with a trained downstream interface; U2, parameter-efficient fine-tuning, including prompt-based and other restricted-parameter approaches; U3, partial fine-tuning of pretrained parameters; U4, full fine-tuning; U5, continued or intermediate pretraining; and U6, mixed or sequential regimes. U0 and U1 could occur as comparators or components within eligible studies but did not, by themselves, establish direct FM-adaptation eligibility. Mechanism family and update regime were therefore treated as separate dimensions, describing \emph{what} aspect of the adaptation pathway changed and \emph{how} the pretrained FM was used or updated, respectively. Full coding definitions, boundaries, and examples are provided in the Supplementary Reproducibility Appendix.

\subsection{Screening, Data Extraction, and Evidence Appraisal}
\label{subsec:screening_linkage}

Three reviewers participated in study screening, data extraction, and adaptation categorization using common eligibility criteria and a structured extraction framework. The study corpus was distributed across reviewers rather than undergoing complete duplicate screening or extraction by every reviewer. Cases that were ambiguous, internally inconsistent, or flagged during verification were discussed among the three reviewers and adjudicated by consensus. Where the available source evidence remained insufficient for a definitive classification, the final status was recorded as unresolved. Retrieval failure was recorded separately from scientific exclusion.
Multiple reports describing the same underlying study were linked using model identity, cohorts, interventions, outcomes, authorship, identifiers, and companion-publication information. Ambiguous report-to-study links were reviewed among the three reviewers and resolved by consensus before finalization of the study-level corpus.

Study-level extraction established the 705-study descriptive evidence map. Extracted variables included FM identity and pretraining source, adaptation intervention, model use and update regime, mechanism family, trainable components, imaging modality, task, datasets, label regime when reported, validation design, comparator, external-validation status, robustness, calibration or uncertainty, efficiency or resource reporting, and source-located quantitative findings. Information not reported in the source was recorded as not reported rather than inferred or imputed. For all 473 studies entering the comparative layer, comparator definitions, adaptation attribution, validation design, and retained quantitative findings were additionally checked against the source reports before the layer was stratified by publication status into the 322-study primary synthesis and 151-study preprint sensitivity layer. Detailed extraction, harmonization, study-linkage, and finding-selection rules are provided in the Supplementary Reproducibility Appendix.

A structured, design-aware appraisal framework was used to accommodate the heterogeneous study designs represented in the corpus. The appraisal considered cohort selection, data independence and potential leakage, validation design, comparator adequacy, outcome analysis, uncertainty reporting, and reporting completeness. These domains were used to characterize methodological limitations and applicability rather than to assign a single numerical quality score. The relevance of evidence to the synthesis was considered separately according to directness to the adaptation question, validation depth, comparator informativeness, reporting completeness, and clinical relevance. Effect magnitude itself did not determine appraisal judgments. The appraisal definitions and decision rules are provided in the Supplementary Reproducibility Appendix.

\subsection{Synthesis}
\label{subsec:synthesis_methods}

Corpus characteristics were summarized descriptively at study level. Model use and update regimes, mechanism families, imaging modalities, and task groups were coded as non-mutually exclusive categories and were therefore not expected to sum to 100\%. Given substantial heterogeneity in models, adaptation strategies, datasets, outcome measures, and validation designs, quantitative effects were not statistically pooled, and no overall comparative ranking of adaptation families or update regimes was inferred. 
Publication status was treated separately from scientific eligibility. Studies with a formal journal, conference, or book publication were classified as formally published, whereas studies available only as preprints at the search cutoff were classified as preprint-only. 

To distinguish the broad descriptive evidence map from evidence more directly informative about adaptation effects, a nested comparative criterion was applied to the 705-study corpus. Comparative eligibility required an attributable within-study contrast based on the same pretrained base and starting weights, or an explicit within-base component ablation, together with matched, source-located quantitative results. This criterion identified 473 studies with controlled comparative evidence, comprising 322 formally published studies included in the primary comparative synthesis and 151 preprint-only studies retained as a publication-status sensitivity layer. Studies for which comparative eligibility remained unresolved were retained in the descriptive evidence map but excluded from the comparative synthesis.

This nested structure separated evidence describing the broader FM-adaptation landscape from within-study evidence more directly informative about differences attributable to an update regime, component, or adaptation intervention. It did not alter scientific eligibility for the broader review. Complete comparative eligibility criteria, decision rules, and arm-level evidence ledgers are provided in the Supplementary Reproducibility Appendix.

% ============================================================
% SECTION 3
% ============================================================

\section{Evidence-Base Results}
\label{sec:evidence_results}

\subsection{Study Selection and Corpus Profile}
\label{subsec:study_selection}

Searches across the five sources identified 4,975 records, including 482 from PubMed, 1,754 from Scopus, 1,097 from Web of Science, 567 from IEEE Xplore, and 1,075 from arXiv. After removal of 2,264 duplicate records and 47 proceedings-container records, 91 supplementary candidate records identified through documented citation tracing and proceedings or record reconciliation were added, yielding 2,755 unique records for screening.

\begin{figure}[!htbp]
\centering
\includegraphics[
width=0.95\textwidth,
height=0.90\textheight,
keepaspectratio
]{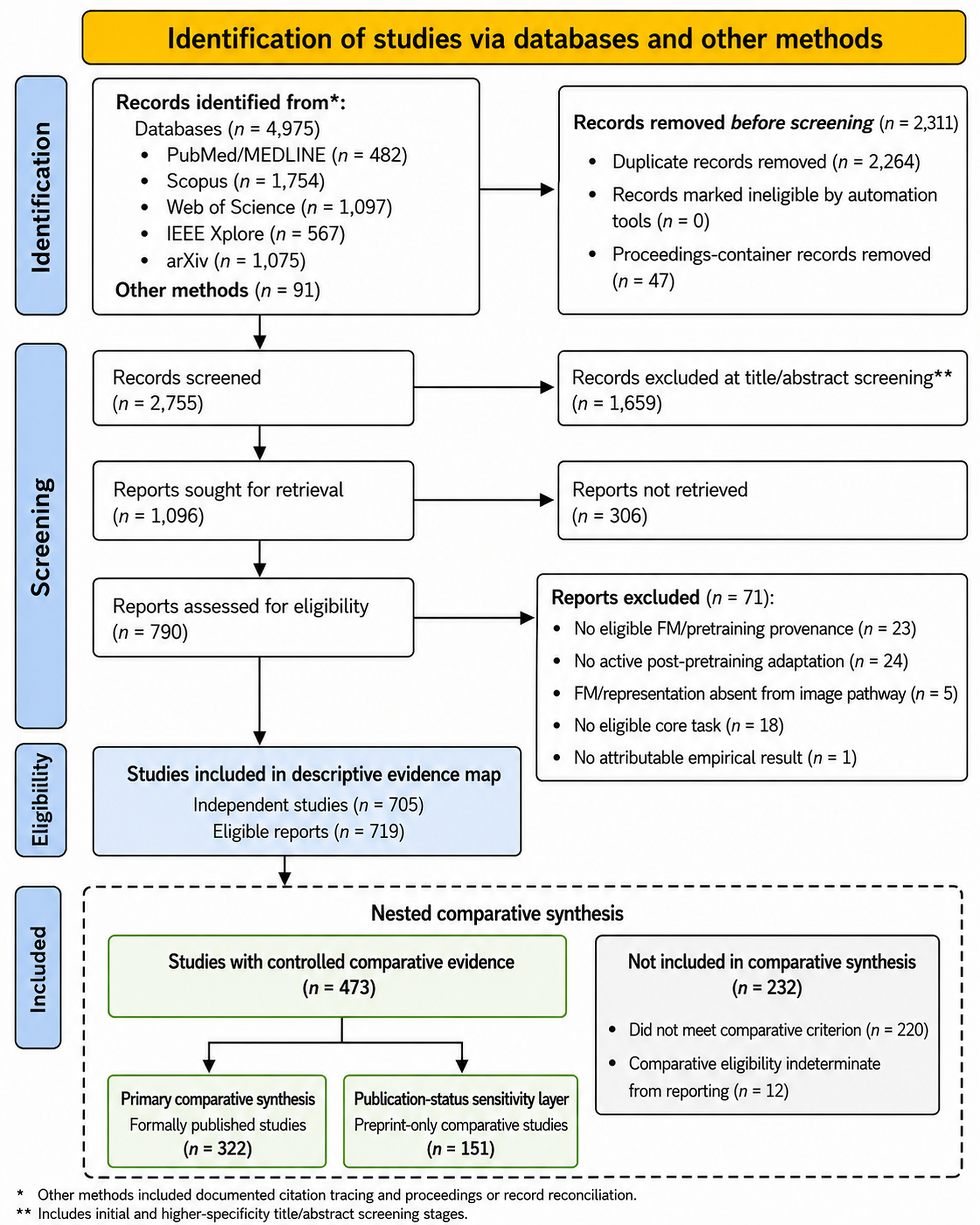}
\caption{Flow of evidence identification, study selection, and nested comparative synthesis. The descriptive evidence map comprised 705 studies from 719 eligible reports. Of these, 473 studies met the comparative criterion. The primary comparative synthesis comprised 322 formally published studies, while 151 preprint-only comparative studies were retained for publication-status sensitivity analysis. Flow diagram adapted from the PRISMA 2020 flow-diagram template.}
\label{fig:prisma_flow}
\end{figure}

Initial title-and-abstract screening excluded 787 records, leaving 1,968 for higher-specificity assessment. A further 872 records were excluded, and 1,096 reports were sought for retrieval. Full text could not be obtained for 306 reports through the documented retrieval procedures, leaving 790 reports for full-text assessment. Of these, 71 were excluded on scientific eligibility criteria. The final evidence base comprised 719 eligible reports representing 705 independent studies after report-to-study linkage (Figure~\ref{fig:prisma_flow}). Retrieval failure was recorded separately from scientific full-text exclusion.

The evidence base was concentrated in recent years. No included study was published or posted in 2020, 2021, or 2022. Twenty-six studies were dated 2023 (3.7\%), 130 were dated 2024 (18.4\%), 261 were dated 2025 (37.0\%), and 288 were dated 2026 up to the 1 August search cutoff (40.9\%). The 2026 count therefore represents a partial year.
Classification or diagnosis was evaluated in 356 studies (50.5\%), segmentation in 330 (46.8\%), and detection or localization in 70 (9.9\%). Secondary task categories included prognosis or biomarker prediction in 22 studies (3.1\%), report generation, visual question answering, reasoning, or explanation in 13 (1.8\%), and retrieval or image--text matching in 4 (0.6\%). Task groups were multi-label, and secondary categories were counted only within studies entering the corpus through at least one qualifying core-task arm. All 13 studies in the report-generation, visual-question-answering, reasoning, or explanation category had explicit image-grounded outputs; no study met the protocol category for a medical-image or image-to-image output. These distinctions are documented in Supplementary Data S17.

At the search cutoff, 485 studies (68.8\%) had at least one formal journal, conference, or book publication, whereas 220 (31.2\%) were available only as preprints. Imaging-modality distributions and adaptation-pattern frequencies are reported in Section~\ref{subsec:modality_adaptation}.

\subsection{Imaging Modalities and Adaptation Landscape}
\label{subsec:modality_adaptation}

The evidence base spanned a broad range of imaging modalities, although representation was uneven. Histopathology or whole-slide imaging was the largest group with 157 studies (22.3\%), followed by CT with 149 (21.1\%), MRI with 141 (20.0\%), and radiography or mammography with 125 (17.7\%). Other or multimodal imaging accounted for 92 studies (13.0\%), ophthalmic imaging for 72 (10.2\%), ultrasound for 64 (9.1\%), dermatology or clinical photography for 55 (7.8\%), endoscopy or surgical video for 48 (6.8\%), microscopy or cytology for 23 (3.3\%), and nuclear medicine for 4 (0.6\%). Modality assignments were multi-label.

Input dimensionality and pretraining provenance are retained at study level in Supplementary Data S1 because the source reports did not support reliable corpus-wide harmonization.
Table~\ref{tab:modality_mechanism_counts} summarizes selected update regimes, mechanism families, and confirmed independent external validation by modality. Across the full corpus, mixed or sequential regimes (U6) were coded in 251 studies (35.6\%), parameter-efficient fine-tuning (U2) in 219 (31.1\%), and full fine-tuning (U4) in 118 (16.7\%). Frozen-FM downstream interfaces (U1) occurred in 61 studies (8.7\%), partial fine-tuning (U3) in 35 (5.0\%), inference-only frozen use (U0) in 32 (4.5\%), and continued or intermediate pretraining (U5) in 16 (2.3\%). Because studies could evaluate more than one update regime, these frequencies describe observed use and are not mutually exclusive.

At the mechanism-family level, RLA was identified in 599 studies (85.0\%), OLA in 426 (60.4\%), ASA in 374 (53.0\%), PLA in 247 (35.0\%), and DCA in 257 (36.5\%). Mechanism-family assignments were also multi-label. Among modality groups containing at least 20 studies, confirmed independent external validation ranged from 17.4\% in microscopy or cytology to 31.3\% in endoscopy or surgical video. These distributions show substantial variation in the adaptation mechanisms and update regimes evaluated across imaging modalities. However, the underlying studies differed in models, tasks, datasets, comparators, and validation designs, and the evidence did not support ranking adaptation strategies by modality. Complete modality-by-update-regime and modality-by-mechanism matrices are provided in Supplementary Data S15--S16.

\begingroup
\fontsize{7.6}{9.2}\selectfont
\renewcommand{\arraystretch}{1.15}
\setlength{\tabcolsep}{1pt}
\setlength{\LTleft}{0pt}
\setlength{\LTright}{0pt}
\Needspace{.5\textheight}
\begin{longtable}{@{}P{2.70cm}Q{0.55cm}Q{1.55cm}Q{1.55cm}Q{1.55cm}Q{1.55cm}Q{1.55cm}Q{1.55cm}Q{1.55cm}Q{1.55cm}@{}}
\caption{Observed distribution of selected model use/update regimes, mechanism families, and confirmed independent external validation by imaging modality. Cells report study count and percentage within modality. Frequencies describe the evaluated evidence base and do not imply comparative effectiveness.}\label{tab:modality_mechanism_counts}\\
\toprule
\textbf{Modality group} &
\textbf{n} &
\textbf{U1} &
\textbf{U2} &
\textbf{U4} &
\textbf{U6} &
\textbf{RLA} &
\textbf{OLA} &
\textbf{ASA} &
\textbf{External} \\
\midrule
\endfirsthead
\multicolumn{10}{@{}l}{\textit{Table \thetable\ (continued)}}\\[3pt]
\toprule
\textbf{Modality group} &
\textbf{n} &
\textbf{U1} &
\textbf{U2} &
\textbf{U4} &
\textbf{U6} &
\textbf{RLA} &
\textbf{OLA} &
\textbf{ASA} &
\textbf{External} \\
\midrule
\endhead
\endfoot
\endlastfoot

Histopathology / whole-slide imaging &
157 & 26 (16.6) & 48 (30.6) & 29 (18.5) & 44 (28.0) &
130 (82.8) & 109 (69.4) & 71 (45.2) & 40 (25.5) \\

CT &
149 & 13 (8.7) & 42 (28.2) & 26 (17.4) & 53 (35.6) &
130 (87.2) & 95 (63.8) & 90 (60.4) & 27 (18.1) \\

MRI &
141 & 15 (10.6) & 42 (29.8) & 23 (16.3) & 46 (32.6) &
117 (83.0) & 94 (66.7) & 82 (58.2) & 33 (23.4) \\

Radiography / mammography &
125 & 4 (3.2) & 31 (24.8) & 27 (21.6) & 47 (37.6) &
101 (80.8) & 73 (58.4) & 51 (40.8) & 31 (24.8) \\

Other / multimodal &
92 & 2 (2.2) & 31 (33.7) & 16 (17.4) & 37 (40.2) &
71 (77.2) & 50 (54.3) & 49 (53.3) & 21 (22.8) \\

Ophthalmic imaging &
72 & 10 (13.9) & 25 (34.7) & 13 (18.1) & 18 (25.0) &
63 (87.5) & 50 (69.4) & 32 (44.4) & 18 (25.0) \\

Ultrasound &
64 & 4 (6.3) & 22 (34.4) & 10 (15.6) & 20 (31.3) &
51 (79.7) & 34 (53.1) & 40 (62.5) & 17 (26.6) \\

Dermatology / clinical photography &
55 & 3 (5.5) & 24 (43.6) & 4 (7.3) & 20 (36.4) &
43 (78.2) & 29 (52.7) & 31 (56.4) & 10 (18.2) \\

Endoscopy / surgical video &
48 & 0 (0.0) & 21 (43.8) & 6 (12.5) & 16 (33.3) &
42 (87.5) & 32 (66.7) & 23 (47.9) & 15 (31.3) \\

Microscopy / cytology &
23 & 2 (8.7) & 7 (30.4) & 5 (21.7) & 8 (34.8) &
20 (87.0) & 16 (69.6) & 16 (69.6) & 4 (17.4) \\

Nuclear medicine &
4 & 0 (0.0) & 1 (25.0) & 1 (25.0) & 2 (50.0) &
4 (100.0) & 4 (100.0) & 2 (50.0) & 0 (0.0) \\

\bottomrule
\end{longtable}

\vspace{2pt}

\scriptsize
\textit{Note.} Percentages use the number of studies in each modality as the denominator. Modality and mechanism-family assignments were multi-label. FM denotes foundation model; CT, computed tomography; MRI, magnetic resonance imaging; U1, a frozen FM with a trained downstream interface; U2, parameter-efficient fine-tuning; U4, full fine-tuning; U6, mixed or sequential regimes; and External, confirmed independent external validation. RLA, OLA, and ASA denote representation-, objective-, and architectural/sequence-level adaptation. Other update regimes and mechanism families are omitted here for space and are reported in the complete supplementary matrices. Nuclear-medicine percentages should be interpreted cautiously because $n=4$.

\endgroup

\subsection{Validation, Reporting, and Evidence Appraisal}
\label{subsec:validation_quality}

Confirmed independent external validation was identified in 173 studies (24.5\%). Robustness or generalization evidence was identified in 191 studies (27.1\%), efficiency or resource evidence in 151 (21.4\%), and calibration or uncertainty evidence in only 10 (1.4\%). These categories reflect assessments identifiable from the reported study-level evidence and do not establish that an unreported analysis was necessarily absent. Table~\ref{tab:validation_reporting} summarizes the corresponding validation and reporting characteristics.

Reporting of adaptation-related parameter requirements was limited. Trainable parameter counts or fractions were quantified in 68 studies (9.6\%), described without numerical quantification in 6 (0.9\%), and not reported in 631 (89.5\%). At the descriptive extraction level, a principal comparator was identifiable in 652 studies (92.5\%), although identification of a comparator did not by itself satisfy the stricter nested comparative criterion.
Under the design-aware appraisal framework, risk of bias was rated high in 46 studies (6.5\%) and as having some concerns in 659 (93.5\%). Applicability concern was low in 244 studies (34.6\%), some concerns in 316 (44.8\%), moderate in 128 (18.2\%), and high in 17 (2.4\%). Overall evidence strength was classified as moderate in 655 studies (92.9\%) and low in 50 (7.1\%), with no study meeting the prespecified criteria for high evidence strength. Collectively, these findings indicate that numerical task performance was commonly reported, whereas evidence relevant to external transport, calibration or uncertainty, and adaptation-specific resource requirements was substantially less frequent. These patterns characterize the depth and reporting of the available evidence rather than comparative or clinical effectiveness.

\begingroup
\fontsize{8.4}{10.5}\selectfont
\renewcommand{\arraystretch}{1.15}
\setlength{\tabcolsep}{2pt}
\setlength{\LTleft}{0pt}
\setlength{\LTright}{0pt}
\Needspace{.5\textheight}
\begin{longtable}{@{}P{2.60cm}P{3.40cm}P{1.40cm}P{1.80cm}P{5.80cm}@{}}
\caption{Observed validation and reporting characteristics of the foundation-model adaptation evidence base. Counts indicate whether prespecified evidence or reporting fields were identified and do not establish clinical utility, safety, regulatory authorization, or comparative superiority of an adaptation strategy.}\label{tab:validation_reporting}\\
\toprule

\textbf{Evidence or reporting domain} &
\textbf{Operational criterion} &
\textbf{Studies, n (\%)} &
\textbf{Denominator, n} &
\textbf{Interpretive boundary} \\

\midrule
\endfirsthead
\multicolumn{5}{@{}l}{\textit{Table \thetable\ (continued)}}\\[3pt]
\toprule

\textbf{Evidence or reporting domain} &
\textbf{Operational criterion} &
\textbf{Studies, n (\%)} &
\textbf{Denominator, n} &
\textbf{Interpretive boundary} \\

\midrule
\endhead
\endfoot
\endlastfoot

Confirmed independent external validation
&
Named independent external cohort or source meeting the prespecified study-level confirmation rule
&
173 (24.5\%)
&
705
&
Demonstrates evaluation beyond development data under the coded definition. It does not by itself establish temporal transport, prospective utility, workflow benefit, or clinical safety.
\\

\addlinespace

Robustness or generalization evidence
&
Non-placeholder robustness, domain-shift, perturbation, or generalization evidence identified in the prespecified study-level extraction
&
191 (27.1\%)
&
705
&
Evaluated shifts and endpoints were heterogeneous. Presence of robustness evidence does not constitute a common robustness standard or support ranking of adaptation mechanisms.
\\

\addlinespace

Calibration or uncertainty evidence
&
Explicit calibration or uncertainty endpoint identified in the prespecified study-level extraction
&
10 (1.4\%)
&
705
&
The small and heterogeneous evidence base does not support comparative conclusions about calibration, uncertainty, or abstention across adaptation families.
\\

\addlinespace

Efficiency or resource evidence
&
Explicit evidence on parameters, memory, computation, communication, runtime, annotation, or other resource requirements
&
151 (21.4\%)
&
705
&
Reporting a resource quantity does not establish a matched efficiency advantage or imply lower clinical, validation, cybersecurity, or regulatory risk.
\\

\addlinespace

Trainable parameters quantified
&
At least one explicit numerical trainable-parameter count or fraction was reported for a retained pathway; an explicit zero was counted only when no model or added parameters were optimized.
&
68 (9.6\%)
&
705
&
Parameter count describes implementation scope. It is not a measure of behavioral change, clinical safety, validation sufficiency, or regulatory burden.
\\

\addlinespace

High evidence strength
&
Met the prespecified high-evidence-strength category in the study-level appraisal
&
0 (0.0\%)
&
705
&
No study met the prespecified high-strength category. This limits high-strength comparative conclusions but does not imply that studies were uniformly weak or that adaptation strategies are equivalent.
\\

\bottomrule

\end{longtable}

\vspace{2pt}

\scriptsize
\textit{Note.} Counts use the frozen 705-study evidence map and prespecified study-level extraction definitions. ``Reported'' indicates that qualifying information was identified in the extracted evidence and does not imply that an unreported analysis was not performed. Complete study-level fields are provided in Supplementary Data S1, with reporting definitions and counts in Supplementary Data S8.

\endgroup

\subsection{Comparative Evidence and Publication-Status Sensitivity}
\label{subsec:comparative_results}

Of the 705 studies in the descriptive evidence map, 473 (67.1\%) met the nested comparative criterion. A further 220 (31.2\%) did not meet the criterion, while comparative eligibility remained unresolved for 12 studies (1.7\%); these 12 studies were excluded from the comparative synthesis. Of the 473 comparative studies, 322 (68.1\%) had a formal journal, conference, or book publication and constituted the primary comparative synthesis. The remaining 151 (31.9\%) were preprint-only and were retained for publication-status sensitivity analysis. The formally published comparative synthesis therefore represented 45.7\% of the full evidence map.

Among comparative studies, at least one qualifying externally validated arm was identified in 97 of 322 formally published studies (30.1\%) and 41 of 151 preprint-only studies (27.2\%). External-validation status remained unclear for five formally published studies and one preprint-only study, respectively. All 138 comparative positives were among the 173 studies classified as having confirmed independent external validation in the parent evidence map. The smaller comparative count reflects the additional requirement that external validation be attributable to a qualifying comparative arm. The same independence and no-target-adaptation criteria were otherwise applied across the parent evidence map and comparative arm ledger.

No broader publication-status difference is inferred from these descriptive proportions. The operational rules for comparative eligibility and external-validation attribution are described in the Supplementary Reproducibility Appendix, and Supplementary Data S18--S22 provide the comparative dispositions, arm-level records, results, sources, and validation bridges.
The nested analytical layers therefore reflect progressively stricter inclusion and attribution requirements rather than estimates of a common causal effect. This structure permits broad characterization of the contemporary evidence base while limiting the principal comparative interpretation to formally published evidence and retaining preprint-only studies as a sensitivity layer.

% Section 3
\section{Taxonomy of Adaptation Strategies}
\label{sec4:taxonomy}

\begin{figure}[!htbp]
\centering
\includegraphics[width=\linewidth]{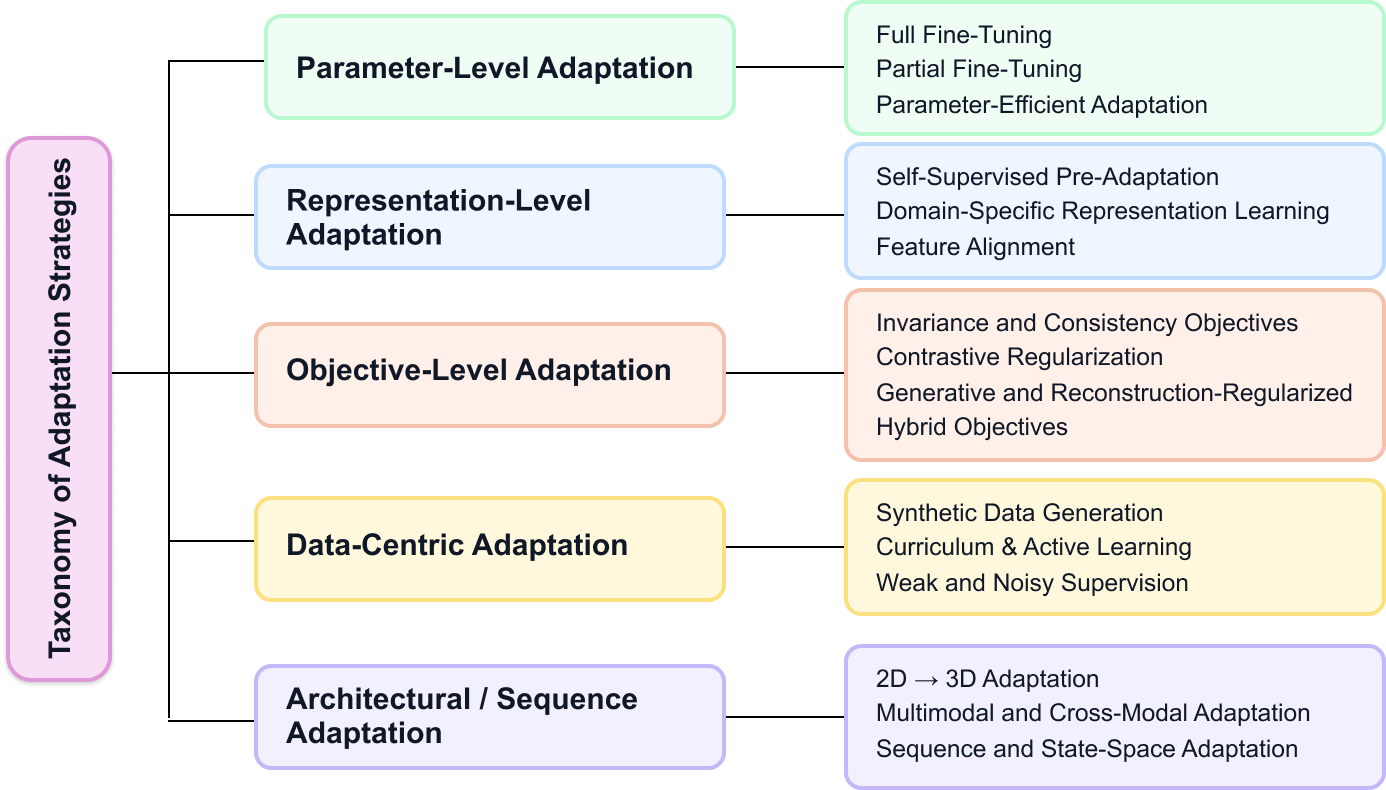}
\caption{Mechanism-centered taxonomy of foundation-model adaptation in medical image analysis. Model-side loci comprise parameter-level adaptation (PLA), representation-level adaptation (RLA), and architectural/sequence-level adaptation (ASA), while objective-level adaptation (OLA) and data-centric adaptation (DCA) describe learning-control modifiers. Mechanism families may overlap and are interpreted separately from the model use or update regime.}
\label{fig:taxonomy}
\end{figure}

This section presents the mechanism-centered taxonomy used to organize post-pretraining FM adaptation in MIA (Figure~\ref{fig:taxonomy}). The framework describes \emph{what aspect of the adaptation pathway is modified}, rather than ranking strategies by adaptation depth, effectiveness, robustness, or clinical readiness. Three families represent model-side loci of intervention: parameter-level adaptation (PLA), representation-level adaptation (RLA), and architectural/sequence-level adaptation (ASA). Objective-level adaptation (OLA) and data-centric adaptation (DCA) represent learning-control modifiers that alter the optimization signal or effective training data. These families are non-mutually exclusive and may occur together within the same adapted system. Model use or update regime is coded separately and therefore does not follow directly from mechanism-family assignment.

The taxonomy is intended as a descriptive abstraction rather than a set of modality-independent design rules. The same mechanism may have different implications across modalities because acquisition physics, spatial dimensionality, annotation structure, and sources of distribution shift differ across CT, MRI, radiography, ultrasound, histopathology, and other imaging settings. Mechanism-specific evidence is therefore interpreted in relation to the task, modality, supervision, comparator, and validation setting in which it was evaluated. Table~\ref{tab:taxonomy_riskview} summarizes the conceptual problem addressed by each family, its plausible mechanism, associated mechanism-level risks, and corresponding evaluation implications. These entries are author-synthesized interpretations and should not be read as comparative rankings or fixed properties of individual adaptation strategies.

\subsection{Parameter-Level Adaptation}\label{subsec:pla}

PLA comprises interventions in which pretrained FM parameters are updated or adaptation-specific trainable parameters are introduced and optimized as part of the post-pretraining intervention. Training an ordinary downstream interface on otherwise fixed FM representations is treated separately under the model use/update regime and does not, by itself, constitute direct FM-adaptation evidence. Within this mechanism family, the extent of pretrained-model updating varies substantially, ranging from parameter-efficient adaptation to partial and full fine-tuning. Full, partial, and parameter-efficient updating correspond to separately coded model use/update regimes rather than subclasses of PLA. They are discussed together here through a parameter-level lens because each involves adaptation-specific trainable parameters, although mechanism-family and update-regime assignments remain separate in the coding framework. The extent of updating describes adaptation capacity rather than expected effectiveness, and does not imply a monotonic relationship with performance, robustness, or clinical utility.

The relevant distinction is therefore \emph{where and how much adaptation capacity is made trainable}, not a presumed ordering of method quality. Updating more parameters may permit correction of deeper domain mismatch, but also gives a small or homogeneous adaptation set greater ability to distort broadly transferable features. Restricting updates can reduce optimization and storage burdens, but can prevent correction of mismatched or biased pretrained features. Trainable-parameter count is consequently an intervention descriptor rather than evidence of accuracy, robustness, retained knowledge, or clinical utility.

\subsubsection{Full Fine-Tuning}

Full fine-tuning updates all or nearly all parameters of a pretrained model during downstream optimization, representing the greatest extent of pretrained-parameter updating among the conventional downstream fine-tuning regimes. Because the pretrained backbone remains broadly modifiable, this regime provides substantial capacity for target-domain optimization. Across direct medical-imaging FM comparisons, however, its advantage was conditional on the task, available labeled data, comparator, and validation distribution \citep{huix2024naturaldomain,pai2024cancerfm,enda2026pathologyfm}.
For example, Huix et al. found consistent gains over frozen-feature learning across four classification datasets when DINOv2 was fully updated, although evaluation was limited to dataset-specific train/validation/test splits without independent external cohorts \citep{huix2024naturaldomain}. In the 3D CT FM evaluated by Pai et al., full fine-tuning achieved numerically higher balanced accuracy than frozen-feature classification at the full-data setting (0.804 versus 0.779), but showed a larger performance decline as labeled data were reduced and lost significant superiority over baseline implementations from 20\% onward \citep{pai2024cancerfm}. More importantly, internal gains did not necessarily translate under distribution shift. In brain-tumor histopathology, Enda et al.\ observed similar local performance between fully fine-tuned and frozen UNI models, whereas linear probing of frozen UNI features performed better on the external EBRAINS cohort \citep{enda2026pathologyfm}. Together, these findings show that full updating can improve in-domain discrimination in some settings, but do not establish consistent advantages in label efficiency or external generalization.

Broader transfer-learning literature identifies plausible risks of overfitting, catastrophic forgetting, and distortion of pretrained representations when adaptation data are limited or distributionally narrow \citep{ref68_peft_review,ref69_unilm_ft_text,ref64_ft_distor_feat}. These risks are particularly relevant in MIA, where institutional, acquisition, and population differences can contribute to substantial external distribution shift \citep{zech2018variable,ref65_transfusion}. However, such failure modes are empirical possibilities rather than inherent consequences of full updating. Full fine-tuning is therefore best characterized as a high-capacity adaptation regime whose observed value depends on supervision, backbone, task, comparator, and validation setting rather than on update extent alone.

\subsubsection{Partial Fine-Tuning}
Partial fine-tuning updates a selected subset of parameters of an eligible pretrained FM while freezing the remainder \citep{ref68_peft_review}. Within the separately coded model-use/update regime, it differs from frozen-backbone interface learning, in which the pretrained FM remains unchanged and only a downstream component is trained. Partial updating therefore permits task-specific modification of selected pretrained parameters while leaving the remainder fixed. Implementations commonly involve selective unfreezing of pretrained layers, blocks, or submodules. Related strategies such as gradual unfreezing originated in broader transfer-learning settings \citep{ref69_unilm_ft_text,ref70_chain_thaw}, but their value in medical-imaging FMs depends on the pretrained model, target domain, data regime, and choice of trainable components.

Medical-imaging FM evidence suggests that this choice can be consequential rather than uniformly beneficial. In OCT classification with RETFound, Chakrapani et al.\ reported higher cross-validated AUC after unfreezing the top four or eight transformer blocks than with a frozen encoder (0.9971 and 0.9943 versus 0.9869), although the two partial-update configurations did not differ significantly and evaluation of the selected model remained single-center \citep{chakrapani2026retfounddril}. Conversely, Nouyed et al.\ found that updating only the final two transformer blocks produced little change relative to frozen probing for MedSigLIP (94.8\% versus 94.4\% accuracy) but substantially lower performance for RETFound (77.1\% versus 86.8\%; $p<0.001$) under the same evaluation design \citep{nouyed2026retinalfm}. These contrasting results indicate that the value of partial updating can be backbone- and configuration-dependent. Strong performance from partially frozen systems should also not be interpreted automatically as evidence for the freezing strategy itself. For example, MedSAM retained the prompt encoder frozen while updating the image encoder and mask decoder and achieved substantially higher external segmentation performance than unadapted SAM and task-specific comparators \citep{ref31_medsam}. However, the adaptation simultaneously involved large-scale training across multiple imaging modalities, so the contribution of selective freezing was not isolated. Overall, partial fine-tuning represents an intermediate, configurable update regime between frozen use and full fine-tuning, but the current evidence does not establish an inherent advantage in performance, generalization, or preservation of pretrained representations. Its value appears contingent on which components are updated, the pretrained backbone, adaptation data, comparator, and validation setting.

\subsubsection{Parameter-Efficient Adaptation}
Parameter-efficient adaptation constrains learning to a small set of task-specific parameters while leaving most pretrained backbone weights unchanged \citep{ref96_peft_mia, ref95_peft_survey}. Rather than fully updating the backbone, these methods introduce or selectively tune lightweight components such as adapters \citep{ref53_medsa, ref82_standard_adapter}, prompts \citep{ref89_prompt_tuning, ref88_prompt_tuning1}, bias terms \citep{ref97_bitfit}, or low-rank parameterizations such as Low-Rank Adaptation (LoRA) \citep{ref100_lora}. These mechanisms substantially reduce the trainable-parameter and per-task storage burden relative to full fine-tuning \citep{ref68_peft_review, ref80_delta_tuning}.

Medical-imaging FM studies show that large reductions in trainable parameters can support substantial adaptation gains, but do not imply superiority over full updating. Med-SA updated approximately 2\% of SAM ViT-H parameters and substantially improved BTCV multi-organ segmentation over unadapted SAM (mean Dice 88.3\% versus 54.8\%) \citep{ref53_medsa}. However, the gain cannot be attributed to parameter efficiency alone because Med-SA combined several adaptation components, and the full-fine-tuning comparisons were not matched experiments differing only in the extent of parameter updating. A more direct comparison was provided by DVPT, which adapted a frozen CLIP ViT-B/16 using fewer than 0.5 million trainable parameters and modestly exceeded full fine-tuning on DDR diabetic-retinopathy grading (76.40\% versus 75.56\% accuracy), although no independent external cohort was evaluated \citep{ref89_prompt_tuning}. In volumetric CT segmentation, Silva-Rodr\'{\i}guez et al.\ found that rank-four LoRA used only 68.1 thousand trainable parameters compared with 62.1 million for full fine-tuning, but the performance ordering depended on supervision: LoRA was slightly worse with one labeled volume and slightly better with ten or thirty \citep{silvarodriguez2025fseft}. Thus, substantial parameter savings can coexist with competitive performance, but the relative benefit of PEFT is configuration- and data-dependent.

Across the broader medical-imaging PEFT literature, differences in backbones, trainable components, comparators, label regimes, and validation designs preclude a general ranking of adapters, prompts, bias tuning, and low-rank methods \citep{ref96_peft_mia,ref86_brain_adapter,ref127_masam,ref54_SAMed,ref130_3dsam_adapter}. Nor do reductions in trainable parameters establish preservation of pretrained representations, reduced catastrophic forgetting, improved calibration, or better external generalization. Conversely, retaining most of the backbone frozen may restrict adaptation when pretrained representations are poorly aligned with the target domain, although this limitation must be demonstrated empirically rather than inferred from parameter count alone \citep{ref65_transfusion,ref76_how_transferable_dl,ref91_revisiting_peft}. Parameter-efficient adaptation is therefore most directly characterized by its reduced trainable-parameter footprint. Its comparative value remains dependent on the backbone, task, available supervision, comparator, and validation setting.

Across full, partial, and parameter-efficient updating, the comparative evidence does not support a monotonic relationship between the number of parameters updated and performance or generalization. Accordingly, update depth functions as an empirical design choice rather than a proxy for adaptation quality or clinical readiness. For deployment-oriented interpretation, trainable-parameter footprint alone is insufficient as the relevant evidence concerns the performance of the complete versioned system under intended-use-specific external validation.

\subsection{Representation-Level Adaptation}
\label{subsec:rla}

RLA is a model-side locus comprising post-pretraining interventions that explicitly reorganize, align, transform, or reweight an existing FM feature space for a downstream medical-imaging domain or task. It is defined by the representational target rather than by update depth. RLA can operate around a frozen encoder or coexist with PEFT, partial updating, or full fine-tuning. Parameter updating that merely causes the representation to change does not constitute RLA by itself, because the intervention must explicitly target the task-facing feature space. Because the taxonomy is multi-label, the same study may also carry PLA, ASA, OLA, or DCA codes. Placement within the RLA narrative indicates that modification of the task-facing representation is the principal mechanism being discussed, without implying an exclusive or primary study-level code.

For narrative synthesis, we organize direct RLA evidence into three overlapping mechanism groups: self-supervised pre-adaptation, domain-specific representation learning, and feature alignment across domains and modalities. These groups are not mutually exclusive. A frozen FM used only as an unchanged feature extractor does not itself constitute RLA, and representation learning used solely to construct the original FM remains foundational pretraining. Direct RLA requires an active intervention on an already pretrained representation whose resulting transformation remains part of the evaluated downstream pathway. Such realignment may improve a selected task while erasing transferable information, amplifying acquisition shortcuts, or narrowing clinically useful heterogeneity \citep{banerjee2023shortcuts}. Whether such realignment improves label efficiency, external generalization, robustness, or clinical utility remains an empirical question rather than an inherent property of the mechanism.

\subsubsection{Self-Supervised Pre-Adaptation}

Self-supervised pre-adaptation introduces an additional representation-learning stage to an already pretrained FM using unlabeled or automatically structured medical data before downstream task optimization \citep{ref59_big_ssl,ref1_3d_ssl_methods_medical}. It is discussed primarily as RLA when that stage produces the refined task-facing representation. The associated contrastive, reconstruction, or consistency loss can simultaneously receive an OLA code, but the analytical distinction is whether the principal product is a transformed representation or an auxiliary constraint on downstream optimization. Self-supervision used to construct the original FM remains pretraining rather than adaptation.

\paragraph*{Contrastive Representation Learning}

Contrastive RLA realigns an existing FM representation by defining which images, views, or image--text pairs should become similar or distinct \citep{ref22_clip}. The available evidence supports this strategy in selected settings, but attribution varies across studies. Qiu et al.\ provided relatively direct same-model evidence: task-adaptive contrastive continuation increased QuiltNet quadratic weighted kappa on SICAP from 0.02 to 0.50 \citep{qiu2025effortless}. However, effects varied across model--task combinations and sometimes declined when additional lower-ranked image--caption pairs were introduced, indicating that benefit depends on pair selection as well as the objective itself.

ALTA and CPLIP showed stronger performance after broader cross-modal adaptation, but neither isolated the contrastive contribution. ALTA achieved zero-shot RSNA Pneumonia AUC of 91.0\% versus 89.3\% for BioViL-T \citep{lian2025alta}, while CPLIP achieved balanced accuracy of 0.701 on CRC100K versus 0.566 for CONCH \citep{javed2024cplip,ref111_villan_fm}. In both cases, differences in starting models, data construction, adaptation components, and auxiliary objectives mean that the results support the complete pipelines rather than contrastive alignment alone.

Contrastive RLA can improve selected downstream endpoints when the imposed similarity structure is informative for the evaluated task, but the available evidence does not establish uniform benefit. Noisy captions, generated descriptions, coarse labels, or site-specific cues may provide easier alignment signals than pathology, and poorly defined negatives may separate clinically related findings \citep{banerjee2023shortcuts,ong2024shortcut}. Improved alignment or benchmark performance does not by itself establish robustness or external transport without matched comparisons, component ablations, and independent validation.

\paragraph*{Generative or Reconstruction-Based Approaches}

Reconstruction-based RLA uses reconstruction or generative recovery to learn a transformed FM representation before that representation is supplied to a downstream task. It may also carry an OLA code, but differs from reconstruction whose residual or generated output directly contributes to task prediction. Reconstruction used solely to construct the original FM remains foundational pretraining \citep{ref51_mae,ref114_simmim}.

In medical imaging, Chung et al. provide a direct two-stage example. Model-specific autoencoders reconstructed retained UNI and other pathology-FM features while a cosine-similarity term aligned their bottleneck spaces. The learned encoder was then used to provide transformed features to the downstream classifier. For early-gastric-cancer lymph-node-metastasis prediction, AUC on an author-described 46-case external cohort increased from 0.63 with original UNI features to 0.71 with the transformed representation \citep{chung2025autoencoder}. Internal cancer-region segmentation accuracy was 0.984, indicating that the observed classification gain should not be generalized across downstream tasks. Because reconstruction and cross-model bottleneck alignment were optimized jointly, the gain supports the complete transformation rather than reconstruction alone. The external source was also insufficiently identified for confirmation under the review's strict external-validation rule.

DiffuGTS provides a complementary example in which the internal representation of a frozen medical diffusion FM was repurposed for tumor segmentation through feature adaptation and latent inpainting \citep{jiang2025diffugts}. Taken together, these studies show that reconstruction and generative mechanisms can redirect pretrained representations, but reconstruction fidelity itself is not evidence of clinically useful representation learning. Their value depends on what information is reconstructed, the starting representation, adaptation data, and whether gains persist under independent external evaluation rather than reflecting anatomical redundancy, acquisition characteristics, or other nuisance structure. Hybrid approaches discussed next combine reconstruction with contrastive, multimodal, hierarchical, or region-level objectives to target complementary representational signals.

\paragraph*{Hybrid Representation-Learning Objectives}

Hybrid adaptation combines complementary learning signals within an existing FM pathway, for example contrastive alignment with reconstruction, masked modeling, teacher regularization, or multimodal supervision. Such combinations are intended to capture complementary representational information, but the available medical-imaging FM evidence does not establish that adding objectives produces intrinsic synergy.
RetFiner refined pretrained retinal FM representations using paired OCT images and electronic-health-record text, jointly optimizing image--text contrastive alignment, image--text matching, masked-language modeling, and autoregressive report generation. The adapted RETFound representation improved average balanced accuracy across seven OCT classification datasets from 78.1\% to 83.8\% \citep{fecso2026retfiner}. A matched loss ablation also favored the combined objective over individual losses, providing evidence of complementarity within that pipeline. However, the ablation used a separately pretrained backbone rather than the RETFound model underlying the principal comparison.

GLINT similarly combined sparsely gated vision--language alignment with dense-feature regularization to a frozen SSL teacher. Component analyses indicated that the complete configuration benefited from interaction between the alignment and regularization components rather than from the auxiliary constraint alone \citep{park2026glint}. This supports complementarity within the evaluated design, but not a general advantage of adding regularization to contrastive adaptation.

ASTRA provides a complementary pathology example, combining reconstruction across multiple frozen FM feature spaces with subsequent semantic alignment. It achieved a mean Dice of 0.738 for text-guided tumor localization on the external TCGA cohort without cohort-specific fine-tuning \citep{wang2026astra}. However, the contribution of the hybrid objectives remained entangled with multi-FM feature access, spatial contextualization, reconstruction, and semantic supervision. Together, these studies show that multiple learning signals can be combined productively, but ``hybrid'' describes an adaptation package rather than a mechanism with guaranteed benefit. Attribution of synergy would require matched component ablations from common starting weights and data, because aggregate gains from a multi-component pipeline do not establish such interaction.

\subsubsection{Domain-Specific Representation Learning}

Domain-specific RLA reorganizes an existing FM representation using explicit medical structure, such as clinical concepts, anatomy, modality relationships, spatial context, or teacher knowledge. Its defining feature is not merely the use of medical data, but that this structure explicitly shapes the task-facing representation. The same intervention may also involve parameter-, objective-, architecture-, or data-level mechanisms. The relevant question is whether the imposed medical structure produces an attributable task or transport benefit rather than simply a more ``domain-aligned'' representation.

MONET adapted pretrained CLIP-derived image and text encoders using dermatology image--text pairs and organized the resulting representation around named skin concepts. Mean concept-recognition AUROC increased from 0.692 for unadapted CLIP to 0.766 for MONET \citep{kim2024monet}. This supports improved alignment with the evaluated dermatology concepts, but does not establish that the selected vocabulary captures all clinically relevant variation or transports across settings. The reported concept evaluation used Fitzpatrick17k and Diverse Dermatology Images. It did not constitute site-stratified independent clinical validation under the review's external-validation criteria.
G2L imposed teacher-guided specialization by distilling H-optimus-0 into the smaller, already pretrained Hibou-L model using cancer-specific pathology slides. On TILS, adapted Hibou-L achieved accuracy of 0.9362, close to 0.9344 for the substantially larger teacher \citep{cho2025g2l}. This shows that the adapted student achieved performance comparable to the substantially larger teacher, but does not by itself isolate the gain from specialization relative to unadapted Hibou-L. Distillation may also transfer the teacher's representational priorities and blind spots, and the reported result does not establish improved transport across centers or rare morphologies.

NasVLM illustrates an important boundary case. Its multi-view reconstruction and image--report alignment were used to construct the original nasal FM and are therefore contextual pretraining evidence rather than direct post-pretraining RLA \citep{liu2025nasvlm}. Separately evaluated downstream adaptation of the completed model may remain eligible under its own intervention, but should not be used to attribute benefit specifically to domain-specific RLA.
Domain-specific RLA is therefore better viewed as controlled specialization than as an inherently more medical representation. Its value depends on whether the imposed concepts, teachers, anatomical relations, or multimodal correspondences preserve clinically useful variation rather than narrowing the representation around task-, terminology-, institution-, or acquisition-specific structure.

\subsubsection{Feature Alignment Across Domains and Modalities}

Feature alignment is an RLA strategy in which an explicit loss, projection, prototype, transport rule, or discrepancy constraint reduces representational differences associated with scanner, institution, acquisition protocol, population, or imaging modality. It differs from simply training in one domain and evaluating in another as the intervention must explicitly target the domain discrepancy. Complete invariance is not necessarily desirable, because domain-associated variation may contain clinically meaningful as well as nuisance information.

In medical imaging, ScanGen trained a projection network around frozen pathology-FM features using a scanner-aware contrastive objective. For Phikon-based EGFR prediction, AUC increased from 0.738 to 0.768 while scanner-associated coefficient of variation decreased from 0.389 to 0.121 \citep{carloni2025scangen}. The concurrent improvement in discrimination and reduction in scanner-associated variation is compatible with useful nuisance suppression, but does not establish transport to a new institution because the principal evaluation remained within a multi-scanner cohort.

Vardi et al.\ combined LoRA adaptation of UNI with local maximum mean discrepancy minimization across hospitals. On NSCLC, balanced accuracy increased from 78.84\% to 84.88\% when unlabeled target-hospital data were available and from 81.61\% to 83.79\% when generalizing to hospitals unseen during training \citep{vardi2026discrepancy}. The larger target-informed gain illustrates an important distinction. Adaptation to an observed target distribution can directly reshape features toward that domain, whereas domain generalization must identify nuisance variation without access to the future target. These results support alignment in both settings but do not establish that the same intervention will remain beneficial across tasks or institutions. Under the review's prespecified external-validation rule, the target-informed experiment does not constitute independent external validation because target-domain data were available during adaptation, whereas the unseen-hospital experiment addresses domain generalization without access to the future target.

RFD provides a complementary cross-modality example, combining SAM-based adapters, prototype alignment, transport, and feature-discrepancy reduction for unsupervised CT--MRI segmentation \citep{xia2026rfd}. Its reported MMWHS results support the complete cross-modality adaptation pipeline, but the individual contribution of discrepancy reduction cannot be isolated because the architecture, prototypes, transport mechanism, and alignment objective changed together. More fundamentally, successful CT--MRI alignment requires preserving modality-specific contrast while aligning shared anatomy rather than forcing the two representations to become indistinguishable.

The principal risk is therefore \emph{over-alignment}. Scanner, site, protocol, population, and modality effects can interact with lesion visibility, tissue contrast, disease prevalence, and labeling practice. Reducing a global domain discrepancy may suppress clinically useful variation or align samples that are similar in acquisition characteristics but different in pathology \citep{geirhos2020shortcut,banerjee2023shortcuts}. Lower discrepancy, weaker scanner predictability, or visually mixed embeddings are therefore intermediate properties rather than evidence of clinical robustness.

Feature alignment is therefore best characterized as targeted control of specified domain variation rather than as a generic route to invariance. Its value depends on whether task-relevant information is preserved while nuisance variation is reduced. Robustness or transport is established by performance in genuinely unseen environments rather than by improved latent similarity alone.

\subsection{Objective-Level Adaptation} \label{subsec:ola}

OLA changes what an eligible retained FM is encouraged to preserve, suppress, align, or discriminate during post-pretraining optimization. It is therefore analytically distinct from the model-side loci such as PLA, RLA, and ASA and can coexist with any of them or with DCA. The mathematical form of a loss does not by itself determine the primary narrative placement of a study. When contrastive or reconstruction learning principally creates or transforms the task-facing representation before a downstream task, it is discussed under RLA even though an OLA code may also be present in the multi-label evidence table. The present section focuses on objectives used as task-coupled constraints during downstream adaptation, including generative objectives whose trained outputs subsequently participate in the inference rule. Within that boundary, consistency specifies which variation should be ignored, contrastive regularization defines which cases should be similar or dissimilar, generative objectives regulate what is preserved or normalized, and hybrid objectives coordinate several such assumptions. OLA is thus control over inductive bias, not evidence that the rewarded behavior is clinically appropriate.

\subsubsection{Invariance and Consistency Objectives}

Invariance and consistency objectives encourage stable predictions or task-facing representations across perturbations assumed to preserve the clinical target. These may involve image augmentation, prompt variation, paired modality views, or teacher--student predictions, but they encode different equivalence relations and are not interchangeable. Their value depends on whether the imposed transformation represents true nuisance variation rather than diagnostically relevant change \citep{gao2024pmt,hu2025better,xu2022adversarial,zhou2022generalizable}. The central risk is not insufficient consistency, but enforcing consistency across the wrong variation.

Direct FM adaptations illustrate this dependence on the chosen constraint. PromptMed kept a pretrained CLIP ViT-B/16 backbone frozen and optimized 601,148 prompt and classifier parameters using class-balanced prompt perturbation and contrastive feature consistency. With 100 labeled NCT-CRC-HE examples, adding the contrastive feature-consistency term to the otherwise matched prompt-noise configuration increased accuracy from 96.75\% to 97.73\% \citep{wang2026promptmed}. This supports the term within that pipeline, but only for augmentations and prompt noise assumed to preserve the histology class. It also illustrates that learning-control modifiers can overlap conceptually. Prompt group-aware SAM3 reached Dice 0.7942 on PanNuke versus 0.7845 for the conventional fine-tuning baseline, while removing the proposed consistency term reduced Dice to 0.7772 and naive all-pairs consistency produced 0.7738 \citep{wu2026promptgroup}. Thus, the form of consistency mattered more than the label ``consistency'': grouping prompts according to a defined semantic relation was helpful, whereas treating every prompt pair as equivalent was not.

The same issue becomes more pronounced across modalities or pseudo-labels. Semi-MedSAM retained pretrained SAM encoders in a teacher--student pathway and enforced agreement over paired white-light and narrow-band endoscopy. With 20\% labeled data, the complete system achieved Dice 0.775 on Dataset~I versus 0.744 for SAM-Med2D in the same semi-supervised setting. After development on Dataset~I, Dice on independent Dataset~III was 0.769 versus 0.744 \citep{li2026semimedsam}. These results support the complete multimodal, adapter, pseudo-label, and consistency system rather than consistency alone. Mechanistically, agreement can exploit complementary views when both preserve the same anatomical target, but registration error, modality-specific visibility, or incorrect teacher predictions can make the imposed equivalence clinically invalid. Consistency may then stabilize error as readily as useful information, while aggressive augmentation can suppress lesion texture or boundaries even when those features appear nuisance-like at image level \citep{zhao2019learning,cho2023anti,qi2025mediaug}.

Invariance and consistency objectives therefore specify which variation the adapted model should ignore, and they do not establish that the ignored variation is clinically irrelevant. Stronger invariance is not intrinsically more robust. Credible evaluation should isolate the consistency term under common starting weights, verify that each transformation preserves pathology and boundaries, examine pseudo-label confirmation errors and rare classes, and test whether the selected equivalence survives external sites and clinically authored prompts. None of these examples established calibrated operating thresholds or prospective clinical benefit.

\subsubsection{Contrastive Regularization During Adaptation}

Here, contrastive regularization denotes a contrastive term coupled to downstream adaptation, encouraging task-defined positives to converge and selected negatives to separate. It may coexist with RLA when it also reorganizes the representation, but it differs narratively from a separate contrastive realignment stage and from the upstream contrastive pretraining used to construct ConVIRT, GLoRIA, or MedCLIP \citep{ref_112_convirt,ref125_gloria,ref32_medclip}. The critical issue for OLA is whether the imposed similarity structure corresponds to clinically meaningful similarity rather than merely producing better-separated embeddings.

The direct evidence suggests that the effect is task- and granularity-dependent. In pediatric brain-tumor whole-slide classification, a frozen UNI2-h encoder was coupled to a trained projection and CLAM interface. Adding a queue-based supervised contrastive term increased seven-class macro recall from 0.705 to 0.768 and three-class macro recall from 0.625 to 0.658, but did not consistently improve the near-ceiling binary task \citep{nguyen2026clinically}. In weakly supervised nuclei segmentation, an instance-aware contrastive term around a frozen SAM encoder produced smaller same-pipeline increments: CPM17 Dice increased from 80.04\% to 80.38\%, while MoNuSeg Dice increased from 74.51\% to 75.22\% \citep{nam2026selfrefining}. These findings are compatible with contrastive pressure being most useful when difficult class or instance relationships remain unresolved. They do not show that greater feature separation is beneficial once the pretrained representation already supports the coarse decision.

The same mechanism creates its principal failure mode because positives and negatives are clinical assumptions expressed as sampling rules. Coarse labels can pull heterogeneous diseases or grades together, while related, co-occurring, or phenotypically similar findings may be pushed apart as false negatives \citep{ref32_medclip}. Report-derived supervision can encode uncertainty, negation, or documentation convention rather than image-grounded pathology \citep{jain2021visualchexbert}. Patient, slide, scanner, or institution may also provide an easier pair-separation signal than disease \citep{zech2018variable}. Contrastive regularization can therefore sharpen embedding geometry while compressing within-class heterogeneity, separating clinically related states, or strengthening provenance shortcuts.

Contrastive regularization is therefore a mechanism for imposing a particular notion of clinical similarity, not a generally beneficial means of improving representation quality. Evidence should report patient- and site-aware pair construction, the rule defining positives and negatives, sensitivity to temperature and hard-negative selection, matched loss ablations from common starting weights, and effects on rare classes and subtypes. The available examples do not establish independent clinical-site validation, calibrated thresholds, or that the learned similarity remains valid under population or acquisition shift.

\subsubsection{Generative and Reconstruction-Regularized Objectives}

Within OLA, generative and reconstruction objectives are coupled to downstream adaptation, with their trained outputs in some cases participating directly in the inference rule. They can coexist with RLA, but a separate post-pretraining reconstruction stage whose principal product is a transformed representation is discussed primarily under RLA, as with Chung et al.; masked-autoencoder training used only to construct the original FM is further upstream \citep{ref51_mae,ref26_ssl_mae_mia}. The central OLA question is whether reconstruction preserves information needed for the clinical decision, not whether the reconstruction is visually or numerically accurate.

AnoPILaD exposes the normalization assumption directly. Stable Diffusion v1.5 was LoRA-adapted on normal gastric lymph-node patches, while retained CONCH encoders selected pathology prompts; anomaly scores were the discrepancy between the observed patch and its reconstructed normal counterpart. Patch-level AUC was 0.9587 locally versus 0.9290 for MemAE, and 0.8884 on the cross-organ Camelyon16 set versus 0.6611 for MemAE \citep{wang2026anopilad}. These results support the complete pathology-prompted reconstruction system, but the slide-level advantage on Camelyon16 was much smaller (0.6745 versus 0.6612). Mechanistically, normalization can expose pathology when abnormal structure is not reproducible from the learned normal prior, but can suppress the signal when the diffusion model reconstructs true disease or exaggerate benign acquisition differences when they lie outside that prior. The reconstruction timestep was also selected using the local test set, introducing test-set selection bias and potentially inflating the local estimate.

DiffusionCLIP demonstrates that the denoising objective can also conflict with localization. It retained a pretrained CLIP ViT-L/14 encoder and trained diffusion modules over multilevel features. Adding diffusion alone changed mean image-level AUROC from 86.1\% to 86.4\% but reduced mean pixel-level AUROC from 90.4\% to 90.1\%; only the complete diffusion-plus-dynamic-loss configuration reached 87.4\% and 92.4\%, respectively \citep{chen2025diffusionclip}. Denoising was therefore not independently beneficial: its value depended on a task-specific loss that redirected what the reconstruction pathway preserved. Because dataset provenance and evaluation independence were not sufficiently explicit for every benchmark, these results should not be treated uniformly as strict external validation.

The central failure mode is proxy mismatch: reconstruction error is not disease evidence \citep{cai2024rethinking,georgescu2023masked}. Expressive models may reproduce true pathology, while scanner texture, background, or benign anatomy may generate large residuals. Contextual feature-reconstruction evidence points to the same issue: Mammo-SAE recovered both clinically aligned and background-responsive features from Mammo-CLIP representations, so preserving FM information did not guarantee preservation of pathology-specific information \citep{nakka2025mammosae}. Evaluation should therefore examine false-negative reconstruction, lesion size and subtype, scanner and site effects, threshold calibration, and inference cost under independent external data. Better denoising or reconstruction fidelity is desirable only when what is reconstructed aligns with the clinical task.

\subsubsection{Hybrid Objectives and Inductive Bias Control}

Hybrid OLA combines task-coupled objectives that reward different properties of the adapted system, such as pixel fidelity, preference ordering, semantic consistency, contrastive correspondence, or reconstruction. It can coexist with RLA when the same losses reorganize representations, but the analytical focus here is how their joint weighting controls downstream behavior. The motivation is complementarity; the risk is incompatibility. Each loss defines a different desirable behavior, and minimizing their weighted sum does not guarantee that clinically important properties improve together.

SAMPO-Path shows why preference and task objectives are complementary but not interchangeable. Pixel supervision specified an absolute segmentation target, whereas two preference losses only required one prompt-conditioned mask to rank above another. With 10\% of PanNuke data, supervised-only IoU was 0.6529 and the complete hybrid reached 0.6828, but preference training without pixel supervision collapsed IoU to 0.2234 \citep{wu2025sampopath}. Ranking alone therefore under-specified anatomical correctness: both candidates can be poor, and a quality proxy can favor relative improvement while sacrificing boundary fidelity or small-object coverage. Excessive preference weight also impaired pixel accuracy. Because the rankings were generated from ground-truth IoU rather than clinician preference, the result supports simulated prompt-quality ordering, not alignment with heterogeneous clinical intent.

KEPIL combined classification, image--report contrastive alignment, and semantic consistency between dropout-perturbed report representations. Classification rewarded disease separation, alignment rewarded paired image--text correspondence, and consistency assumed that the removed embedding information was nuisance rather than clinical meaning. On ChestXray-14, AUC increased from 76.89\% for classification plus alignment to 79.45\% after prompt enrichment and 80.95\% after consistency was added; on CheXpert, the corresponding values were 87.07\%, 90.47\%, and 91.21\% \citep{luo2026kepil}. These staged gains support the selected combination, but strong consistency can suppress dimensions encoding negation, certainty, temporality, severity, or laterality, while contrastive alignment can preserve reporting conventions that are more stable than the visual finding. The study did not test whether the learned invariance preserved these distinctions when clinically meaningful formulations changed.

STAMP imposed a different pair of biases while LoRA-adapting Virchow2 on paired H\&E--spatial-transcriptomics patches. Morpho-molecular contrastive alignment rewarded structure shared across modalities, whereas pathway reconstruction rewarded recovery of molecular activity. The combination increased mean AUC across five actionable driver mutations from 0.777 to 0.805 internally and from 0.780 to 0.796 on independent external cohorts \citep{zhou2026stamp}. Complementarity is plausible, but the objectives need not preserve the same information: contrastive alignment can compress intratumor heterogeneity, while reconstruction can privilege prevalent, high-amplitude pathways over weak or rare molecular signals. Their individual contributions and weighting were not isolated, and prospective observational assay-bypass simulations still produced false-negative and false-positive recommendations despite retrospective threshold calibration.

Hybrid objectives are therefore mechanisms for controlling competing inductive biases, not inherently stronger forms of adaptation. Equal coefficients do not imply equal influence when loss scales and gradient magnitudes differ, and gradients may agree for common cases while conflict for boundaries, rare subtypes, uncertain reports, or heterogeneous tumor regions \citep{chen2018gradnorm,yu2020gradient}. Evidence should include the task-only anchor, single-objective additions, complete combination, coefficient and scheduling sensitivity, and property-specific outcomes such as boundary error, semantic-negation failure, rare-pathway recovery, calibration, and subgroup performance. Increasing objective complexity is justified only when the added constraints address distinct failure modes without trading away task-critical behavior.

\subsection{Data-Centric Adaptation} \label{subsec:dca}
Where PLA, RLA, OLA, and ASA primarily modify model parameters, representations, optimization objectives, or architectural/sequence structure, DCA modifies the data and supervision through which downstream adaptation occurs. It reshapes the effective training signal by changing distributional coverage, the allocation of samples or tasks, or the precision and reliability of supervision \citep{zhang2026data}. This includes synthetic data generation, curriculum and active learning, and weak or noisy supervision.

DCA has no fixed parameter-update depth and can accompany frozen-backbone adaptation, PEFT, partial updating, or full fine-tuning. Its benefit is likewise not inherent to using more data or fewer annotations as it depends on whether the intervention adds clinically relevant coverage, allocates supervision appropriately, or preserves useful information despite weaker labels. Poorly aligned data-centric interventions can instead amplify synthetic artifacts, model-defined sampling biases, reporting conventions, under-representation, or label noise.

\subsubsection{Synthetic Data Generation}
\label{subsec:synthetic}

Synthetic-data DCA introduces artificially generated images, volumes, masks, lesions, text, or other supervisory signals \emph{after} FM pretraining. It changes the data used to adapt the retained FM and can coexist with any update regime or other mechanism family. Generator development or fidelity evaluation alone is therefore insufficient unless synthetic data contribute to an attributable downstream FM adaptation. The relevant question is not whether synthetic samples appear realistic, but whether they add task-relevant variation or supervision that improves downstream performance on real data. Independent clinical evaluation provides stronger evidence that such gains transfer beyond the development setting \citep{giuffre2023harnessing,koetzier2024generating}.

Direct FM evidence illustrates several forms of this strategy. SynthFM-3D fine-tuned SAM 2.1 Hiera-B+ on 10,000 procedurally generated 3D volumes varying anatomy, texture, boundary definition, contrast, and noise. On real CT aorta segmentation, Dice increased from 68.18\% for the original checkpoint to 72.87\% after synthetic adaptation, with transfer also reported to unseen ultrasound data \citep{chakrabarty2026synthfm3d}. The same-checkpoint comparison supports the complete synthetic-training intervention, but does not identify which simulated factors produced the gain, and no equal-budget real-data adaptation was tested. Procedural synthesis may therefore broaden shape and boundary variation while still omitting pathology-dependent deformation, devices, demographic anatomy, or disease--acquisition relationships.

Synthetic supervision can also operate on the language side of a multimodal FM. Zhang et al.\ froze a 2-billion-parameter SmolVLM base and trained LoRA adapters using synthetic dermatology question--answer supervision, inherited figure captions, or both; question candidates and answers were generated through external language models. The combined QA and caption model achieved 70.20\% classification accuracy, compared with 68.24\% for the general-purpose SmolVLM-Instruct comparator \citep{zhang2026smolvlmsynthetic}. However, the adapted models began from SmolVLM-Base whereas the comparator was SmolVLM-Instruct, so this contrast does not isolate the effect of synthetic supervision. Moreover, Gemini contributed to dataset annotation/curation and synthetic QA generation, creating potential semantic dependence between the constructed supervision and evaluation setting. Its improved agreement may partly reflect teacher-specific terminology or answer structure rather than stronger visual grounding or clinical reasoning. Evaluation on curated publication figures further limits conclusions about routine clinical images and clinician-authored questions.

MediCLIP represents a third form, using synthetic abnormalities rather than complete images or language supervision. MediCLIP used CutPaste, GaussIntensityChange and Source, the latter introducing deformation-based synthetic abnormalities and masks from normal images, while prompts and visual adapters aligned retained CLIP features with normal and abnormal concepts \citep{ref146_mediclip}. Comparisons with different architectures support the overall system but cannot attribute gains specifically to synthesis. Within the same pipeline, greater synthetic diversity was also not uniformly beneficial across datasets. In the 16-shot BrainMRI setting, combining all three corruption types produced 94.8\% AUROC, compared with 95.2\% for Gaussian-intensity modification plus CutPaste. Such corruptions provide inexpensive localization supervision, but synthetic blending, intensity, or deformation cues may still be easier to discriminate than genuine disease morphology.

Across these examples, synthetic data provide configurable supervision but introduce three recurring risks: \emph{support omission}, when clinically relevant variation is absent from the generator; \emph{source shortcuts}, when synthetic artifacts or source status predict the target; and \emph{semantic circularity}, when the generation process also shapes evaluation. These risks reinforce that synthetic realism is not equivalent to clinical validity \citep{giuffre2023harnessing,koetzier2024generating,ye2024spurious,raja2024synthetic}. Visual fidelity, diversity, and privacy are likewise distinct properties and should not be treated as interchangeable evidence of usefulness.

Stronger evidence requires real-only, synthetic-only, and mixed-data comparisons using the same pretrained checkpoint, update rule, optimization budget, and real-data quantity, with transparent reporting of synthetic provenance and potential class--source confounding. Evaluation should ultimately use independent real clinical cohorts, preferably under distribution shift, and assess clinically relevant failure modes rather than fidelity alone. Current FM evidence therefore supports selected, task-aligned uses of synthetic data as a coverage and supervision mechanism, but not the broader assumption that greater synthetic volume, realism, or teacher capability necessarily produces more reliable clinical adaptation.

\subsubsection{Curriculum and Active Learning}
\label{subsec:curriculum_active}

Curriculum and active-learning DCA alter the allocation of supervision rather than its nominal label content. Curriculum learning changes the order or sampling probability of available samples, tasks, or classes, whereas active learning selects which currently unlabeled cases should next receive annotation \citep{bengio2009curriculum,9392296,wang2024comprehensive}. These mechanisms can coexist with any update regime, but the DCA claim concerns the exposure or acquisition policy itself rather than inference-time prompt sequencing or uncertainty estimation alone.

Direct FM evidence shows that such policies can improve adaptation, but also that their effect depends strongly on how informativeness or difficulty is defined. In VALIANT, frozen CLIP, BioViL, or MedCLIP checkpoints were treated as black-box FMs, while a visual-prompt decoder was learned through zeroth-order optimization and unlabeled cases were ranked by instability of their prompt representations under input perturbation, with unsupervised anatomical maps additionally informing prompt generation. On CheXpert at a 50\% label budget, the BioViL-based configuration achieved 90.7\% AUC, compared with 65.7\% for random sampling, 85.2\% for BALD, and 84.9\% for Coreset; a MedCLIP variant reached 92.6\% \citep{mahapatra2026valiant}. The reported matching of frozen FMs, data splits, budgets, acquisition-batch sizes, and seeds across active-learning baselines strengthens attribution to the overall acquisition system. It does not, however, isolate prompt instability from the accompanying visual-prompt pathway and anatomical-coherence mechanism, and perturbation sensitivity need not correspond uniquely to clinically informative cases.

A related issue appears in curriculum learning under label noise. CUFIT used retained VFMs including DINOv2 to train a linear probe on all samples, then progressively selected cases whose model predictions agreed with their observed labels for intermediate and final adapters. On HAM10000 with 40\% simulated symmetric label noise, accuracy reached 79.1\% versus 74.1\% for CoDis \citep{yu2024cufit}. This supports the complete agreement-based curriculum, but model--label agreement is only a proxy for label correctness. Difficult but correctly labeled or domain-shifted cases may be excluded, whereas systematic errors that agree with the model may be retained. Simulated symmetric corruption therefore establishes robustness to a controlled noise process rather than to reader disagreement, grading ambiguity, or other clinically structured annotation errors.

CURE extends the same principle from individual samples to tasks and semantic groups. MedGemma-4B-IT was LoRA-adapted across phrase grounding, grounded report generation, and anatomy-grounded report generation while an error-aware sampler increased exposure to underperforming datasets, anatomical regions, and semantic classes according to validation performance \citep{messina2026cure}. In a matched ablation, curriculum reweighting increased anatomy-grounded-report IoU from $0.360\pm0.009$ to $0.411\pm0.009$ and MS-CXR phrase-grounding micro-IoU from $0.398\pm0.019$ to $0.430\pm0.018$. Larger gains from the final pipeline cannot be assigned to curriculum alone because additional pre-adaptation and hyperparameter optimization were introduced. More importantly, placing validation performance inside the training-control loop means that the chosen metric and reweighting schedule themselves determine which failures receive additional optimization pressure.

Across these studies, the central issue is therefore not adaptive sampling per se, but the proxy used to allocate supervision. Prompt instability, model--label agreement, uncertainty, and task-level error encode different notions of informativeness or difficulty and need not align with clinical value. Policies favoring model-consistent cases may reinforce existing blind spots, whereas policies dominated by uncertainty or error may concentrate effort on noise or irreducibly ambiguous cases. Neither approach creates missing population, scanner, or disease coverage unless such constraints are represented explicitly \citep{nath2020diminishing,faghani2023quantifying}.
 Active-learning studies should assess annotation effort and subgroup coverage rather than label count alone, while curriculum studies should test sensitivity to the difficulty metric and reweighting schedule. Current FM evidence supports particular adaptive allocation policies on retrospective benchmarks, but not the general assumption that model-defined difficulty or informativeness corresponds to clinical value.

\subsubsection{Weak and Noisy Supervision}
\label{subsec:weak_noisy}

Weak- and noisy-supervision DCA changes the source, precision, or reliability of the signal used to adapt a retained FM. Weak labels may provide only image-level classes, points, scribbles, or boxes. Semi-supervised methods derive training targets from unlabeled data and noisy-label methods attempt to limit the influence of incorrect observed labels. These settings have different error processes and should not be conflated. A sparse but correct point loses spatial information, whereas an incorrect dense mask supplies precise but erroneous supervision. The relevant question is therefore whether the weaker or noisier signal improves an otherwise comparable post-pretraining pathway, rather than simply whether fewer dense annotations are used.

Direct FM evidence shows that unlabeled data can materially strengthen adaptation when converted into pseudo-supervision. In the Divergent--Convergent Framework (DCF), MedSAM and MedSAM2 were adapted by fine-tuning the decoder, while unlabeled pixels were assigned different training roles based on the confidence and disagreement of the model \citep{wan2026dcf}. Under matched labeled budgets and adaptation pathways, DCF substantially improved segmentation over supervised-only adaptation across the reported MRI and CT benchmarks. The result supports semi-supervised FM adaptation, but also exposes its central limitation that pseudo-labels do not create independent ground truth. When predictions originate from related models or representations, systematic errors may be reproduced with high confidence, so additional unlabeled data can amplify useful structure alongside existing FM biases. Confidence and disagreement can regulate this process, but cannot guarantee that retained pseudo-labels are clinically correct.

Sparse supervision introduces a different information constraint. WSPolyp-SAM used bounding boxes to derive SAM pseudo-masks before adapting the prompt encoder and mask decoder, and the authors reported near-full-supervision performance in several comparisons using the same ViT backbone \citep{cai2024wspolypsam}. However, the boxes were derived from existing masks, and the reported 70--73\% saving in annotation-time was based on an assumed box-to-mask annotation-time ratio rather than prospective annotator measurement. Point-supervised nuclei segmentation similarly combined sparse points with SAM adaptation and self-refined pseudo-labels \citep{nam2026selfrefining}. It improved over InstaSAM in several settings, including controlled point displacement, but not universally. On CPM17 with precise points, Dice was 82.9\% versus 83.9\%. Together, these results show that weaker spatial labels can retain enough information for effective adaptation, but their missing information must be reconstructed by the model. Boxes constrain location without defining exact boundaries, while points identify instances without specifying extent. Any recovered structure therefore depends partly on the FM's prior rather than on observed annotation.

The mechanism-level risk is consequently an \emph{information bottleneck}. Weak supervision reduces annotation burden by discarding part of the target information and relying on model priors, pseudo-labels, or refinement procedures to reconstruct what is missing. This can be advantageous when the retained FM supplies an appropriate prior, but harmful when clinically important morphology lies outside that prior. Noisy supervision presents a complementary problem in which information is present but unreliable and filtering or correction mechanisms must decide which observations to trust. Such decisions can preferentially suppress atypical, low-contrast, minority, or difficult cases if disagreement with the model is treated as evidence of label error. Report-derived labels introduce analogous uncertainty through negation, temporality, omissions, and institution-specific reporting conventions \citep{Wang_2017_CVPR,irvin2019chexpert}.

Weak and noisy supervision therefore do not simply reduce annotation cost, but also redistribute where uncertainty enters adaptation. Pseudo-labels can propagate model blind spots, sparse labels leave clinically relevant structure unspecified, and confidence-based filtering can trade coverage for apparent label precision. Stronger evidence requires matched FM comparisons, clinically realistic weak or noisy labels, direct measurement of annotation and correction burden, and external clinical evaluation kept independent of pseudo-label generation or label-refinement decisions. Current FM evidence supports selected weak- and semi-supervised adaptation strategies, but not a general assumption that replacing expert supervision with incomplete or model-generated targets preserves clinically relevant information or external reliability.

\subsection{Architectural/Sequence Adaptation}\label{subsec:asa}
Where PLA, RLA, OLA, and DCA primarily modify parameter updating, task-facing representations, optimization objectives, or the data and supervision used for adaptation, ASA modifies the computational pathway through which a retained FM processes and integrates information. Post-pretraining components can introduce spatial, volumetric, multimodal, temporal, or state-based interactions, thereby embedding structural assumptions about which information should interact and over what range. ASA has no fixed update depth and may coexist with frozen-backbone adaptation, PEFT, partial updating, or full fine-tuning.

Within this review, an architectural intervention constitutes direct FM adaptation only when an eligible pretrained FM remains central and the structural change is introduced after pretraining. Structural properties already established during FM pretraining, such as native volumetric processing, do not themselves constitute post-pretraining ASA and are used only as contextual evidence for the relevant inductive bias. The following subsections examine three recurring forms of ASA: extending 2D representations across volumetric depth, integrating complementary modalities, and propagating information across ordered sequences or memory. Their value depends on whether the imposed assumptions about spatial continuity, cross-modal complementarity, or sequential dependence match the clinical information required by the task.

\subsubsection{2D--3D Adaptation}

Two-to-three-dimensional adaptation modifies how a retained 2D-pretrained FM incorporates through-plane information after pretraining, for example through depth-aware adapters, volumetric decoders, or learned slice fusion. Applying a 2D FM independently to individual slices does not itself constitute 3D adaptation. Slice-wise, 2.5D, converted-3D, and native-3D processing represent distinct structural choices rather than an ordinal progression in which greater volumetric context is necessarily preferable.

Med-SA retained the frozen SAM ViT-H backbone and introduced bottleneck adapters with Space--Depth Transpose for in-plane and through-plane interaction, updating approximately 13 million of 636 million parameters. On BTCV, its BBox-0.75 configuration achieved $89.8\pm0.3\%$ mean Dice compared with 86.9\% for Swin UNETR \citep{ref53_medsa}. This supports the complete parameter-efficient volumetric adaptation pathway, but does not isolate the contribution of depth exchange from adapters, prompting, decoder adaptation, or comparator differences.

More extensive or intermediate conversions show similarly configuration-dependent effects. 3DSAM-adapter introduced 3D patch embedding, depth positional information, spatial adapters, a volumetric prompt pathway, and a 3D decoder around SAM-B. With one point per volume, kidney-tumor Dice on KiTS21 was 73.78\% versus 73.07\% for nnU-Net, whereas liver-tumor Dice on LiTS17 was lower at 54.78\% versus 60.10\% \citep{ref130_3dsam_adapter}. MemSAM-2.5D instead processed three adjacent CT slices using z-axis state fusion and prototype memory, improving Dice from 76.8\% to 79.0\% on HCC-TACE-Seg and from 75.0\% to 77.1\% on the independent WAW-TACE cohort \citep{hou2026memsam25d}. These results support the evaluated volumetric pathways without establishing that greater depth integration is generally superior.

Native 3D FMs provide a distinct comparison context. Merlin is a volumetric CT foundation model pretrained using three-dimensional CT together with clinical text and structured clinical information, and subsequently adapted for downstream tasks including disease prediction and 3D semantic segmentation \citep{blankemeier2026merlin}. Unlike Med-SA, 3DSAM-adapter, and MemSAM-2.5D, Merlin does not convert a retained 2D representation into a volumetric one; volumetric representation learning is already incorporated during pretraining. It therefore illustrates native-3D FM reuse rather than evidence that native 3D pretraining outperforms 2D--3D adaptation.

The principal distinction is therefore between constructing through-plane context around a 2D representation and adapting a representation pretrained volumetrically. The former introduces dependence on slice spacing, orientation, resampling, and depth aggregation, whereas native 3D models introduce their own dependencies on volumetric pretraining coverage, voxel geometry, resolution, patching, and computational cost. Because dimensionality, architecture, pretraining data, supervision, and evaluation designs differ simultaneously across current studies, the evidence does not support a general ranking of slice-wise, 2.5D, converted-3D, or native-3D approaches. Evaluation should instead report acquisition geometry, resampling, input resolution, prompt conditions, computational requirements, and external performance under relevant acquisition variation.

\subsubsection{Multimodal and Cross-Modal Adaptation}

Multimodal ASA modifies how a retained FM routes, combines, or exchanges information across modalities after pretraining through modality-specific adapters, fusion modules, or cross-attention. This differs from constructing a new multimodal FM through joint pretraining. The auxiliary information may be required at inference, used optionally, or supplied only during training as privileged information. Training-only information constitutes multimodal ASA here only when it enters through an explicit post-pretraining cross-modal computational pathway rather than solely through labels or objective terms.

Direct FM evidence spans several forms of cross-modal integration. CoMed-SAM retained pretrained SAM components, introduced modality-specific LoRA adapters for different MRI contrasts, and combined their embeddings through a trainable local--global fusion module with modality-dropout training. Dice reached $0.9238\pm0.0286$ on IVDM3Seg and $0.9345\pm0.0476$ on CHAOS with the available modalities, while performance remained comparatively robust when one phase was absent \citep{kim2025comedsam}. CorrFABR used frozen DINOv2 CT and CONCH pathology representations to learn pathology-informed CT biomarkers, with pathology available during training but not required at inference. On the held-out KiTS21 test set, the complete model achieved ROC--AUC $0.855\pm0.005$, compared with 0.679 for a tumor-diameter baseline \citep{bhattacharya2025corrfabr}. 

Across these forms, the principal opportunity is access to information that may be complementary to the original visual representation. The corresponding mechanism-level risk is that fusion does not itself determine whether a modality is complementary, redundant, misleading, or spuriously predictive. When one stream is easier to exploit, it can dominate the fused representation and reduce reliance on clinically important evidence from another modality. Cross-modal interaction can also transfer shortcuts between streams, particularly when reports or structured variables encode reporting conventions, referral patterns, acquisition characteristics, or institutional practice rather than pathology \citep{irvin2019chexpert,zech2018variable}.

A further potential limitation is \emph{selective compression}. Cross-modal alignment encourages representations to preserve information that can be reconciled across modalities \citep{ref_112_convirt,ref125_gloria,ref145_mvfa}, but clinically useful modality-specific information need not be shared. Stronger fusion or alignment may therefore suppress discordant yet valid evidence. This is particularly relevant for visual--language adaptation, where the text representation defines which visual distinctions are rewarded.  Broad anchors such as normal and abnormal may support transfer while weakly representing lesion type, severity, extent, laterality, or uncertainty \citep{ref145_mvfa}.

Multimodal adaptation is also intrinsically dependent on modality availability and reliability. Missing-modality training can improve robustness to predefined absent-input patterns, but absence is not equivalent to corruption, misregistration, delay, or contradiction \citep{kim2025comedsam,wu2026deep}. When an auxiliary modality is available only during training, the deployed model observes a learned correlate of that information rather than the modality itself, as in privileged cross-modal adaptation \citep{bhattacharya2025corrfabr}. Fusion likewise provides no intrinsic guarantee that the clinically correct source will dominate when modalities disagree. The value of multimodal adaptation therefore depends on whether additional streams provide complementary and transportable information under realistic patterns of availability and disagreement. Current evidence supports multimodal FM adaptation in selected settings, but not the broader assumption that adding modalities, maximizing cross-modal agreement, or increasing fusion capacity is intrinsically beneficial.

\subsubsection{Sequence and State-Space Adaptation}

Sequence and state-space adaptation extends a retained FM beyond independent-image processing by propagating information across ordered frames, slices, patches, or prior observations. Implementations include temporal adapters, selective state-space blocks, and external memory, which differ in how prior information is retained and revisited. State-space formulations such as Mamba are particularly attractive for long sequences because their sequence-mixing operation scales linearly with sequence length rather than quadratically as in global self-attention \citep{gu2024mamba}. The common architectural aim is to make longer-range context available without abandoning transferable pretrained representations.

Direct FM evidence spans temporal, volumetric, and memory-based variants. MediViSTA retained SAM while adding temporal-fusion adapters, a frequency-feature pathway, and multiscale decoding for echocardiographic video segmentation. On CAMUS left-ventricular endocardium segmentation, it reported 94.42\% Dice compared with 93.22\% for Med-SA \citep{kim2025medivista}. ShapeMamba-EM retained SAM-Med3D and incorporated 3D Mamba adapters together with local shape information for volumetric electron-microscopy segmentation. On the Gauy axon benchmark, Dice reached 0.809 compared with 0.790 for U3D-BCD \citep{shi2024shapemamba}. SAMed-2 instead combined temporal adaptation with confidence-driven feature memory around SAM-2. In zero-shot evaluation across 10 datasets described as external to its medical pretraining data, mean Dice was 0.6938; task-specific U-Nets fine-tuned for the corresponding tasks achieved 0.6879 \citep{yan2025samed2}. These results support the evaluated sequence-aware FM systems, while only the within-pipeline ablations can attribute gains specifically to temporal, state-space, or memory components.

The principal mechanism-level assumption is that sequence order represents clinically meaningful context. This is natural for temporal imaging, but less direct for reconstructed volumes, where neighboring slice indices can correspond to different physical distances or orientations across acquisition protocols \citep{ref52_mim3d,ref130_3dsam_adapter}. Propagating information across an ordered sequence can improve continuity when neighboring observations are genuinely related, but may also smooth abrupt yet real changes or dilute localized evidence when the propagation scale exceeds that of the target \citep{ma2024umamba,chen2024msvmunet}. Longer context is therefore useful only when the ordering and spatial or temporal scale of propagation correspond to the clinical phenomenon being modeled.

State-space mechanisms introduce a potential \emph{selective-retention bottleneck}. Their computational efficiency arises from propagating a finite evolving state rather than explicitly constructing pairwise interactions among all sequence elements, while selective state updates determine how input information influences that state \citep{gu2024mamba}. Such compression cannot guarantee that the information retained by the learned dynamics is also the information of greatest clinical importance. Weak or infrequent findings may receive limited influence, whereas strongly represented but non-specific features can persist across subsequent observations. External memory extends the accessible context but retains a related selection problem because storage and retrieval depend on learned criteria such as confidence or feature similarity \citep{yan2025samed2}.
Sequence length and memory capacity should therefore not be treated as monotonic sources of benefit. Greater context can improve representation of extended structures and temporal evolution, while also incorporating irrelevant observations and increasing dependence on ordering and state-selection assumptions \citep{gu2024mamba,ma2024umamba}. Likewise, linear complexity of the state-space sequence-mixing operation does not imply linear end-to-end computational cost once high-resolution feature extraction, volumetric processing, multidirectional pathways, fusion, and decoding are included \citep{shi2024shapemamba,chen2024msvmunet}.

Evaluation should consequently compare sequence-aware adaptation with matched frame- or slice-wise processing and examine sensitivity to sequence length, order, spacing, missing observations, and state or memory reset. Performance should be considered alongside local lesion or boundary sensitivity and end-to-end computational cost rather than continuity alone. Current evidence supports sequence, state-space, and memory mechanisms when clinically relevant information extends beyond the current observation, but does not establish that longer context, smoother propagation, or larger memory is intrinsically more accurate or reliable.

\begingroup
\fontsize{8.5}{10.7}\selectfont
\renewcommand{\arraystretch}{1.15}
\setlength{\tabcolsep}{2pt}
\setlength{\LTleft}{0pt}
\setlength{\LTright}{0pt}
\Needspace{.35\textheight}
\begin{longtable}{@{}P{1.10cm}P{3.20cm}P{4.20cm}P{3.60cm}P{3.90cm}@{}}
\caption{Author-interpreted mechanism map for foundation-model adaptation in medical image analysis. Entries summarize plausible mechanisms, potential risks, and evaluation considerations synthesized from heterogeneous evidence; they are not causal estimates, comparative rankings, or strategy recommendations.}\label{tab:taxonomy_riskview}\\
\toprule

\textbf{Strategy} &
\textbf{Primary problem} &
\textbf{Plausible mechanism} &
\textbf{Potential risk} &
\textbf{Evaluation focus} \\

\midrule
\endfirsthead
\multicolumn{5}{@{}l}{\textit{Table \thetable\ (continued)}}\\[3pt]
\toprule

\textbf{Strategy} &
\textbf{Primary problem} &
\textbf{Plausible mechanism} &
\textbf{Potential risk} &
\textbf{Evaluation focus} \\

\midrule
\endhead
\endfoot
\endlastfoot

\textbf{PLA}
&
Parameter--task/domain mismatch
&
Full, partial, or restricted parameter updating
\citep{ref96_peft_mia,nouyed2026retinalfm}
&
Feature distortion or forgetting; residual mismatch with restricted updating
&
Same-checkpoint update-depth comparison; out-of-distribution performance
\\

\addlinespace

\textbf{RLA}
&
Task- or domain-misaligned representations
&
Feature transformation, projection, or post-pretraining realignment of the task-facing representation
\citep{lian2025alta,qiu2025effortless}
&
Loss of useful variation; reinforcement of confounded structure
&
Transport across sites, populations, and acquisition conditions
\\

\addlinespace

\textbf{OLA}
&
Objective--task behavior mismatch
&
Optimization for separability, invariance, consistency, or task-specific constraints
\citep{wang2026promptmed,wu2025sampopath}
&
Overemphasis of encoded objectives; attenuation of subtle or rare signals
&
Task gain together with objective-specific failure behavior
\\

\addlinespace

\textbf{DCA}
&
Limited, biased, or unreliable adaptation signal
&
Sampling, weighting, synthesis, weak supervision, or adaptive data selection
\citep{wan2026dcf,mahapatra2026valiant}
&
Coverage gaps; sampling bias; synthetic shortcuts; label error
&
Data coverage and provenance; annotation burden; independent real-data validation
\\

\addlinespace

\textbf{ASA}
&
Mismatch with spatial, multimodal, or sequential structure
&
Volumetric processing, modality fusion, sequence modeling, or memory pathways
\citep{ref53_medsa,kim2025comedsam,yan2025samed2}
&
Incorrect structural assumptions; dependence on geometry, modality availability, or sequence order
&
Realistic geometry, missing modalities, sequence variation, and acquisition shift
\\

\bottomrule

\end{longtable}

\vspace{2mm}

\footnotesize
\textit{Note.}
Entries are author-interpreted summaries synthesized from heterogeneous direct, indirect, and methodological evidence rather than comparative clinical effect estimates, causal risk estimates, or strategy recommendations. Citations are representative traceability anchors rather than exhaustive evidence for every statement in a row. FM denotes foundation model; PLA, parameter-level adaptation; RLA, representation-level adaptation; OLA, objective-level adaptation; DCA, data-centric adaptation; ASA, architectural/sequence-level adaptation.

\endgroup

% Section 5
\section{Task-Conditioned Analysis Across Medical Imaging Tasks}
\label{sec:task_comparative}

Aggregate benchmark scores can obscure \emph{task-dependent failure modes} in MIA, and rankings may vary with metric choice, test data, aggregation rules, and annotation conventions \citep{maier2018rankings}. More importantly, similar numerical changes can have different clinical consequences across tasks: miscalibration can alter classification decisions, boundary errors can distort segmented measurements, missed localized findings can delay detection, and unsupported language outputs can misstate image evidence \citep{kelly2019key,ref47_metrics}. These outcomes are not interchangeable; heterogeneous task endpoints are therefore not pooled into a common effect or used to construct a universal ranking of adaptation families.

The following subsections examine adaptation as a task-conditioned problem. Direct examples identify the retained pretrained FM, the post-pretraining intervention, and the reported task outcome. Critical interpretation is then synthesized separately at the task--mechanism level, focusing on the information or behavior an intervention emphasizes, the assumptions it introduces, and the failure patterns those assumptions may produce. Broader methodological literature is used to contextualize task risks and evaluation requirements rather than as direct evidence that a particular FM-adaptation strategy is superior.
The relative value of PLA, RLA, OLA, DCA, and ASA therefore remains conditional on the task, supervision, modality, data distribution, and intended use rather than being an intrinsic property of the adaptation family.
Table~\ref{tab:controlled_comparisons} therefore reports selected source-verifiable comparisons from the formally published comparative core and states the boundary of attribution supported by each contrast.

\subsection{Classification: Long-Tailed Structure, Decision Risk, and Calibration}
\label{subsec:task_classification}

Clinical image classification is a risk-sensitive decision problem rather than only a ranking task. Long-tailed class frequencies, multi-label co-occurrence, uncertain labels, and distribution shift interact with adaptation, while discrimination and decision reliability remain distinct: a model may rank cases well yet produce poorly calibrated probabilities or unsuitable operating thresholds \citep{zhang2023deep,irvin2019chexpert,guo2017calibration}.

Direct FM evidence illustrates several adaptation responses. TFA-LT retained frozen CLIP ViT-B/16 encoders and combined adapters trained under weighted sampling and undersampling. On APTOS-2019, total and few-class accuracy reached 75.60\% and 73.11\%, versus 59.47\% and 49.56\% for balanced contrastive learning; on ISIC2018, however, total accuracy was 70.48\% versus 71.19\% \citep{ref126_tfa-lt}. Thus, increasing tail-class influence can improve rare-class performance without guaranteeing higher aggregate performance across datasets. Dutt et al. found no universally preferred update depth: on BreastUS, CLIP ViT-B/16 with LoRA achieved F1 of 0.94 versus 0.91 for full fine-tuning, while relative rankings varied across datasets and training volumes \citep{ref96_peft_mia}. RETFound showed that full downstream fine-tuning can also transfer across datasets, achieving AUROC 0.943 in held-out internal evaluation on APTOS-2019 and 0.822 when fine-tuned on APTOS-2019 and evaluated on IDRiD \citep{zhou2023foundation}.

EAGLE extends this evidence toward clinical deployment. A Prov-GigaPath-initialized pathology encoder was fine-tuned end-to-end for slide-level EGFR mutation prediction, achieving AUC 0.847 internally, 0.870 across external centers, and 0.890 in a prospective real-time silent trial. Threshold-based workflow analysis indicated that rapid molecular testing could be reduced by up to 43\% while maintaining the prespecified clinical-standard performance \citep{campanella2025deployment}. This progression illustrates why benchmark discrimination alone is insufficient when classification outputs are intended to influence downstream decisions.

At the mechanism level, class-rebalancing approaches to long-tail adaptation change the effective contribution of classes during optimization but cannot create missing diversity within rare classes. Oversampling, undersampling, reweighting, or balanced ensembling can increase tail exposure, yet repeated presentation of a small tail set may amplify annotation errors or idiosyncratic features, while reduced head-class exposure can weaken estimation of common-class variability \citep{zhang2023deep,kang2019decoupling}. Tail performance is therefore both an optimization and a coverage problem.

Restricted and full updating introduce a related plasticity--capacity trade-off. PEFT can preserve pretrained structure and reduce trainable capacity when labels are scarce, but may leave substantial domain mismatch unresolved; full fine-tuning offers greater flexibility while allowing task- or site-specific signals to reshape the complete representation \citep{ref96_peft_mia}. Trainable-parameter count is consequently an efficiency property rather than evidence of lower clinical overfitting, better calibration, or superior rare-class sensitivity.

Calibration adds a distinct decision-level requirement. Reweighting, focal objectives, or contrastive training can alter score geometry without ensuring that confidence represents outcome probability \citep{guo2017calibration}. Prevalence shift changes predictive values, while acquisition or population shift can alter both ranking and calibration \citep{ma2022testtime}. Weak and multi-label supervision add further ambiguity because report-derived negatives may reflect non-mention rather than true absence, and disease co-occurrence can allow one finding to act as a proxy for another \citep{irvin2019chexpert}. Improved classification therefore need not imply reliance on the clinically intended image evidence.

Evaluation should consequently separate overall discrimination from rare-class sensitivity, calibration, and performance at clinically relevant operating thresholds, with external, subgroup, site, and temporal results reported separately where available. Current FM evidence shows that tail-aware adaptation and restricted or full parameter updating can each be effective in selected classification settings, while EAGLE further demonstrates that an adapted FM can retain useful discrimination when evaluated prospectively in a real-time clinical workflow. The evidence does not support a classification-wide preference for any single adaptation family. The central question is whether adaptation preserves rare-class information and maintains reliable decision behavior under the intended deployment distribution.

\subsection{Segmentation: Boundary Geometry, Spatial Coherence, and Interaction}
\label{subsec:task_segmentation}

Segmentation in MIA produces geometric outputs that directly inform downstream measurements and interventions, including tumor burden estimation, surgical planning, organ-at-risk delineation, and radiotherapy targeting. Unlike classification, segmentation errors are spatially structured, and small boundary deviations can propagate into meaningful differences in volume estimates, margin assessment, or treatment-field delineation \citep{OSTMEIER2023102927,muller2022towards}. A central deployment risk is therefore not semantic misclassification alone, but geometric instability.

In medical imaging, evidence shows substantial gains from downstream adaptation. MedSAM adapted SAM ViT-B for medical segmentation using more than 1.5 million image--mask pairs and perturbed bounding-box prompts. It reported median DSC 87.8\% on nasopharyngeal-cancer segmentation, while MedSAM-assisted editing reduced annotation time by approximately 82\% for two radiologists in an adrenal-tumor annotation experiment \citep{ref31_medsam}. Med-SA instead retained frozen SAM parameters and trained approximately 13 million adapter-related parameters with volumetric depth exchange and prompt conditioning. On BTCV, one-point mean Dice reached 88.3\% versus 54.8\% for unadapted SAM under the corresponding prompt condition. The BBox-0.75 configuration reached \(89.8\pm0.3\%\), compared with 86.9\% for the automatic Swin UNETR comparator; this latter contrast is not interaction-matched because Med-SA receives bounding-box information \citep{ref53_medsa}.

Volumetric adaptation provides complementary evidence. 3DSAM-adapter converted retained SAM-B representations to volumetric processing through 3D patch embedding, depth encoding, adapters, and a 3D decoder. On KiTS21 kidney-tumor segmentation, 3DSAM-adapter achieved 73.78\% Dice using one point per volume; the reported unadapted SAM-B baseline achieved 36.30\% using one point per slice, while automatic nnU-Net achieved 73.07\% \citep{ref130_3dsam_adapter}. These contrasts support the complete volumetric adaptation system but are not fully interaction-matched.
MA-SAM similarly combined low-rank adaptation with 3D adapters and decoder updating, reporting 87.2\% mean Dice on BTCV and stronger transfer from BTCV to AMOS22 MRI than nnU-Net in the reported zero- and five-shot settings \citep{ref127_masam}. Beyond SAM-based systems, PanDerm fine-tuned its retained dermatology encoder with a UPerNet decoder for ISIC2018 lesion segmentation and achieved DSC 0.910, compared with 0.904 for the evaluated MedSAM configuration \citep{yan2025multimodal}.

At the task-mechanism level, prompting changes how much target geometry must be inferred from the image. A point primarily identifies location, whereas a box additionally constrains approximate extent. More informative prompts can therefore improve overlap without improving autonomous target discovery. Prompt-conditioned adaptation also makes performance dependent on prompt precision, target multiplicity, and user interaction. For multifocal, poorly circumscribed, or unusually located disease, the supplied prompt may under-specify the intended target even when the underlying representation is strong \citep{ref21_sam,ref31_medsam}.

Volumetric adaptation introduces a related spatial trade-off. Through-plane information can improve anatomical continuity, but its usefulness depends on whether slice spacing, orientation, and receptive-field scale correspond to the anatomy being segmented. Resampling and depth mixing may improve coherence while blurring thin structures or diluting evidence confined to a small number of slices \citep{ref130_3dsam_adapter,ref127_masam}. Greater 3D context is therefore not intrinsically more geometrically faithful as its benefit depends on the relationship between acquisition geometry and target scale.

Dense supervision and objective choice further determine which geometric errors are emphasized. Region-based objectives may underweight small lesions or thin structures, whereas boundary- or surface-aware losses emphasize contours but inherit uncertainty in the reference annotation \citep{kervadec19a,karimi2019reducing}. Shape or topology constraints can discourage implausible masks, but overly rigid priors may suppress genuine pathological disruption. No adaptation mechanism can eliminate reference ambiguity when the target boundary is itself inconsistently defined across experts or institutions \citep{jungo2018effect}.

Segmentation evaluation should therefore separate overlap from boundary and lesion-level reliability, with size-stratified and external results reported where available \citep{ref47_metrics,muller2022towards}. Promptable systems additionally require standardized prompt conditions and interaction burden because performance depends jointly on the image, prompt, and correction pathway. Current evidence shows that FM adaptation can improve overlap and annotation efficiency and can enable effective volumetric processing in selected settings, but does not establish that any adaptation locus consistently produces geometrically reliable contours across tasks.

\subsection{Detection and Localization: Rarity, Sensitivity, and Missed Findings}
\label{subsec:task_detection}

Detection and localization differ from classification because clinically relevant errors are often lesion-level as a case may be correctly labeled abnormal while a small or consequential finding remains missed or poorly localized. Extreme foreground--background imbalance, small targets, incomplete lesion annotation, hard negatives, and false-positive workload therefore shape adaptation and evaluation \citep{lin2017focal,yan2018deeplesion,setio2017validation}. Case-level discrimination and spatial localization should consequently be considered separately.
Direct FM evidence shows that adaptation can improve both candidate detection and anomaly localization. Qin et al.\ adapted pretrained GLIP-T across thirteen medical object-detection datasets using medical prompts and partial parameter updating. On LUNA16 pulmonary-nodule detection, AP increased from 31.9\% when the visual backbone was fully frozen to 40.0\% under the authors' partial-updating configuration, while AP50 increased from 78.8\% to 84.7\% \citep{qin2023medical_vlm}. These results show that both visual adaptation and target specification can materially alter detection performance.

Anomaly-localization studies provide complementary evidence. MediCLIP retained a CLIP visual backbone and learned normal/abnormal prompts with multiscale visual adapters, using synthetically generated abnormalities to provide spatial supervision. With four CheXpert support images, image-level AUROC reached $70.1\pm1.6\%$, compared with $66.0\pm1.9\%$ for SQUID \citep{ref146_mediclip}. MVFA similarly retained frozen CLIP encoders and adapted multilevel visual features toward generic normal and abnormal language anchors. Under leave-one-dataset-out evaluation, mean anomaly-classification AUC improved by 6.24 percentage points and pixel-level localization AUC by 2.03 points relative to April-GAN \citep{ref145_mvfa}.

At the task-mechanism level, detection depends critically on how the target search space is specified. Text prompts can supply semantic and anatomical priors that compensate for weak correspondence between pretrained concepts and medical findings, but richer specification also narrows what the model is encouraged to search for. Prompts describing typical location, morphology, or appearance may therefore improve detection of canonical presentations while underrepresenting atypical disease. Increased prompt informativeness should not be interpreted as equivalent to increased autonomous visual sensitivity \citep{qin2023medical_vlm}.

Anomaly adaptation imposes a related definition problem. Mapping heterogeneous pathology onto generic normal and abnormal concepts can support transfer across diseases and organs, but it compresses distinctions among benign variation, artifacts, treatment effects, and clinically urgent abnormalities. Synthetic supervision further makes the abnormality distribution dependent on the transformations used to generate lesions or defects. Visually diverse synthetic perturbations do not guarantee coverage of the morphology or context of rare clinical disease \citep{ref146_mediclip}. Multiscale adaptation addresses another detection-specific trade-off: fine-resolution features preserve small targets but can respond strongly to normal edges and acquisition noise, whereas coarser semantic features provide context at the risk of averaging away small or subtle findings \citep{ref145_mvfa}.

Rarity amplifies each of these limitations. Reweighting, oversampling, or hard-example emphasis can increase the contribution of uncommon positives, but cannot create missing variation within the rare class and may magnify annotation errors or repeated site-specific patterns \citep{lin2017focal}. Incompletely annotated datasets create an additional ambiguity because apparently false-positive detections may correspond to genuine but unlabeled abnormalities \citep{yan2018deeplesion}. Detection adaptation must therefore balance sensitivity to weak, local signals against suppression of the large number of plausible normal mimics.

Evaluation should reflect this lesion-level risk. Case-level discrimination metrics, where relevant, and object-detection AP should be complemented by lesion-level sensitivity, false-positive burden or FROC analysis, and stratification by lesion size and clinically relevant rarity where spatial annotations permit \citep{setio2017validation,ehteshami2017diagnostic,ref48_detection_metrics}. External and site-specific evaluation is particularly important because pooled metrics can conceal selective loss of sensitivity under acquisition or prevalence shift. Current FM evidence shows that prompt design, constrained updating, synthetic anomaly supervision, and multiscale alignment can improve selected detection and localization endpoints, but does not establish that higher average scores preserve rare-finding sensitivity at an acceptable false-positive burden across deployment settings.

\begingroup
\fontsize{8.2}{10.3}\selectfont
\renewcommand{\arraystretch}{1.15}
\setlength{\tabcolsep}{2pt}
\setlength{\LTleft}{0pt}
\setlength{\LTright}{0pt}
\Needspace{.5\textheight}
\begin{longtable}{@{}P{1.50cm}P{2.80cm}P{2.65cm}P{2.95cm}P{2.40cm}P{3.60cm}@{}}
\caption{Selected controlled comparisons of post-pretraining foundation-model adaptation from the formally published primary comparative synthesis. Results are study-specific and do not imply cross-study ranking.}\label{tab:controlled_comparisons}\\
\toprule
\textbf{Study} &
\textbf{FM, task, modality} &
\textbf{Adaptation vs comparator} &
\textbf{Controlled aspect} &
\textbf{Reported result} &
\textbf{Attribution supported} \\
\midrule
\endfirsthead
\multicolumn{6}{@{}l}{\textit{Table \thetable\ (continued)}}\\[3pt]
\toprule
\textbf{Study} &
\textbf{FM, task, modality} &
\textbf{Adaptation vs comparator} &
\textbf{Controlled aspect} &
\textbf{Reported result} &
\textbf{Attribution supported} \\
\midrule
\endhead
\endfoot
\endlastfoot

\citep{ref96_peft_mia}
&
CLIP ViT-Base; breast-ultrasound classification
&
LoRA vs full fine-tuning
&
Same pretrained model, task, protocol, and F1 endpoint; update regime differed
&
BreastUS F1: 0.94 vs 0.91
&
Supports LoRA vs full updating for this configuration; not a general PEFT advantage
\\

\addlinespace

 \citep{nouyed2026retinalfm}
&
RETFound ViT-L/16; retinal classification
&
Frozen embeddings + logistic regression vs partial fine-tuning
&
Same checkpoint and ten-fold IDRiD protocol; update depth differed
&
IDRiD ROC--AUC: $93.3\pm2.7\%$ vs $86.9\pm5.9\%$; DeepDRiD: 79.9\% vs 59.5\%
&
Supports frozen-interface vs partial updating for this RETFound configuration; not a general preference for freezing
\\

\addlinespace

 \citep{wan2026dcf}
&
MedSAM/MedSAM2; MRI and CT segmentation
&
DCF semi-supervised adaptation vs supervised-only decoder tuning
&
Same starting FM, 10\% label budget, frozen encoders, trainable decoder, and test set
&
ACDC Dice: 89.583\% vs 77.023\%
&
Supports the complete DCF pathway; individual routing, prototype, disagreement, and consistency effects not isolated
\\

\addlinespace

 \citep{mahapatra2026valiant}
&
MedCLIP; chest-radiograph classification
&
Prompt-instability active learning vs Coreset
&
Same frozen FM, CheXpert split, 50\% label budget, batch size, seeds, and protocol
&
AUROC: 92.6\% vs 84.9\% across ten runs
&
Supports the combined prompt/adaptive-acquisition pathway; individual prompt-instability effects not isolated
\\

\addlinespace

 \citep{gao2026medpeftcl}
&
CLIPSeg; sequential multi-dataset segmentation
&
Retention-oriented PEFT pathway vs PEFT baseline
&
Same starting pathway and continual-learning task sequences
&
Mean Dice: 78.84\% vs 61.31\%; forgetting: 1.91\% vs 20.32\%
&
Supports the complete retention-oriented pathway for the evaluated task order; component-specific and cross-setting effects not established
\\

\bottomrule
\end{longtable}

\vspace{2pt}

\scriptsize
\textit{Note.}
Selected formally published comparisons met the prespecified comparative criterion and were not selected by effect size. ``Controlled'' denotes a within-study contrast rather than causal identification. FM, foundation model; CLIP, Contrastive Language--Image Pre-training; ViT, Vision Transformer; LoRA, Low-Rank Adaptation; PEFT, parameter-efficient fine-tuning; MRI, magnetic resonance imaging; CT, computed tomography; ROC--AUC/AUROC, area under the receiver operating characteristic curve; DCF, Divergent--Convergent Framework. BreastUS, IDRiD, DeepDRiD, ACDC, and CheXpert denote the datasets or benchmarks named in the source studies. Complete records are provided in Supplementary Data S18--S22.

\endgroup

\section{Failure Mechanisms and Reliability Risks}
\label{sec:failure_modes}

Post-pretraining adaptation can introduce or amplify failures that affect the reliability and transferability of foundation models in medical image analysis. These include negative transfer, shortcut reliance, data leakage, and multimodal semantic failure. Such risks may arise from target-specific updating, invariance constraints, data exposure, structural modifications, or multimodal supervision.
This section consolidates these recurring failure mechanisms and reliability risks, focusing on their underlying assumptions, observable effects, and validation implications. Some represent direct adaptation failures, whereas others primarily threaten the validity or reliability of evaluation. Table~\ref{tab:failure_mechanisms} summarizes the principal failure mechanisms, representative evidence, mechanism-level trade-offs, and corresponding validation probes.

\subsection{Over-Adaptation, Catastrophic Forgetting, and Representational Drift}
\label{subsec:overadaptation}

Over-adaptation is one form of negative transfer in which specialization to the adaptation target degrades capability relevant elsewhere in the intended-use domain. It should be distinguished from \emph{catastrophic forgetting}, which requires loss of a previously learned capability after sequential updating, and from \emph{representational drift}, which refers to measurable change in pretrained feature geometry or content. General transfer-learning evidence shows that extensive fine-tuning can distort pretrained representations and that retaining aspects of the pretrained solution can improve robustness under distribution shift \citep{ref64_ft_distor_feat,wortsman2022robust}. However, neither deeper updating nor external performance degradation alone establishes forgetting or harmful drift.

Direct FM evidence shows that retention and update depth affect previously acquired capability, but not monotonically. MedPEFT-CL retained CLIPSeg and combined parameter-efficient updates with replay and consolidation across sequential segmentation tasks, increasing average Dice from 61.31\% to 78.84\% and reducing average forgetting from 20.32\% to 1.91\% \citep{gao2026medpeftcl}. In the preprint sensitivity layer, MedSeqFT combined replay and distillation during sequential adaptation of VoCo and UniMiSS+, improving mean Dice and HD95 over conventional full fine-tuning. However, its small reported parameter changes do not establish preservation of feature geometry \citep{ye2025medseqft}. Adaptation depth was likewise not uniformly beneficial as partial fine-tuning of RETFound reduced ROC--AUC relative to a frozen probe on IDRiD ($86.9\pm5.9\%$ vs.\ $93.3\pm2.7\%$) and external DeepDRiD (59.5\% vs.\ 79.9\%) \citep{nouyed2026retinalfm}, while MeLo outperformed full fine-tuning on some endpoints but not others \citep{ref136_melo}.

At the mechanism level, adaptation creates a \emph{plasticity--retention trade-off}. Target-domain gradients can reorganize features toward distinctions useful for current supervision while weakening information needed for other tasks, sites, modalities, or disease patterns. The severity of this trade-off depends on where and how the representation is changed, the diversity and noise of the adaptation data, and the relationship between source and target domains. Full updating permits broader correction of genuine domain mismatch but exposes more of the representation to target-specific signals. PEFT restricts the update surface but can preserve an unsuitable backbone or still encode substantial behavioral change through relatively few parameters.

Retention mechanisms shift rather than eliminate this trade-off. Replay preserves only the variation represented by retained examples, distillation can preserve teacher errors and omissions alongside useful behavior, and consolidation constrains adaptation according to estimated parameter importance that may change across tasks or domains. Stronger anchoring can therefore protect prior capability while limiting correction of genuine domain mismatch. Broader medical-AI external-validation evidence establish distribution shift as an important failure context \citep{yu2022external,zech2018variable,guo2026impact}, but external degradation alone should not be attributed to over-adaptation or forgetting without an appropriate matched comparison.

Evaluation should therefore match the claim being made. Catastrophic forgetting requires before--after measurement of previously learned capabilities during sequential adaptation. Representational drift requires direct feature-level analysis rather than weight displacement alone, and over-adaptation requires evidence of target specialization accompanied by deterioration of a capability relevant to intended use. Current evidence supports retention-aware adaptation in selected settings, but does not establish that shallower updating, smaller parameter movement, or stronger anchoring is intrinsically safer or more transferable.

\subsection{Shortcut Learning and Spurious Correlations}
\label{subsec:shortcut_learning}

Shortcut learning occurs when a model relies on an accessible predictive cue that does not represent the intended clinical relation and fails when that cue is removed, reversed, or transported \citep{geirhos2020shortcut}. This should be distinguished from a \emph{spurious correlation}, which may exist in the data without being used by the predictor. Likewise, scanner, hospital, or demographic information being decodable from an FM representation does not itself demonstrate shortcut reliance. Stronger attribution requires controlled cue manipulation, disruption of the cue--label association, or comparison of the same biological content under different acquisition conditions \citep{banerjee2023shortcuts,degrave2021ai}.

Direct FM evidence shows that acquisition-associated structure can be substantial and modifiable. PathoROB evaluated 20 pathology FMs on 99,392 patches spanning 28 biological classes and 34 medical centers and found that several representations were organized more strongly by center than by biology; controlled center--label confounding linked increasing access to technical cues with downstream errors \citep{komen2026pathorob}. Zhang similarly used an adversarial projection around frozen pathology-FM features: for UNI, hospital-source AUC decreased from 0.948 to 0.579 while disease AUC remained high, changing from 0.982 to 0.961 \citep{zhang2026hospitalbias}. These findings show that domain information can be suppressed while retaining substantial task information, but they do not imply that all clinically irrelevant variation has been removed.

Scanner-controlled evidence strengthens attribution by holding biological content fixed. ScanGen aligned representations of the same specimens digitized on different scanners; for Phikon-based EGFR prediction, ROC--AUC increased from 0.738 to 0.768 while across-scanner prediction variability decreased from 0.389 to 0.121 \citep{carloni2025scangen}. Conversely, deeper updating is not necessarily more shortcut-prone: in the preprint sensitivity layer, Filiot et al.\ fully fine-tuned pathology FMs with an acquisition-robustness objective, increasing the mean PathoROB robustness index from approximately 0.72 to 0.87 while generally retaining or improving downstream performance \citep{filiot2026robustifying}. Shortcut susceptibility therefore depends on what adaptation encourages the model to preserve or suppress, not simply on update depth.

At the mechanism level, the central difficulty is that nuisance and clinically relevant information are not always separable. Scanner, stain, hospital, protocol, demographic, or workflow variables may correlate with artifacts but also with prevalence, disease severity, lesion visibility, referral pathways, or tissue processing. Enforcing invariance can therefore remove a genuine shortcut, suppress legitimate effect modification, or do both. Adversarial suppression is limited by the discriminator used, paired alignment by the variation represented in the pairs, and normalization or robustification by the nuisance factors represented during development \citep{komen2026pathorob,zhang2026hospitalbias,carloni2025scangen}. Stronger invariance is therefore not intrinsically safer.

The same attribution discipline applies beyond pathology. Chest-radiography FMs can encode demographic structure and exhibit subgroup performance differences \citep{glocker2023risk}, while hospital- and acquisition-dependent degradation has been documented in radiology \citep{zech2018variable,ong2024shortcut}; neither observation alone identifies the cue used for prediction. Multimodal adaptation can introduce analogous shortcuts through report style, disease co-occurrence, or metadata \citep{ref22_clip,ref32_medclip}. Evaluation should therefore distinguish \emph{cue encoding} from \emph{cue reliance} by manipulating suspected nuisance factors while preserving clinical content and examining prediction changes across disease--nuisance strata and external shifts. Current evidence shows that FM adaptation can reduce acquisition-associated dependencies, but not that removing all domain information or maximizing invariance is universally beneficial.

\subsection{Data Leakage and Implicit Test Contamination}
\label{subsec:domain_leakage}

Data leakage in FM adaptation can occur across several distinct boundaries. The same patient, examination, slide, or derived image may appear in both development and test partitions. Evaluation content may have entered the FM pretraining corpus. Target-domain data may influence post-pretraining adaptation before being presented as an independent test. Repeated use of a public benchmark may also produce adaptive overfitting at the research-process level. These mechanisms should not be conflated. Patient-level splitting does not establish pretraining-corpus independence, and an institution-external cohort is not adaptation-unseen if its unlabeled data informed continued training. Broad and incompletely documented FM corpora make such lineage particularly difficult to reconstruct \citep{longpre2024data}.

Direct FM studies illustrate protection against different leakage routes. PanDerm used a cleaned HAM10000 split designed to prevent lesions from the same patient crossing training, validation, and test partitions. Fully fine-tuned PanDerm reported 98.03\% accuracy versus 97.40\% for linear probing \citep{yan2025multimodal}. Prov-GigaPath evaluated downstream adaptation on a 403-patient colorectal cohort collected after March 2023, whereas its stated FM pretraining data preceded March 2023, providing temporal separation from the pretraining corpus \citep{xu2024provgigapath}. EAGLE provides a complementary post-development example. An adapted Prov-GigaPath pathway was evaluated in a prospective real-time silent trial of 197 primary-tumor samples and achieved AUC 0.890 using thresholds selected in an earlier cohort \citep{campanella2025deployment}. These designs address different forms of independence, including downstream patient separation, temporal separation from pretraining, and prospective post-development evaluation. They should not be treated as equivalent safeguards.

The problem extends beyond exact duplicate records. Slice-level splitting can place images from the same volume in both development and evaluation sets, while pathology tiles can cross partitions when splitting occurs after patch extraction. Both can substantially inflate performance relative to patient-level separation \citep{yagis2021data_leakage,tampu2022inflation}. Exact hashes may also miss resized, cropped, compressed, re-annotated, or otherwise derived versions of the same source image, whereas similarity-based matching can falsely identify distinct images with common anatomy. Image--text FMs introduce further linkage through reports, identifiers, timestamps, or outcome-related language. Public medical-imaging resources often lack sufficiently complete provenance and persistent identifiers to resolve these relationships reliably \citep{jimenez2024copycats}. Unresolved overlap should therefore be reported as provenance uncertainty rather than automatically labeled proven contamination.

Exposure to the target distribution is not inherently invalid. Continued self-supervision, transductive adaptation, test-time adaptation, or use of unlabeled target-center images may be legitimate strategies when reported transparently. The interpretive problem arises when performance after such exposure is presented as independent external generalization. Conversely, a geographically or temporally separate cohort may still share scanners, preprocessing, referral pathways, or institutional conventions. External validation, prospective evaluation, pretraining-disjoint evaluation, and adaptation-unseen evaluation therefore describe different properties and should be reported separately. Likewise, the possibility that a public benchmark was accessible during FM development does not establish contamination or performance inflation without evidence linking the evaluated content to model development \citep{maier2018rankings}.

Evaluation should therefore emphasize data lineage rather than rely on a single external-validation label. Pretraining, intermediate training, adaptation, model selection, and final testing should be documented separately, with splitting performed at the highest relevant dependency level before slice, tile, crop, or text-pair generation. Exact and near-duplicate screening should include linked derivatives where feasible, and unresolved provenance should be reported explicitly. When target-domain data inform adaptation, the analysis should be described as transductive, semi-supervised, domain-adaptive, or test-time adaptive rather than independent external validation. The strongest evidence comes from locked models, prompts, thresholds, and analysis plans evaluated on geographically or temporally disjoint data with no known pretraining or adaptation exposure.

\subsection{Multimodal Hallucination and Semantic Misalignment}
\label{subsec:multimodal_hallucination}

Multimodal hallucination, omission, and semantic misalignment are related but distinct failures. Hallucination occurs when a model asserts a clinical finding that is unsupported by the available image evidence, whereas omission refers to failure to report a finding that is visually supported. Semantic misalignment occurs when an image or region is associated with an incorrect textual concept even when no free-text report is generated. Report-derived supervision introduces a further source of error because reports may contain uncertainty, temporal comparisons, clinical context, and omissions that do not map directly to visible pathology \citep{irvin2019chexpert,huang2025survey,kim2025medical}. Adaptation can modify these pathways through report supervision, instruction tuning, reinforcement learning, synthetic question--answer data, retrieval, or explicit spatial grounding.

Direct FM evidence shows that adaptation can improve report-level agreement without necessarily establishing image-grounded factuality. MedGemma was fully fine-tuned for MIMIC-CXR report generation, increasing RadGraph F1 from 0.295 to 0.303 \citep{sellergren2025medgemma}. In the preprint sensitivity layer, DiagCoT adapted Qwen2-VL-7B-Instruct through image--report alignment, chain-of-thought supervised tuning, and reinforcement optimization. On MIMIC-CXR, BLEU-1 increased from 0.2096 with supervised chain-of-thought tuning alone to 0.3260 for the complete pathway \citep{luo2025diagcot}. These results show that adaptation can improve structured relation recovery and reference-text agreement, but RadGraph and BLEU do not directly measure whether every generated clinical claim is supported by the image.

Grounding provides a more direct connection between language and visual evidence. CURE LoRA-adapted MedGemma-4B-IT across phrase grounding and grounded report-generation tasks. Anatomy-grounded-report IoU reached $0.601\pm0.008$ compared with $0.249\pm0.008$ for MAIRA-2, while its anatomy-specific evaluation reported an abnormality hallucination rate of 8.78\% compared with 26.50\% \citep{messina2026cure}. Synthetic language supervision provides a different pathway. Zhang et al.\ adapted SmolVLM using generated dermatology question--answer pairs and figure captions, reaching 70.20\% classification accuracy with combined supervision compared with 68.63\% for synthetic QA alone \citep{zhang2026smolvlmsynthetic}. These examples show that grounding and synthetic supervision can improve their respective reported outcomes, but localization accuracy, answer correctness, classification performance, and report factuality remain distinct properties.

At the mechanism level, the central problem is that linguistic agreement is not equivalent to visual entailment. Autoregressive objectives reward likely report sequences and may reproduce frequent statements while underrepresenting rare or weakly documented findings. Chain-of-thought or reinforcement objectives can improve coherence while preserving an incorrect visual premise if the supervision or reward also accepts it. Spatial grounding narrows the link between a phrase and image evidence, but correct localization does not guarantee correct laterality, severity, temporality, certainty, or diagnosis. Synthetic supervision can reproduce the terminology, omissions, and reasoning patterns of the generating teacher, while report-derived objectives can transfer institutional phrasing or clinical context that is not visually observable \citep{irvin2019chexpert,jain2021visualchexbert}. Stronger multimodal alignment can therefore improve semantic consistency without ensuring that the adapted model relies on the clinically intended image evidence.

Evaluation should consequently separate factual support, omission, grounding, and linguistic similarity rather than combining them into a single multimodal score. Generated outputs require claim-level assessment of unsupported and missed findings, while grounded systems should test whether the cited region actually supports the associated clinical statement. Prompt variation and removal of image or textual context can further reveal whether responses depend on visual evidence or language priors. Current evidence shows that post-pretraining adaptation can improve report agreement, spatial grounding, and selected hallucination-related outcomes, but does not establish that greater language fluency, stronger semantic alignment, or higher reference-text similarity necessarily produces more image-grounded or clinically reliable outputs.

\begingroup
\fontsize{8.3}{10.4}\selectfont
\renewcommand{\arraystretch}{1.15}
\setlength{\tabcolsep}{2pt}
\setlength{\LTleft}{0pt}
\setlength{\LTright}{0pt}
\Needspace{.35\textheight}
\begin{longtable}{@{}P{2.35cm}P{4.40cm}P{3.05cm}P{3.25cm}P{2.95cm}@{}}
\caption{Failure-mechanism map for foundation-model (FM) adaptation in medical image analysis. Representative evidence is illustrative, while mechanism-level trade-offs and validation probes are author-synthesized.}\label{tab:failure_mechanisms}\\
\toprule
\textbf{Failure mechanism} &
\textbf{Representative direct FM evidence} &
\textbf{Key mechanism trade-off} &
\textbf{Validation probe} &
\textbf{Potential consequence} \\
\midrule
\endfirsthead
\multicolumn{5}{@{}l}{\textit{Table \thetable\ (continued)}}\\[3pt]
\toprule
\textbf{Failure mechanism} &
\textbf{Representative direct FM evidence} &
\textbf{Key mechanism trade-off} &
\textbf{Validation probe} &
\textbf{Potential consequence} \\
\midrule
\endhead
\endfoot
\endlastfoot

Over-adaptation, catastrophic forgetting, and representational drift
&
MedPEFT-CL reduced forgetting from 20.32\% to 1.91\% \citep{gao2026medpeftcl}. MedSeqFT improved VoCo Dice from 65.3\% to 68.3\% \citep{ye2025medseqft}.
&
Greater plasticity can improve current-task fit while weakening previously useful information.
&
Before--after retained-task testing, matched update-depth comparisons, and direct feature-similarity analysis.
&
Loss of prior capability or unstable performance across tasks, sites, or time.
\\

\addlinespace

Shortcut learning and spurious correlations
&
ScanGen increased EGFR AUC from 0.738 to 0.768 while reducing scanner variability from 0.389 to 0.121 \citep{carloni2025scangen}. Hospital-adversarial UNI adaptation reduced hospital AUC from 0.948 to 0.579 \citep{zhang2026hospitalbias}.
&
Removing nuisance information may also suppress clinically relevant domain-dependent information.
&
Separate cue encoding from cue reliance using controlled cue interventions, paired rescans, and external-site testing.
&
Transport failure, subgroup error, or reliance on non-clinical cues.
\\

\addlinespace

Data leakage and implicit test contamination
&
PanDerm used patient-separated HAM10000 splits \citep{yan2025multimodal}. Prov-GigaPath used a post-pretraining-cutoff cohort \citep{xu2024provgigapath}, while EAGLE added prospective silent evaluation \citep{campanella2025deployment}.
&
Different forms of data separation protect different stages of the development pipeline.
&
Audit data lineage, split at the highest dependency level, screen duplicates, and report pretraining and adaptation exposure separately.
&
Inflated or non-independent estimates of clinical generalization.
\\

\addlinespace

Multimodal hallucination and semantic misalignment
&
MedGemma increased RadGraph F1 from 0.295 to 0.303 \citep{sellergren2025medgemma}. CURE reported an abnormality hallucination rate of 8.78\% versus 26.50\% for MAIRA-2 \citep{messina2026cure}.
&
Better fluency, text agreement, or grounding does not guarantee visual entailment.
&
Separate false assertions, omissions, grounding, and text similarity using claim-level and image-grounded evaluation.
&
Unsupported or omitted findings and misleading clinical outputs.
\\

\bottomrule
\end{longtable}

\vspace{2pt}
\scriptsize
\textit{Note.} Evidence examples illustrate measured outcomes and do not rank failure mechanisms. Mechanism trade-offs and validation probes are author-interpreted synthesis. AUC: area under the receiver operating characteristic curve; EGFR: epidermal growth factor receptor.

\endgroup

\section{From Adaptation to Clinical Readiness}
\label{sec:clinical_readiness}
Technical improvement after FM adaptation does not by itself establish clinical readiness. Robustness, regulatory authorization, and clinical readiness are related but distinct claims. Clinical readiness additionally depends on the intended use, performance under relevant external and temporal shifts, calibration and subgroup reliability, prospective and workflow evaluation, human factors, equity, cybersecurity, and post-deployment monitoring \citep{kelly2019key,raji2020closing}.

This section therefore examines adaptation in relation to deployment constraints, validation and regulatory requirements, and the extent to which current FM evidence has progressed beyond development-set evaluation. Direct findings from the reviewed corpus are distinguished from broader methodological, clinical-evaluation, and regulatory evidence. The aim is to characterize what different forms of evidence can support, rather than to assign adaptation mechanisms or update regimes a fixed level of clinical readiness.

\subsection{Deployment Scenarios and Adaptation Constraints}
\label{subsec:deployment_scenarios}
Deployment context constrains the evidence required for adaptation rather than determining a preferred mechanism. Relevant factors include labeled and unlabeled data availability, mismatch between the pretrained FM and target modality, expected site and temporal shift, task requirements, error consequences, compute constraints, and the ability to govern the complete system. Intervention locus and update depth are treated as separate dimensions and may vary independently. Current comparative clinical evidence is insufficient to assign any of these combinations a fixed level of safety, readiness, or validation burden. Broader evidence on feature distortion, shortcuts, and dataset shift instead motivates testing these possibilities directly \citep{ref64_ft_distor_feat,geirhos2020shortcut,finlayson2021dataset_shift}.

Direct FM comparisons illustrate why data volume or trainable-parameter count alone cannot determine the appropriate update depth. Dutt et al.\ similarly found no universally preferred adaptation strategy: on BreastUS, a CLIP-pretrained ViT-Base with LoRA achieved F1 of 0.94 versus 0.91 for full fine-tuning, while relative rankings varied across datasets, model sizes, and training-data regimes \citep{ref96_peft_mia}. In an IDRiD retinal comparison, partial RETFound fine-tuning produced ROC--AUC $86.9\pm5.9\%$ versus $93.3\pm2.7\%$ for a frozen probe on IDRiD, with a larger difference on external DeepDRiD at 59.5\% versus 79.9\% \citep{nouyed2026retinalfm}. These findings support a capacity--preservation trade-off rather than a general preference for freezing or PEFT. Self-supervised target-domain adaptation offers another option when labels are scarce, but exposure to local unlabeled data may itself encode acquisition-specific structure and should be reported explicitly \citep{huang2023ssl_medical}.

Multi-institutional deployment introduces a related trade-off between fitting observed sites and transporting to unseen ones. Vardi et al.\ used LoRA with local discrepancy minimization to adapt UNI across hospitals and reported stronger gains when target-hospital information was available than under generalization to unseen hospitals \citep{vardi2026discrepancy}. In contrast, EAGLE used end-to-end fine-tuning of a Prov-GigaPath pathway and achieved AUC 0.870 across external centers and 0.890 in a prospective silent trial of 197 primary-tumor samples \citep{campanella2025deployment}. Together, these examples show that neither site-specific alignment nor deep updating has a deployment-wide advantage. Evaluation should instead determine whether adaptation preserves performance across sites, time, subgroups, calibration, and the intended workflow.

Modularity should likewise be interpreted as an engineering and governance property rather than evidence of clinical robustness. Adapters, prompts, and low-rank updates isolate a smaller trainable parameter set and may simplify storage, versioning, comparison, or rollback \citep{ref68_peft_review,ref95_peft_survey,ref100_lora}, but the validated system also includes the base checkpoint, preprocessing, prompts, retrieval resources, calibration, thresholds, software dependencies, and human workflow. Regulatory frameworks similarly focus on intended use, risk, validation, lifecycle control, and the impact of modifications rather than the fraction of parameters updated \citep{fda2025pccp,tga2026software,tga2025aievidence}. Adaptation depth and modularity should therefore be treated as design variables whose value depends on the deployment context and the evidence available to support the resulting system.

\subsection{Validation and Regulatory Considerations}
\label{subsec:validation_regulatory}

Validation evidence is multidimensional rather than a binary internal versus external distinction. Evaluation on a held-out split, another public dataset, an institution-external cohort, a temporally separated or prospective cohort, a silent clinical workflow, and an interventional study answer different questions about generalization and clinical utility. Independence from downstream training also does not establish independence from FM pretraining or target-domain adaptation. Among the 705 mapped studies, 173 (24.5\%) had confirmed independent external validation and only 10 (1.4\%) reported calibration or uncertainty evidence. None met the prespecified high-evidence-strength criteria. These figures characterize the evidence base rather than imply clinical failure among studies lacking these assessments.

For intended-use claims, validation should specify the target population, setting, user, input acquisition, output, operating threshold, comparator workflow, and anticipated clinical action \citep{tejani2024claim,ref_decide_ai2022}. Discrimination or overlap should be complemented by task-appropriate error analysis, calibration when probabilistic outputs inform decisions, subgroup and rare-condition performance, and temporal or institution-level transport. Prospective silent evaluation tests performance on future cases but does not establish how clinicians use the output or whether care improves. Claims of workflow utility require evaluation during clinical use, while causal claims about benefit and harm require an appropriate comparative interventional design \citep{ref_decide_ai2022}. Human factors, equity, privacy, cybersecurity, and operational resilience therefore remain distinct components of clinical readiness rather than properties inferred from an external AUC or Dice score.

Validation must also follow the complete deployed configuration through change. Alterations to adapters, prompts, retrieval resources, preprocessing, calibration, thresholds, or software dependencies can change clinically relevant behavior even when the backbone remains fixed. Parameter isolation may make a modification easier to identify or version, but it does not determine the scope of regression testing or demonstrate preserved safety. Lifecycle evaluation should therefore consider the behavioral impact of the complete system and its intended use rather than the proportion of parameters modified \citep{raji2020closing,mennella2024ethical}.

Regulatory interpretation must remain jurisdiction- and device-specific. IMDRF N41 provides harmonized principles for Software as a Medical Device (SaMD) clinical evaluation and treats evidence generation as an iterative lifecycle process rather than binding national regulation \citep{imdrf_samd_clinical_2017}. In the United States, the FDA's August 2025 final PCCP guidance provides nonbinding recommendations for specified planned modifications, the methods used to develop, validate, and implement them, and their impact assessment within applicable 510(k), De Novo, or PMA pathways. An authorized PCCP applies to the specified modifications and does not extend automatically to unspecified future changes \citep{fda2025pccp}. In Australia, TGA requirements depend on whether the software is a medical device, its intended purpose and risk classification, and the applicable Essential Principles, with current guidance addressing validation, configuration control, cybersecurity, change management, and post-market responsibilities \citep{tga2026software,tga2025aievidence}. Neither framework defines regulatory obligations according to the proportion of model parameters updated, and regulatory authorization should not be interpreted as evidence of clinical utility or readiness.

\subsection{FM-Specific Evidence Under Realistic Evaluation}
\label{subsec:fm_specific_realistic_evaluation}
Realistic evaluation is used here for evidence obtained beyond the development split under a clinically relevant change in institution, time, case mix, acquisition, specimen type, or workflow. These settings test different aspects of transport and should not be treated as equivalent to clinical readiness. The examples below are therefore interpreted according to the specific evaluation boundary they address. Direct evidence refers to an attributable post-pretraining intervention around a retained FM, while fixed-model assessments are used only as contextual evidence.

EAGLE provides the clearest prospective example in the reviewed adaptation evidence. A Prov-GigaPath-initialized pathology pathway was fine-tuned for slide-level EGFR mutation prediction in lung adenocarcinoma and achieved AUC 0.847 internally, 0.870 across external centers, and 0.890 in a prospective real-time silent trial of 197 primary-tumor samples \citep{campanella2025deployment}. Internal performance also differed by specimen type, with AUC 0.90 for primary and 0.75 for metastatic specimens. Because predictions did not guide care, the silent trial establishes prospective technical performance rather than an effect on clinician behavior or patient outcomes. The reported estimate of up to 43\% fewer rapid molecular tests is therefore a modeled workflow consequence of the prespecified threshold strategy rather than an observed interventional benefit.

Other studies address narrower forms of transport. Prov-GigaPath evaluated downstream adaptation in a 403-patient colorectal cohort collected after the stated March 2023 pretraining-data cutoff, providing temporal separation from FM pretraining within the same health system \citep{xu2024provgigapath}. RETFound provides cross-dataset evidence. After downstream fine-tuning for diabetic-retinopathy classification, AUROC was 0.943 on APTOS-2019 and 0.822 when the APTOS-trained model was evaluated on IDRiD \citep{zhou2023foundation}. These results demonstrate temporal or cross-dataset transport of the complete adapted pathways, but they do not isolate adaptation depth or establish institution-independent clinical utility.

Contextual evidence illustrates an additional boundary. In a standardized evaluation of 1,117 rare and atypical dermatology cases, image-only PanDerm achieved 72.2\% multiclass accuracy, compared with 68.2\% among readers with less than three years of experience and 74.2\% among readers with more than ten years. The multimodal PanDerm configuration achieved 66.3\% \citep{anriot2026limits}. Because this study evaluated fixed FM configurations rather than an attributable post-pretraining intervention, it informs difficult-case and modality evaluation rather than adaptation effectiveness. Taken together, these studies show that external-site, temporal, cross-dataset, prospective-silent, and difficult-case evaluation reveal different limits of transport. None alone establishes clinical utility, calibrated benefit and harm, regulatory readiness, or superiority of a particular update regime.

\section{Open Challenges and Research Directions}
\label{sec:open_challenges}

The evidence gaps identified in this review extend beyond individual adaptation mechanisms to how adapted FMs are updated, distributed, verified, and governed over time. Direct FM evidence remains limited for longitudinal adaptation, distributed clinical deployment, calibration, and lifecycle evaluation, while broader medical-AI and methodological literature provides mainly contextual guidance \citep{ref12_challenges_per_fm,ref9_fm_medical_comprehensive_survey}. The following subsections therefore frame continual and federated adaptation, verification and governance, and prospective evaluation as open empirical questions rather than preferred design directions. Consistent with the review taxonomy, no research priority is inferred from mechanism family or update depth.

\subsection{Continual Adaptation Without Degradation}

Clinical environments change over time through shifts in scanners, acquisition protocols, patient populations, disease prevalence, and clinical definitions \citep{kore2024empirical}. Direct FM evidence for continual adaptation, however, remains concentrated in ordered benchmark tasks rather than longitudinal clinical change. MedPEFT-CL retained a CLIPSeg-based pathway and combined parameter-efficient updates with replay and consolidation, increasing average Dice from 61.31\% to 78.84\% and reducing average forgetting from 20.32\% to 1.91\% across sequential segmentation tasks \citep{gao2026medpeftcl}. In the preprint sensitivity layer, MedSeqFT combined sequential updating, replay, and distillation, improving mean Dice over conventional full fine-tuning for both VoCo and UniMiSS+ \citep{ye2025medseqft}. These findings demonstrate retention under controlled task sequences, but not preservation of calibration, subgroup sensitivity, or previously validated behavior under real clinical drift.

At the mechanism level, continual adaptation is a stability--plasticity trade-off that is not resolved simply by reducing the number of updated parameters. Replay preserves only the variation represented in memory, distillation can retain teacher errors and shortcuts, and consolidation constrains adaptation according to estimated parameter importance. Conversely, stronger anchoring can prevent correction of genuine domain mismatch. General continual-learning and medical-imaging literature provides relevant methodological context \citep{ref148_experience_replay,wang2024continual_learning_survey,kumari2025continual,perkonigg2021dynamic}, but does not establish which balance is appropriate for a particular FM or intended use.

The main unresolved question is whether continual FM adaptation can preserve clinically relevant operating characteristics across genuine temporal change. Future longitudinal studies could distinguish acquisition drift, prevalence shift, label-definition change, and task addition, while examining retained-task performance, calibration, rare findings, subgroup behavior, and external transport after successive updates. Current evidence does not support a preferred trigger, memory mechanism, or update regime for continual clinical adaptation.

\subsection{Federated and Distributed Adaptation Under Privacy Constraints}

Federated adaptation allows institutions to update a shared FM without centralizing raw clinical data, but it does not by itself establish privacy, transportability, or a preferred update regime. Direct FM evidence illustrates this limitation. Asokan et al.\ retained a common SAM initialization and compared federated LoRA with selected decoder updating against full SAM fine-tuning for three-dimensional segmentation. Mean Dice was 0.6046 versus 0.5444 on Fed-KiTS2019 and 0.9834 versus 0.9811 on Fed-IXI, but 0.8867 versus 0.9084 on prostate MRI \citep{asokan2024federated}. The change in ranking across datasets indicates that reducing the communicated parameter set does not confer a consistent performance advantage.

At the mechanism level, federated adaptation must reconcile heterogeneous local distributions while limiting information disclosure. Differences in scanners, acquisition protocols, case mix, prevalence, and annotation practice can produce conflicting update directions across clients. Aggregating small adapters or low-rank updates can still combine incompatible gradients or propagate site-specific shortcuts, while personalized modules may improve local fit at the cost of increasingly different site-specific models. Frozen backbones reduce the amount of communicated trainable information but can preserve domain mismatch, whereas deeper updating increases adaptation capacity together with communication requirements. Federated exchange also does not guarantee confidentiality because model updates or intermediate representations may themselves reveal information \citep{rieke2020future,koutsoubis2024privacy,ref150_pfl}.

The unresolved question is therefore whether distributed FM adaptation can improve collaborative learning while preserving acceptable performance across both participating and unseen institutions. Future comparative studies could evaluate pooled, local, and federated variants from the same pretrained checkpoint while separating average from worst-site performance, calibration, subgroup behavior, communication cost, and privacy risk. Privacy mechanisms should be evaluated against an explicit threat model rather than treated as proof of confidentiality. Current evidence does not establish whether global, personalized, parameter-efficient, or deeper federated updating provides a generally preferable clinical trade-off.

\subsection{Verifiable and Governed Adaptation}

Trustworthiness cannot be inferred from an adaptation label, explanation map, or improvement in a single aggregate endpoint. Adaptation may improve the intended task while altering calibration, rare-finding sensitivity, subgroup performance, retained capabilities, or dependence on acquisition and contextual cues \citep{markus2021role,pfohl2022comparison}. Verification therefore requires distinguishing the behavior intended to change from behaviors expected to remain stable. Explanation methods may help investigate individual predictions, but they do not establish that an adapted FM is calibrated, externally transportable, equitable, or clinically useful.

The current evidence base leaves substantial gaps in this respect. Only 10 of 705 studies reported calibration or uncertainty evidence, while 173 had confirmed independent external validation, and these properties were rarely assessed under a common protocol. Future research could compare complete pre- and post-adaptation configurations while examining task-specific performance alongside calibration, rare findings, subgroup and worst-site behavior, retained-task capability, and dependence on prompts, retrieval, or nuisance information. The relevant endpoints differ by task, so Dice does not replace boundary assessment in segmentation, discrimination does not replace calibration in classification, and case-level performance does not replace lesion-level sensitivity in detection. Any acceptance limits or monitoring thresholds remain intended-use-specific research variables rather than universal properties of an adaptation mechanism.

Verification must also extend from the model to the complete human--FM system. Retrospective testing, external or temporal evaluation, prospective silent operation, clinician-assisted use, and interventional evaluation answer different questions and should not be collapsed into a single readiness claim \citep{ref_decide_ai2022,imdrf_gmlp_2025}. Open questions include how behavioral change should be monitored across versions, when revalidation or rollback is warranted, and how human factors, workflow effects, equity, cybersecurity, and post-deployment drift should be incorporated into adaptation evaluation. Current evidence does not support a universal verification threshold or indicate that any update regime or mechanism is intrinsically easier to govern.

Lifecycle governance introduces a related research problem. Regulatory expectations apply to the complete product, intended use, modification, and jurisdiction rather than to adaptation depth or trainable-parameter count \citep{imdrf_gmlp_2025,fda2025pccp,tga2026software,tga2025aievidence}. Parameter isolation may improve technical traceability, but whether versioned modules, prespecified change descriptions, rollback procedures, or other controls make behavioral change easier to validate remains an empirical question. Future work should therefore examine how successive FM modifications can be documented, compared, monitored, and revalidated without assuming that modular or parameter-efficient updating carries lower clinical or regulatory risk.

\subsection{Prospective Evaluation of the Complete Human and FM System}

Realistic evaluation remains distinct from evidence that an adapted system improves care. EAGLE provides a direct example. An end-to-end adapted Prov-GigaPath pathway for EGFR prediction achieved AUC 0.847 internally and 0.890 in a prospective real-time silent trial. Median model turnaround from molecular accession was 0.74 hours, compared with 48.78 hours for the reported Idylla rapid-PCR pathway \citep{campanella2025deployment}. These results demonstrate prospective technical operation, but clinicians did not act on the predictions. They therefore do not establish changes in diagnostic decisions, treatment timing, workload, patient outcomes, or overall benefit and harm.

Contextual evidence illustrates a related boundary. In 1,117 rare and atypical dermatology cases, fixed image-only PanDerm achieved 72.2\% multiclass accuracy, compared with 68.2\% among readers with less than three years of experience and 74.2\% among readers with more than ten years. Its multimodal configuration achieved 66.3\% \citep{anriot2026limits}. Because this study evaluated fixed FM configurations rather than an attributable post-pretraining intervention, it does not establish an adaptation effect. It does show that case mix, reader experience, and available clinical context can alter the apparent relation between FM and clinician performance, so benchmark superiority should not be interpreted as evidence of workflow utility.

A major open question is how evidence should progress from technical performance to evaluation of the complete clinical system. Retrospective testing, external or temporal validation, prospective silent operation, clinician-assisted evaluation, and interventional assessment answer different questions and should not be treated as equivalent evidence of readiness \citep{ref_decide_ai2022}. Human factors include how outputs are presented, correction and review burden, clinician reliance, and performance across user experience levels. Relevant system outcomes include timeliness, downstream testing, subgroup effects, equity, cybersecurity, and patient-relevant benefit and harm. The appropriate evaluation depends on intended use and risk, and the current evidence does not establish that technical accuracy, prospective silent operation, or regulatory authorization alone demonstrates clinical readiness.

\section{Limitations of This Review}
\label{sec:limitations}

This review was not prospectively registered. Searches were limited to five sources, English-language reports, bounded supplementary identification, and a 1 August 2026 cutoff. Of 1,096 reports sought for retrieval, 306 (27.9\%) remained unavailable after the documented retrieval procedures, so retrieval bias and incomplete evidence coverage remain possible. The evidence base was also highly time-sensitive, with 220 of 705 studies available only as preprints and the 2026 search period representing a partial year.

Several review processes required judgment despite structured reviewer assessment and consensus adjudication. These included eligibility, report-to-study linkage, adaptation coding, finding selection, and evidence appraisal. The appraisal framework was developed for this heterogeneous evidence base rather than drawn from a validated risk-of-bias instrument. Mechanism and update-regime coding was multi-label, some mixed or sequential pathways could not be fully decomposed, and information not identified in the prespecified extraction was recorded as not reported rather than inferred. Residual classification uncertainty therefore remains possible.

Substantial heterogeneity in FM definitions, pretraining, modalities, tasks, comparators, validation designs, and outcome measures precluded defensible corpus-wide effect pooling or universal strategy rankings. The nested comparative synthesis improved attribution by restricting interpretation to controlled, source-located within-study contrasts, but such comparisons remain sensitive to incomplete reporting, implementation differences, comparator confounding, and heterogeneous validation. A controlled within-study advantage therefore does not establish transportability, clinical utility, or superiority across settings.

The underlying literature also provided limited evidence for higher-level clinical claims. Independent external validation and calibration or uncertainty assessment were uncommon, and no study met the prespecified high-evidence-strength criteria. Prospective utility, workflow effects, human factors, equity, cybersecurity, and post-deployment monitoring were not systematically available across the corpus. Some interpretation therefore relies on contextual medical-AI, methodological, or regulatory literature. Such evidence can identify relevant risks and evaluation questions, but cannot establish comparative benefit, clinical safety, regulatory acceptability, or readiness of a particular FM-adaptation strategy.

\section{Conclusion}
\label{sec:conclusion}

This review synthesizes post-pretraining FM adaptation in MIA using three complementary dimensions: model-side locus, learning-control modifier, and update regime. Across the reviewed evidence, reported outcomes varied across retained FMs, tasks, modalities, reported supervision regimes, comparators, and evaluation settings. Controlled comparisons did not support a universal ranking of adaptation families or update depths, and parameter efficiency, representational alignment, or deeper updating should not be interpreted as intrinsic indicators of robustness, clinical suitability, or reduced validation burden.

The available evidence remains substantially stronger for technical performance than for external transport, calibration, prospective clinical utility, and lifecycle behavior. Interpretation of adaptation findings should therefore remain bounded to the intervention, comparator, population, and validation setting actually evaluated. More same-checkpoint comparisons with attributable contrasts, external and prospective evaluation, calibration, subgroup assessment, human-system evaluation, and lifecycle evidence would help establish whether particular adapted configurations remain reliable for their intended clinical use. Rather than identifying a universally optimal adaptation strategy, the current evidence points to the need to determine the conditions under which specific adaptation approaches produce reproducible and clinically meaningful behavior.
\backmatter

\section*{Declarations}

\noindent\textbf{Funding}\par
The authors did not receive support from any organization for the submitted work.

\medskip
\noindent\textbf{Competing interests}\par
The authors declare that they have no competing interests.

\medskip
\noindent\textbf{Ethics approval and consent to participate}\par
Not applicable.

\medskip
\noindent\textbf{Consent for publication}\par
Not applicable.

\medskip
\noindent\textbf{Data availability}\par
This review did not generate primary patient-level data. Data generated during the evidence synthesis, including the study-level evidence map, principal findings, study-selection and publication-status records, modality and adaptation summaries, and comparative-synthesis data, are provided in the Supplementary Reproducibility Appendix.

\medskip
\noindent\textbf{Materials availability}\par
Not applicable.

\medskip
\noindent\textbf{Code availability}\par
Not applicable.

\medskip
\noindent\textbf{Author contributions}\par
Karma Phuntsho conceived the review, performed the literature synthesis, developed the review framework, and prepared the original manuscript draft. Ickjai Lee provided major intellectual input into the review structure, critical analysis, and manuscript revision. Euijoon Ahn served as the principal academic advisor, providing overall guidance, critical review, and high-level feedback on the manuscript. Abdullah and Kyungmi Lee contributed to manuscript review and revision. All authors read and approved the final manuscript.

\begin{appendices}




\end{appendices}


\bibliography{sn-bibliography}% common bib file


\begin{thebibliography}{197}
\ifx \bisbn   \undefined \def \bisbn  #1{ISBN #1}\fi
\ifx \binits  \undefined \def \binits#1{#1}\fi
\ifx \bauthor  \undefined \def \bauthor#1{#1}\fi
\ifx \batitle  \undefined \def \batitle#1{#1}\fi
\ifx \bjtitle  \undefined \def \bjtitle#1{#1}\fi
\ifx \bvolume  \undefined \def \bvolume#1{\textbf{#1}}\fi
\ifx \byear  \undefined \def \byear#1{#1}\fi
\ifx \bissue  \undefined \def \bissue#1{#1}\fi
\ifx \bfpage  \undefined \def \bfpage#1{#1}\fi
\ifx \blpage  \undefined \def \blpage #1{#1}\fi
\ifx \burl  \undefined \def \burl#1{\textsf{#1}}\fi
\ifx \doiurl  \undefined \def \doiurl#1{\url{https://doi.org/#1}}\fi
\ifx \betal  \undefined \def \betal{\textit{et al.}}\fi
\ifx \binstitute  \undefined \def \binstitute#1{#1}\fi
\ifx \binstitutionaled  \undefined \def \binstitutionaled#1{#1}\fi
\ifx \bctitle  \undefined \def \bctitle#1{#1}\fi
\ifx \beditor  \undefined \def \beditor#1{#1}\fi
\ifx \bpublisher  \undefined \def \bpublisher#1{#1}\fi
\ifx \bbtitle  \undefined \def \bbtitle#1{#1}\fi
\ifx \bedition  \undefined \def \bedition#1{#1}\fi
\ifx \bseriesno  \undefined \def \bseriesno#1{#1}\fi
\ifx \blocation  \undefined \def \blocation#1{#1}\fi
\ifx \bsertitle  \undefined \def \bsertitle#1{#1}\fi
\ifx \bsnm \undefined \def \bsnm#1{#1}\fi
\ifx \bsuffix \undefined \def \bsuffix#1{#1}\fi
\ifx \bparticle \undefined \def \bparticle#1{#1}\fi
\ifx \barticle \undefined \def \barticle#1{#1}\fi
\bibcommenthead
\ifx \bconfdate \undefined \def \bconfdate #1{#1}\fi
\ifx \botherref \undefined \def \botherref #1{#1}\fi
\ifx \url \undefined \def \url#1{\textsf{#1}}\fi
\ifx \bchapter \undefined \def \bchapter#1{#1}\fi
\ifx \bbook \undefined \def \bbook#1{#1}\fi
\ifx \bcomment \undefined \def \bcomment#1{#1}\fi
\ifx \oauthor \undefined \def \oauthor#1{#1}\fi
\ifx \citeauthoryear \undefined \def \citeauthoryear#1{#1}\fi
\ifx \endbibitem  \undefined \def \endbibitem {}\fi
\ifx \bconflocation  \undefined \def \bconflocation#1{#1}\fi
\ifx \arxivurl  \undefined \def \arxivurl#1{\textsf{#1}}\fi
\csname PreBibitemsHook\endcsname

%%% 1
\bibitem[\protect\citeauthoryear{Azad
  et~al.}{2023}]{ref9_fm_medical_comprehensive_survey}
\begin{botherref}
\oauthor{\bsnm{Azad}, \binits{B.}},
\oauthor{\bsnm{Azad}, \binits{R.}},
\oauthor{\bsnm{Eskandari}, \binits{S.}},
\oauthor{\bsnm{Bozorgpour}, \binits{A.}},
\oauthor{\bsnm{Kazerouni}, \binits{A.}},
\oauthor{\bsnm{Rekik}, \binits{I.}},
\oauthor{\bsnm{Merhof}, \binits{D.}}:
Foundational models in medical imaging: A comprehensive survey and future
  vision.
arXiv preprint arXiv:2310.18689
(2023)
\end{botherref}
\endbibitem

%%% 2
\bibitem[\protect\citeauthoryear{Asokan et~al.}{2024}]{asokan2024federated}
\begin{bchapter}
\bauthor{\bsnm{Asokan}, \binits{M.}},
\bauthor{\bsnm{Benjamin}, \binits{J.G.}},
\bauthor{\bsnm{Yaqub}, \binits{M.}},
\bauthor{\bsnm{Nandakumar}, \binits{K.}}:
\bctitle{A federated learning-friendly approach for parameter-efficient
  fine-tuning of sam in 3d segmentation}.
In: \bbtitle{International Conference on Medical Image Computing and
  Computer-Assisted Intervention},
pp. \bfpage{226}--\blpage{235}
(\byear{2024}).
\bcomment{Springer}
\end{bchapter}
\endbibitem

%%% 3
\bibitem[\protect\citeauthoryear{Azizi et~al.}{2021}]{ref59_big_ssl}
\begin{bchapter}
\bauthor{\bsnm{Azizi}, \binits{S.}},
\bauthor{\bsnm{Mustafa}, \binits{B.}},
\bauthor{\bsnm{Ryan}, \binits{F.}},
\bauthor{\bsnm{Beaver}, \binits{Z.}},
\bauthor{\bsnm{Freyberg}, \binits{J.}},
\bauthor{\bsnm{Deaton}, \binits{J.}},
\bauthor{\bsnm{Loh}, \binits{A.}},
\bauthor{\bsnm{Karthikesalingam}, \binits{A.}},
\bauthor{\bsnm{Kornblith}, \binits{S.}},
\bauthor{\bsnm{Chen}, \binits{T.}}, \betal:
\bctitle{Big self-supervised models advance medical image classification}.
In: \bbtitle{Proceedings of the IEEE/CVF International Conference on Computer
  Vision},
pp. \bfpage{3478}--\blpage{3488}
(\byear{2021})
\end{bchapter}
\endbibitem

%%% 4
\bibitem[\protect\citeauthoryear{Anriot et~al.}{2026}]{anriot2026limits}
\begin{barticle}
\bauthor{\bsnm{Anriot}, \binits{J.}},
\bauthor{\bsnm{Yan}, \binits{S.}},
\bauthor{\bsnm{Coste}, \binits{C.}},
\bauthor{\bsnm{Tschandl}, \binits{P.}},
\bauthor{\bsnm{Verlingue}, \binits{L.}},
\bauthor{\bsnm{Andremasse}, \binits{C.}},
\bauthor{\bsnm{Amini-Adle}, \binits{M.}},
\bauthor{\bsnm{Perrot}, \binits{J.L.}},
\bauthor{\bsnm{Ge}, \binits{Z.}},
\bauthor{\bsnm{Kittler}, \binits{H.}},
\bauthor{\bsnm{Thomas}, \binits{L.}}:
\batitle{Limits of artificial intelligence models for skin cancer diagnosis in
  realistic settings}.
\bjtitle{JAMA Dermatology}
\bvolume{162}(\bissue{7}),
\bfpage{701}--\blpage{708}
(\byear{2026})
\doiurl{10.1001/jamadermatol.2026.1492}
\end{barticle}
\endbibitem

%%% 5
\bibitem[\protect\citeauthoryear{Banerjee et~al.}{2023}]{banerjee2023shortcuts}
\begin{barticle}
\bauthor{\bsnm{Banerjee}, \binits{I.}},
\bauthor{\bsnm{Bhattacharjee}, \binits{K.}},
\bauthor{\bsnm{Burns}, \binits{J.L.}},
\bauthor{\bsnm{Trivedi}, \binits{H.}},
\bauthor{\bsnm{Purkayastha}, \binits{S.}},
\bauthor{\bsnm{Seyyed-Kalantari}, \binits{L.}},
\bauthor{\bsnm{Patel}, \binits{B.N.}},
\bauthor{\bsnm{Shiradkar}, \binits{R.}},
\bauthor{\bsnm{Gichoya}, \binits{J.}}:
\batitle{“shortcuts” causing bias in radiology artificial intelligence:
  causes, evaluation, and mitigation}.
\bjtitle{Journal of the American College of Radiology}
\bvolume{20}(\bissue{9}),
\bfpage{842}--\blpage{851}
(\byear{2023})
\end{barticle}
\endbibitem

%%% 6
\bibitem[\protect\citeauthoryear{Bommasani et~al.}{2021}]{ref10_fm}
\begin{botherref}
\oauthor{\bsnm{Bommasani}, \binits{R.}},
\oauthor{\bsnm{Hudson}, \binits{D.A.}},
\oauthor{\bsnm{Adeli}, \binits{E.}},
\oauthor{\bsnm{Altman}, \binits{R.}},
\oauthor{\bsnm{Arora}, \binits{S.}},
\oauthor{\bsnm{Arx}, \binits{S.}},
\oauthor{\bsnm{Bernstein}, \binits{M.S.}},
\oauthor{\bsnm{Bohg}, \binits{J.}},
\oauthor{\bsnm{Bosselut}, \binits{A.}},
\oauthor{\bsnm{Brunskill}, \binits{E.}}, et al.:
On the opportunities and risks of foundation models.
arXiv preprint arXiv:2108.07258
(2021)
\end{botherref}
\endbibitem

%%% 7
\bibitem[\protect\citeauthoryear{Blankemeier
  et~al.}{2026}]{blankemeier2026merlin}
\begin{barticle}
\bauthor{\bsnm{Blankemeier}, \binits{L.}},
\bauthor{\bsnm{Kumar}, \binits{A.}},
\bauthor{\bsnm{Cohen}, \binits{J.P.}},
\bauthor{\bsnm{Liu}, \binits{J.}},
\bauthor{\bsnm{Van~Veen}, \binits{D.}},
\bauthor{\bsnm{Paschali}, \binits{M.}},
\bauthor{\bsnm{Chen}, \binits{Z.}},
\bauthor{\bsnm{Delbrouck}, \binits{J.-B.}},
\bauthor{\bsnm{Reis}, \binits{E.}},
\bauthor{\bsnm{Holland}, \binits{R.}}, \betal:
\batitle{Merlin: a computed tomography vision--language foundation model and
  dataset}.
\bjtitle{Nature}
\bvolume{652}(\bissue{8112}),
\bfpage{1318}--\blpage{1328}
(\byear{2026})
\doiurl{10.1038/s41586-026-10181-8}
\end{barticle}
\endbibitem

%%% 8
\bibitem[\protect\citeauthoryear{Bengio et~al.}{2009}]{bengio2009curriculum}
\begin{bchapter}
\bauthor{\bsnm{Bengio}, \binits{Y.}},
\bauthor{\bsnm{Louradour}, \binits{J.}},
\bauthor{\bsnm{Collobert}, \binits{R.}},
\bauthor{\bsnm{Weston}, \binits{J.}}:
\bctitle{Curriculum learning}.
In: \bbtitle{Proceedings of the 26th Annual International Conference on Machine
  Learning},
pp. \bfpage{41}--\blpage{48}
(\byear{2009})
\end{bchapter}
\endbibitem

%%% 9
\bibitem[\protect\citeauthoryear{Bhattacharya
  et~al.}{2025}]{bhattacharya2025corrfabr}
\begin{barticle}
\bauthor{\bsnm{Bhattacharya}, \binits{I.}},
\bauthor{\bsnm{Stacke}, \binits{K.}},
\bauthor{\bsnm{Chan}, \binits{E.}},
\bauthor{\bsnm{Lee}, \binits{J.H.}},
\bauthor{\bsnm{Tse}, \binits{J.R.}},
\bauthor{\bsnm{Liang}, \binits{T.}},
\bauthor{\bsnm{Brooks}, \binits{J.D.}},
\bauthor{\bsnm{Sonn}, \binits{G.A.}},
\bauthor{\bsnm{Rusu}, \binits{M.}}:
\batitle{Aggressiveness classification of clear cell renal cell carcinoma using
  registration-independent radiology--pathology correlation learning}.
\bjtitle{Medical Physics}
\bvolume{52}(\bissue{1}),
\bfpage{300}--\blpage{320}
(\byear{2025})
\doiurl{10.1002/mp.17476}
\end{barticle}
\endbibitem

%%% 10
\bibitem[\protect\citeauthoryear{Chen et~al.}{2023}]{ref52_mim3d}
\begin{bchapter}
\bauthor{\bsnm{Chen}, \binits{Z.}},
\bauthor{\bsnm{Agarwal}, \binits{D.}},
\bauthor{\bsnm{Aggarwal}, \binits{K.}},
\bauthor{\bsnm{Safta}, \binits{W.}},
\bauthor{\bsnm{Balan}, \binits{M.M.}},
\bauthor{\bsnm{Brown}, \binits{K.}}:
\bctitle{Masked image modeling advances 3d medical image analysis}.
In: \bbtitle{Proceedings of the IEEE/CVF Winter Conference on Applications of
  Computer Vision (WACV)},
pp. \bfpage{1970}--\blpage{1980}
(\byear{2023})
\end{bchapter}
\endbibitem

%%% 11
\bibitem[\protect\citeauthoryear{Carloni et~al.}{2025}]{carloni2025scangen}
\begin{bchapter}
\bauthor{\bsnm{Carloni}, \binits{G.}},
\bauthor{\bsnm{Brattoli}, \binits{B.}},
\bauthor{\bsnm{Keum}, \binits{S.}},
\bauthor{\bsnm{Park}, \binits{J.}},
\bauthor{\bsnm{Lee}, \binits{T.}},
\bauthor{\bsnm{Ho~Ahn}, \binits{C.}},
\bauthor{\bsnm{Pereira}, \binits{S.}}:
\bctitle{Pathology foundation models are scanner sensitive: Benchmark and
  mitigation with contrastive scangen loss}.
In: \bbtitle{International Workshop on Foundation Models for General Medical
  AI},
pp. \bfpage{44}--\blpage{53}
(\byear{2025}).
\bcomment{Springer}
\end{bchapter}
\endbibitem

%%% 12
\bibitem[\protect\citeauthoryear{Chen et~al.}{2018}]{chen2018gradnorm}
\begin{bchapter}
\bauthor{\bsnm{Chen}, \binits{Z.}},
\bauthor{\bsnm{Badrinarayanan}, \binits{V.}},
\bauthor{\bsnm{Lee}, \binits{C.-Y.}},
\bauthor{\bsnm{Rabinovich}, \binits{A.}}:
\bctitle{Gradnorm: Gradient normalization for adaptive loss balancing in deep
  multitask networks}.
In: \bbtitle{International Conference on Machine Learning},
pp. \bfpage{794}--\blpage{803}
(\byear{2018}).
\bcomment{PMLR}
\end{bchapter}
\endbibitem

%%% 13
\bibitem[\protect\citeauthoryear{Cai et~al.}{2024}]{cai2024rethinking}
\begin{bchapter}
\bauthor{\bsnm{Cai}, \binits{Y.}},
\bauthor{\bsnm{Chen}, \binits{H.}},
\bauthor{\bsnm{Cheng}, \binits{K.-T.}}:
\bctitle{Rethinking autoencoders for medical anomaly detection from a
  theoretical perspective}.
In: \bbtitle{International Conference on Medical Image Computing and
  Computer-Assisted Intervention},
pp. \bfpage{544}--\blpage{554}
(\byear{2024}).
\bcomment{Springer}
\end{bchapter}
\endbibitem

%%% 14
\bibitem[\protect\citeauthoryear{Chen et~al.}{2024}]{chen2024towards}
\begin{barticle}
\bauthor{\bsnm{Chen}, \binits{R.J.}},
\bauthor{\bsnm{Ding}, \binits{T.}},
\bauthor{\bsnm{Lu}, \binits{M.Y.}},
\bauthor{\bsnm{Williamson}, \binits{D.F.}},
\bauthor{\bsnm{Jaume}, \binits{G.}},
\bauthor{\bsnm{Song}, \binits{A.H.}},
\bauthor{\bsnm{Chen}, \binits{B.}},
\bauthor{\bsnm{Zhang}, \binits{A.}},
\bauthor{\bsnm{Shao}, \binits{D.}},
\bauthor{\bsnm{Shaban}, \binits{M.}}, \betal:
\batitle{Towards a general-purpose foundation model for computational
  pathology}.
\bjtitle{Nature medicine}
\bvolume{30}(\bissue{3}),
\bfpage{850}--\blpage{862}
(\byear{2024})
\end{barticle}
\endbibitem

%%% 15
\bibitem[\protect\citeauthoryear{Cho et~al.}{2023}]{cho2023anti}
\begin{bchapter}
\bauthor{\bsnm{Cho}, \binits{H.}},
\bauthor{\bsnm{Han}, \binits{Y.}},
\bauthor{\bsnm{Kim}, \binits{W.H.}}:
\bctitle{Anti-adversarial consistency regularization for data augmentation:
  applications to robust medical image segmentation}.
In: \bbtitle{International Conference on Medical Image Computing and
  Computer-Assisted Intervention},
pp. \bfpage{555}--\blpage{566}
(\byear{2023}).
\bcomment{Springer}
\end{bchapter}
\endbibitem

%%% 16
\bibitem[\protect\citeauthoryear{Chakrapani
  et~al.}{2026}]{chakrapani2026retfounddril}
\begin{barticle}
\bauthor{\bsnm{Chakrapani}, \binits{P.K.}},
\bauthor{\bsnm{Kumar}, \binits{P.}},
\bauthor{\bsnm{Bhandary}, \binits{S.V.}},
\bauthor{\bsnm{Maiya}, \binits{G.}},
\bauthor{\bsnm{Shenoy}, \binits{S.}},
\bauthor{\bsnm{Fernandes}, \binits{S.}}:
\batitle{Fine tuning {RETFound} with clinically guided foveal {ROI} for
  automated {DRIL} classification in diabetic macular edema {OCT}}.
\bjtitle{Diagnostics}
\bvolume{16}(\bissue{11}),
\bfpage{1654}
(\byear{2026})
\doiurl{10.3390/diagnostics16111654}
\end{barticle}
\endbibitem

%%% 17
\bibitem[\protect\citeauthoryear{Campanella
  et~al.}{2025}]{campanella2025deployment}
\begin{barticle}
\bauthor{\bsnm{Campanella}, \binits{G.}},
\bauthor{\bsnm{Kumar}, \binits{N.}},
\bauthor{\bsnm{Nanda}, \binits{S.}},
\bauthor{\bsnm{Singi}, \binits{S.}},
\bauthor{\bsnm{Fluder}, \binits{E.}},
\bauthor{\bsnm{Kwan}, \binits{R.}},
\bauthor{\bsnm{Muehlstedt}, \binits{S.}},
\bauthor{\bsnm{Pfarr}, \binits{N.}},
\bauthor{\bsnm{Sch{\"u}ffler}, \binits{P.J.}},
\bauthor{\bsnm{H{\"a}ggstr{\"o}m}, \binits{I.}},
\bauthor{\bsnm{Neittaanm{\"a}ki}, \binits{N.}},
\bauthor{\bsnm{Aky{\"u}rek}, \binits{L.M.}},
\bauthor{\bsnm{Basnet}, \binits{A.}},
\bauthor{\bsnm{Jamaspishvili}, \binits{T.}},
\bauthor{\bsnm{Nasr}, \binits{M.R.}},
\bauthor{\bsnm{Croken}, \binits{M.M.}},
\bauthor{\bsnm{Hirsch}, \binits{F.R.}},
\bauthor{\bsnm{Elkrief}, \binits{A.}},
\bauthor{\bsnm{Yu}, \binits{H.}},
\bauthor{\bsnm{Ardon}, \binits{O.}},
\bauthor{\bsnm{Goldgof}, \binits{G.M.}},
\bauthor{\bsnm{Hameed}, \binits{M.}},
\bauthor{\bsnm{Houldsworth}, \binits{J.}},
\bauthor{\bsnm{Arcila}, \binits{M.}},
\bauthor{\bsnm{Fuchs}, \binits{T.J.}},
\bauthor{\bsnm{Vanderbilt}, \binits{C.}}:
\batitle{Real-world deployment of a fine-tuned pathology foundation model for
  lung cancer biomarker detection}.
\bjtitle{Nature Medicine}
\bvolume{31}(\bissue{9}),
\bfpage{3002}--\blpage{3010}
(\byear{2025})
\doiurl{10.1038/s41591-025-03780-x}
\end{barticle}
\endbibitem

%%% 18
\bibitem[\protect\citeauthoryear{Cho et~al.}{2025}]{cho2025g2l}
\begin{botherref}
\oauthor{\bsnm{Cho}, \binits{Y.}},
\oauthor{\bsnm{Lee}, \binits{S.}},
\oauthor{\bsnm{Lee}, \binits{G.}},
\oauthor{\bsnm{Lee}, \binits{M.}},
\oauthor{\bsnm{Park}, \binits{J.}},
\oauthor{\bsnm{Shin}, \binits{D.}}:
G2l: From giga-scale to cancer-specific large-scale pathology foundation models
  via knowledge distillation.
arXiv preprint arXiv:2510.11176
(2025)
\end{botherref}
\endbibitem

%%% 19
\bibitem[\protect\citeauthoryear{Chen et~al.}{2022}]{ref91_revisiting_peft}
\begin{bchapter}
\bauthor{\bsnm{Chen}, \binits{G.}},
\bauthor{\bsnm{Liu}, \binits{F.}},
\bauthor{\bsnm{Meng}, \binits{Z.}},
\bauthor{\bsnm{Liang}, \binits{S.}}:
\bctitle{Revisiting parameter-efficient tuning: Are we really there yet?}
In: \bbtitle{Proceedings of the 2022 Conference on Empirical Methods in Natural
  Language Processing},
pp. \bfpage{2612}--\blpage{2626}
(\byear{2022})
\end{bchapter}
\endbibitem

%%% 20
\bibitem[\protect\citeauthoryear{Chen et~al.}{2024}]{ref127_masam}
\begin{barticle}
\bauthor{\bsnm{Chen}, \binits{C.}},
\bauthor{\bsnm{Miao}, \binits{J.}},
\bauthor{\bsnm{Wu}, \binits{D.}},
\bauthor{\bsnm{Zhong}, \binits{A.}},
\bauthor{\bsnm{Yan}, \binits{Z.}},
\bauthor{\bsnm{Kim}, \binits{S.}},
\bauthor{\bsnm{Hu}, \binits{J.}},
\bauthor{\bsnm{Liu}, \binits{Z.}},
\bauthor{\bsnm{Sun}, \binits{L.}},
\bauthor{\bsnm{Li}, \binits{X.}},
\bauthor{\bsnm{Liu}, \binits{T.}},
\bauthor{\bsnm{Heng}, \binits{P.-A.}},
\bauthor{\bsnm{Li}, \binits{Q.}}:
\batitle{Ma-sam: Modality-agnostic sam adaptation for 3d medical image
  segmentation}.
\bjtitle{Medical Image Analysis}
\bvolume{98},
\bfpage{103310}
(\byear{2024})
\doiurl{10.1016/j.media.2024.103310}
\end{barticle}
\endbibitem

%%% 21
\bibitem[\protect\citeauthoryear{Chung et~al.}{2025}]{chung2025autoencoder}
\begin{bchapter}
\bauthor{\bsnm{Chung}, \binits{W.}},
\bauthor{\bsnm{Park}, \binits{Y.}},
\bauthor{\bsnm{Nam}, \binits{Y.}}:
\bctitle{Autoencoder-based feature transformation with multiple foundation
  models in computational pathology}.
In: \bbtitle{Foundation Models for General Medical AI}.
\bsertitle{Lecture Notes in Computer Science},
vol. \bseriesno{15184}
(\byear{2025}).
\doiurl{10.1007/978-3-031-73471-7_5}
\end{bchapter}
\endbibitem

%%% 22
\bibitem[\protect\citeauthoryear{Chakrabarty
  et~al.}{2026}]{chakrabarty2026synthfm3d}
\begin{barticle}
\bauthor{\bsnm{Chakrabarty}, \binits{S.}},
\bauthor{\bsnm{Sengupta}, \binits{S.}},
\bauthor{\bsnm{Avinash}, \binits{G.}},
\bauthor{\bsnm{Soni}, \binits{R.}}:
\batitle{Synthetic volumetric data generation enables zero-shot generalization
  of foundation models in 3d medical image segmentation}.
\bjtitle{arXiv preprint arXiv:2601.12297}
(\byear{2026})
\doiurl{10.48550/arXiv.2601.12297}
\end{barticle}
\endbibitem

%%% 23
\bibitem[\protect\citeauthoryear{Chen et~al.}{2025}]{chen2025diffusionclip}
\begin{barticle}
\bauthor{\bsnm{Chen}, \binits{Y.}},
\bauthor{\bsnm{Tao}, \binits{H.}},
\bauthor{\bsnm{Yang}, \binits{Z.}},
\bauthor{\bsnm{Cao}, \binits{Y.}},
\bauthor{\bsnm{Jiang}, \binits{C.}},
\bauthor{\bsnm{Hu}, \binits{L.}},
\bauthor{\bsnm{Xiong}, \binits{P.}},
\bauthor{\bsnm{Qiu}, \binits{H.}}:
\batitle{Diffusion-based vision-language model for zero-shot anomaly detection
  in medical images}.
\bjtitle{Engineering Applications of Artificial Intelligence}
\bvolume{161},
\bfpage{112181}
(\byear{2025})
\doiurl{10.1016/j.engappai.2025.112181}
\end{barticle}
\endbibitem

%%% 24
\bibitem[\protect\citeauthoryear{Cai et~al.}{2024}]{cai2024wspolypsam}
\begin{barticle}
\bauthor{\bsnm{Cai}, \binits{T.}},
\bauthor{\bsnm{Yan}, \binits{H.}},
\bauthor{\bsnm{Ding}, \binits{K.}},
\bauthor{\bsnm{Zhang}, \binits{Y.}},
\bauthor{\bsnm{Zhou}, \binits{Y.}}:
\batitle{{WSPolyp-SAM}: Weakly supervised and self-guided fine-tuning of {SAM}
  for colonoscopy polyp segmentation}.
\bjtitle{Applied Sciences}
\bvolume{14}(\bissue{12}),
\bfpage{5007}
(\byear{2024})
\doiurl{10.3390/app14125007}
\end{barticle}
\endbibitem

%%% 25
\bibitem[\protect\citeauthoryear{Chen et~al.}{2024}]{chen2024msvmunet}
\begin{bchapter}
\bauthor{\bsnm{Chen}, \binits{C.}},
\bauthor{\bsnm{Yu}, \binits{L.}},
\bauthor{\bsnm{Min}, \binits{S.}},
\bauthor{\bsnm{Wang}, \binits{S.}}:
\bctitle{Msvm-unet: Multi-scale vision mamba unet for medical image
  segmentation}.
In: \bbtitle{2024 IEEE International Conference on Bioinformatics and
  Biomedicine (BIBM)},
pp. \bfpage{3111}--\blpage{3114}
(\byear{2024}).
\bcomment{IEEE}
\end{bchapter}
\endbibitem

%%% 26
\bibitem[\protect\citeauthoryear{D’Antonoli et~al.}{2026}]{d2026foundation}
\begin{barticle}
\bauthor{\bsnm{D’Antonoli}, \binits{T.A.}},
\bauthor{\bsnm{Bluethgen}, \binits{C.}},
\bauthor{\bsnm{Cuocolo}, \binits{R.}},
\bauthor{\bsnm{Klontzas}, \binits{M.E.}},
\bauthor{\bsnm{Ponsiglione}, \binits{A.}},
\bauthor{\bsnm{Kocak}, \binits{B.}}:
\batitle{Foundation models for radiology: fundamentals, applications,
  opportunities, challenges, risks, and prospects}.
\bjtitle{Diagnostic and Interventional Radiology}
\bvolume{32}(\bissue{3}),
\bfpage{259}
(\byear{2026})
\end{barticle}
\endbibitem

%%% 27
\bibitem[\protect\citeauthoryear{Dutt et~al.}{2024}]{ref96_peft_mia}
\begin{bchapter}
\bauthor{\bsnm{Dutt}, \binits{R.}},
\bauthor{\bsnm{Ericsson}, \binits{L.}},
\bauthor{\bsnm{Sanchez}, \binits{P.}},
\bauthor{\bsnm{Tsaftaris}, \binits{S.A.}},
\bauthor{\bsnm{Hospedales}, \binits{T.}}:
\bctitle{Parameter-efficient fine-tuning for medical image analysis: The missed
  opportunity}.
In: \beditor{\bsnm{Burgos}, \binits{N.}},
\beditor{\bsnm{Petitjean}, \binits{C.}},
\beditor{\bsnm{Vakalopoulou}, \binits{M.}},
\beditor{\bsnm{Christodoulidis}, \binits{S.}},
\beditor{\bsnm{Coupe}, \binits{P.}},
\beditor{\bsnm{Delingette}, \binits{H.}},
\beditor{\bsnm{Lartizien}, \binits{C.}},
\beditor{\bsnm{Mateus}, \binits{D.}} (eds.)
\bbtitle{Proceedings of the 7th International Conference on Medical Imaging
  with Deep Learning}.
\bsertitle{Proceedings of Machine Learning Research},
vol. \bseriesno{250},
pp. \bfpage{406}--\blpage{425}.
\bpublisher{PMLR},
\blocation{Paris, France}
(\byear{2024}).
\burl{https://proceedings.mlr.press/v250/dutt24a.html}
\end{bchapter}
\endbibitem

%%% 28
\bibitem[\protect\citeauthoryear{DeGrave et~al.}{2021}]{degrave2021ai}
\begin{barticle}
\bauthor{\bsnm{DeGrave}, \binits{A.J.}},
\bauthor{\bsnm{Janizek}, \binits{J.D.}},
\bauthor{\bsnm{Lee}, \binits{S.-I.}}:
\batitle{Ai for radiographic covid-19 detection selects shortcuts over signal}.
\bjtitle{Nature Machine Intelligence}
\bvolume{3}(\bissue{7}),
\bfpage{610}--\blpage{619}
(\byear{2021})
\end{barticle}
\endbibitem

%%% 29
\bibitem[\protect\citeauthoryear{Dalmaz et~al.}{2024}]{ref150_pfl}
\begin{barticle}
\bauthor{\bsnm{Dalmaz}, \binits{O.}},
\bauthor{\bsnm{Mirza}, \binits{M.U.}},
\bauthor{\bsnm{Elmas}, \binits{G.}},
\bauthor{\bsnm{Ozbey}, \binits{M.}},
\bauthor{\bsnm{Dar}, \binits{S.U.}},
\bauthor{\bsnm{Ceyani}, \binits{E.}},
\bauthor{\bsnm{Oguz}, \binits{K.K.}},
\bauthor{\bsnm{Avestimehr}, \binits{S.}},
\bauthor{\bsnm{{\c{C}}ukur}, \binits{T.}}:
\batitle{One model to unite them all: Personalized federated learning of
  multi-contrast mri synthesis}.
\bjtitle{Medical Image Analysis}
\bvolume{94},
\bfpage{103121}
(\byear{2024})
\end{barticle}
\endbibitem

%%% 30
\bibitem[\protect\citeauthoryear{Ding et~al.}{2022}]{ref80_delta_tuning}
\begin{botherref}
\oauthor{\bsnm{Ding}, \binits{N.}},
\oauthor{\bsnm{Qin}, \binits{Y.}},
\oauthor{\bsnm{Yang}, \binits{G.}},
\oauthor{\bsnm{Wei}, \binits{F.}},
\oauthor{\bsnm{Yang}, \binits{Z.}},
\oauthor{\bsnm{Su}, \binits{Y.}},
\oauthor{\bsnm{Hu}, \binits{S.}},
\oauthor{\bsnm{Chen}, \binits{Y.}},
\oauthor{\bsnm{Chan}, \binits{C.-M.}},
\oauthor{\bsnm{Chen}, \binits{W.}}, et al.:
Delta tuning: A comprehensive study of parameter efficient methods for
  pre-trained language models.
arXiv preprint arXiv:2203.06904
(2022)
\end{botherref}
\endbibitem

%%% 31
\bibitem[\protect\citeauthoryear{Ehteshami~Bejnordi
  et~al.}{2017}]{ehteshami2017diagnostic}
\begin{barticle}
\bauthor{\bsnm{Ehteshami~Bejnordi}, \binits{B.}},
\bauthor{\bsnm{Veta}, \binits{M.}},
\bauthor{\bsnm{Diest}, \binits{P.}},
\bauthor{\bsnm{Van~Ginneken}, \binits{B.}},
\bauthor{\bsnm{Karssemeijer}, \binits{N.}},
\bauthor{\bsnm{Litjens}, \binits{G.}},
\bauthor{\bsnm{Van Der~Laak}, \binits{J.A.}},
\bauthor{\bsnm{{CAMELYON16 consortium}}},
\bauthor{\bsnm{Hermsen}, \binits{M.}},
\bauthor{\bsnm{Manson}, \binits{Q.F.}}, \betal:
\batitle{Diagnostic assessment of deep learning algorithms for detection of
  lymph node metastases in women with breast cancer}.
\bjtitle{Jama}
\bvolume{318}(\bissue{22}),
\bfpage{2199}--\blpage{2210}
(\byear{2017})
\end{barticle}
\endbibitem

%%% 32
\bibitem[\protect\citeauthoryear{Enda et~al.}{2026}]{enda2026pathologyfm}
\begin{barticle}
\bauthor{\bsnm{Enda}, \binits{K.}},
\bauthor{\bsnm{Oda}, \binits{Y.}},
\bauthor{\bsnm{Tanei}, \binits{Z.-i.}},
\bauthor{\bsnm{Satoh}, \binits{K.}},
\bauthor{\bsnm{Motegi}, \binits{H.}},
\bauthor{\bsnm{Terasaka}, \binits{S.}},
\bauthor{\bsnm{Yamaguchi}, \binits{S.}},
\bauthor{\bsnm{Ogawa}, \binits{T.}},
\bauthor{\bsnm{Wang}, \binits{L.}},
\bauthor{\bsnm{Tsuda}, \binits{M.}},
\bauthor{\bsnm{Tanaka}, \binits{S.}}:
\batitle{Transfer learning strategies for pathological foundation models: A
  systematic evaluation in brain tumor classification}.
\bjtitle{Pathology International}
\bvolume{76}(\bissue{2}),
\bfpage{70098}
(\byear{2026})
\doiurl{10.1111/pin.70098}
\end{barticle}
\endbibitem

%%% 33
\bibitem[\protect\citeauthoryear{Fischer et~al.}{2024}]{ref88_prompt_tuning1}
\begin{barticle}
\bauthor{\bsnm{Fischer}, \binits{M.}},
\bauthor{\bsnm{Bartler}, \binits{A.}},
\bauthor{\bsnm{Yang}, \binits{B.}}:
\batitle{Prompt tuning for parameter-efficient medical image segmentation}.
\bjtitle{Medical Image Analysis}
\bvolume{91},
\bfpage{103024}
(\byear{2024})
\doiurl{10.1016/j.media.2023.103024}
\end{barticle}
\endbibitem

%%% 34
\bibitem[\protect\citeauthoryear{Faghani et~al.}{2023}]{faghani2023quantifying}
\begin{barticle}
\bauthor{\bsnm{Faghani}, \binits{S.}},
\bauthor{\bsnm{Moassefi}, \binits{M.}},
\bauthor{\bsnm{Rouzrokh}, \binits{P.}},
\bauthor{\bsnm{Khosravi}, \binits{B.}},
\bauthor{\bsnm{Baffour}, \binits{F.I.}},
\bauthor{\bsnm{Ringler}, \binits{M.D.}},
\bauthor{\bsnm{Erickson}, \binits{B.J.}}:
\batitle{Quantifying uncertainty in deep learning of radiologic images}.
\bjtitle{Radiology}
\bvolume{308}(\bissue{2}),
\bfpage{222217}
(\byear{2023})
\end{barticle}
\endbibitem

%%% 35
\bibitem[\protect\citeauthoryear{Felbo et~al.}{2017}]{ref70_chain_thaw}
\begin{bchapter}
\bauthor{\bsnm{Felbo}, \binits{B.}},
\bauthor{\bsnm{Mislove}, \binits{A.}},
\bauthor{\bsnm{S{\o}gaard}, \binits{A.}},
\bauthor{\bsnm{Rahwan}, \binits{I.}},
\bauthor{\bsnm{Lehmann}, \binits{S.}}:
\bctitle{Using millions of emoji occurrences to learn any-domain
  representations for detecting sentiment, emotion and sarcasm}.
In: \bbtitle{Proceedings of the 2017 Conference on Empirical Methods in Natural
  Language Processing},
pp. \bfpage{1615}--\blpage{1625}
(\byear{2017})
\end{bchapter}
\endbibitem

%%% 36
\bibitem[\protect\citeauthoryear{Fecso et~al.}{2026}]{fecso2026retfiner}
\begin{bchapter}
\bauthor{\bsnm{Fecso}, \binits{R.}},
\bauthor{\bsnm{Morano}, \binits{J.}},
\bauthor{\bsnm{Schmidt-Erfurth}, \binits{U.}},
\bauthor{\bsnm{Bogunovi\'{c}}, \binits{H.}}:
\bctitle{{RetFiner}: A vision-language refinement scheme for retinal foundation
  models}.
In: \bbtitle{Lecture Notes in Computer Science}
(\byear{2026}).
\doiurl{10.1007/978-3-032-04971-1_51}
\end{bchapter}
\endbibitem

%%% 37
\bibitem[\protect\citeauthoryear{Finlayson
  et~al.}{2021}]{finlayson2021dataset_shift}
\begin{barticle}
\bauthor{\bsnm{Finlayson}, \binits{S.G.}},
\bauthor{\bsnm{Subbaswamy}, \binits{A.}},
\bauthor{\bsnm{Singh}, \binits{K.}},
\bauthor{\bsnm{Bowers}, \binits{J.}},
\bauthor{\bsnm{Kupke}, \binits{A.}},
\bauthor{\bsnm{Zittrain}, \binits{J.}},
\bauthor{\bsnm{Kohane}, \binits{I.S.}},
\bauthor{\bsnm{Saria}, \binits{S.}}:
\batitle{The clinician and dataset shift in artificial intelligence}.
\bjtitle{New England Journal of Medicine}
\bvolume{385}(\bissue{3}),
\bfpage{283}--\blpage{286}
(\byear{2021})
\end{barticle}
\endbibitem

%%% 38
\bibitem[\protect\citeauthoryear{Filiot et~al.}{2026}]{filiot2026robustifying}
\begin{barticle}
\bauthor{\bsnm{Filiot}, \binits{A.}},
\bauthor{\bsnm{Thaeter}, \binits{O.}},
\bauthor{\bsnm{Schmauch}, \binits{B.}},
\bauthor{\bsnm{Guillou}, \binits{L.}}:
\batitle{Robustifying pathology foundation models via fine-tuning}.
\bjtitle{arXiv preprint arXiv:2607.22861}
(\byear{2026})
\doiurl{10.48550/arXiv.2607.22861}
\end{barticle}
\endbibitem

%%% 39
\bibitem[\protect\citeauthoryear{Gu and Dao}{2024}]{gu2024mamba}
\begin{bchapter}
\bauthor{\bsnm{Gu}, \binits{A.}},
\bauthor{\bsnm{Dao}, \binits{T.}}:
\bctitle{Mamba: Linear-time sequence modeling with selective state spaces}.
In: \bbtitle{First Conference on Language Modeling}
(\byear{2024}).
\burl{https://openreview.net/forum?id=tEYskw1VY2}
\end{bchapter}
\endbibitem

%%% 40
\bibitem[\protect\citeauthoryear{Georgescu}{2023}]{georgescu2023masked}
\begin{barticle}
\bauthor{\bsnm{Georgescu}, \binits{M.-I.}}:
\batitle{Masked autoencoders for unsupervised anomaly detection in medical
  images}.
\bjtitle{Procedia Computer Science}
\bvolume{225},
\bfpage{969}--\blpage{978}
(\byear{2023})
\end{barticle}
\endbibitem

%%% 41
\bibitem[\protect\citeauthoryear{Geirhos et~al.}{2020}]{geirhos2020shortcut}
\begin{barticle}
\bauthor{\bsnm{Geirhos}, \binits{R.}},
\bauthor{\bsnm{Jacobsen}, \binits{J.-H.}},
\bauthor{\bsnm{Michaelis}, \binits{C.}},
\bauthor{\bsnm{Zemel}, \binits{R.}},
\bauthor{\bsnm{Brendel}, \binits{W.}},
\bauthor{\bsnm{Bethge}, \binits{M.}},
\bauthor{\bsnm{Wichmann}, \binits{F.A.}}:
\batitle{Shortcut learning in deep neural networks}.
\bjtitle{Nature Machine Intelligence}
\bvolume{2}(\bissue{11}),
\bfpage{665}--\blpage{673}
(\byear{2020})
\end{barticle}
\endbibitem

%%% 42
\bibitem[\protect\citeauthoryear{Glocker et~al.}{2023}]{glocker2023risk}
\begin{barticle}
\bauthor{\bsnm{Glocker}, \binits{B.}},
\bauthor{\bsnm{Jones}, \binits{C.}},
\bauthor{\bsnm{Roschewitz}, \binits{M.}},
\bauthor{\bsnm{Winzeck}, \binits{S.}}:
\batitle{Risk of bias in chest radiography deep learning foundation models}.
\bjtitle{Radiology: Artificial Intelligence}
\bvolume{5}(\bissue{6}),
\bfpage{230060}
(\byear{2023})
\end{barticle}
\endbibitem

%%% 43
\bibitem[\protect\citeauthoryear{Guo et~al.}{2026}]{guo2026impact}
\begin{botherref}
\oauthor{\bsnm{Guo}, \binits{B.}},
\oauthor{\bsnm{Lu}, \binits{D.}},
\oauthor{\bsnm{Szumel}, \binits{G.}},
\oauthor{\bsnm{Gui}, \binits{R.}},
\oauthor{\bsnm{Li}, \binits{P.}},
\oauthor{\bsnm{Wang}, \binits{T.}},
\oauthor{\bsnm{Konz}, \binits{N.}},
\oauthor{\bsnm{Mazurowski}, \binits{M.A.}}:
The impact of scanner domain shift on deep learning performance in medical
  imaging: an experimental study.
International Journal of Computer Assisted Radiology and Surgery,
1--11
(2026)
\end{botherref}
\endbibitem

%%% 44
\bibitem[\protect\citeauthoryear{Gao and Morel}{2026}]{gao2026medpeftcl}
\begin{bchapter}
\bauthor{\bsnm{Gao}, \binits{Z.}},
\bauthor{\bsnm{Morel}, \binits{P.}}:
\bctitle{{MedPEFT-CL}: Dual-phase parameter-efficient continual learning with
  medical semantic adapter and bidirectional memory consolidation}.
In: \bbtitle{Proceedings of the IEEE/CVF Winter Conference on Applications of
  Computer Vision (WACV)},
pp. \bfpage{7699}--\blpage{7708}
(\byear{2026})
\end{bchapter}
\endbibitem

%%% 45
\bibitem[\protect\citeauthoryear{Guo et~al.}{2017}]{guo2017calibration}
\begin{bchapter}
\bauthor{\bsnm{Guo}, \binits{C.}},
\bauthor{\bsnm{Pleiss}, \binits{G.}},
\bauthor{\bsnm{Sun}, \binits{Y.}},
\bauthor{\bsnm{Weinberger}, \binits{K.Q.}}:
\bctitle{On calibration of modern neural networks}.
In: \bbtitle{International Conference on Machine Learning},
pp. \bfpage{1321}--\blpage{1330}
(\byear{2017}).
\bcomment{PMLR}
\end{bchapter}
\endbibitem

%%% 46
\bibitem[\protect\citeauthoryear{Giuffr{\`e} and
  Shung}{2023}]{giuffre2023harnessing}
\begin{barticle}
\bauthor{\bsnm{Giuffr{\`e}}, \binits{M.}},
\bauthor{\bsnm{Shung}, \binits{D.L.}}:
\batitle{Harnessing the power of synthetic data in healthcare: innovation,
  application, and privacy}.
\bjtitle{NPJ digital medicine}
\bvolume{6}(\bissue{1}),
\bfpage{186}
(\byear{2023})
\end{barticle}
\endbibitem

%%% 47
\bibitem[\protect\citeauthoryear{Gong et~al.}{2024}]{ref130_3dsam_adapter}
\begin{barticle}
\bauthor{\bsnm{Gong}, \binits{S.}},
\bauthor{\bsnm{Zhong}, \binits{Y.}},
\bauthor{\bsnm{Ma}, \binits{W.}},
\bauthor{\bsnm{Li}, \binits{J.}},
\bauthor{\bsnm{Wang}, \binits{Z.}},
\bauthor{\bsnm{Zhang}, \binits{J.}},
\bauthor{\bsnm{Heng}, \binits{P.-A.}},
\bauthor{\bsnm{Dou}, \binits{Q.}}:
\batitle{3dsam-adapter: Holistic adaptation of sam from 2d to 3d for promptable
  tumor segmentation}.
\bjtitle{Medical Image Analysis}
\bvolume{98},
\bfpage{103324}
(\byear{2024})
\end{barticle}
\endbibitem

%%% 48
\bibitem[\protect\citeauthoryear{Gao et~al.}{2024}]{gao2024pmt}
\begin{bchapter}
\bauthor{\bsnm{Gao}, \binits{N.}},
\bauthor{\bsnm{Zhou}, \binits{S.}},
\bauthor{\bsnm{Wang}, \binits{L.}},
\bauthor{\bsnm{Zheng}, \binits{N.}}:
\bctitle{Pmt: Progressive mean teacher via exploring temporal consistency for
  semi-supervised medical image segmentation}.
In: \bbtitle{European Conference on Computer Vision},
pp. \bfpage{144}--\blpage{160}
(\byear{2024}).
\bcomment{Springer}
\end{bchapter}
\endbibitem

%%% 49
\bibitem[\protect\citeauthoryear{Hou et~al.}{2026}]{hou2026memsam25d}
\begin{barticle}
\bauthor{\bsnm{Hou}, \binits{Y.}},
\bauthor{\bsnm{Chen}, \binits{N.}},
\bauthor{\bsnm{Huo}, \binits{T.}},
\bauthor{\bsnm{Wang}, \binits{W.}}:
\batitle{{MemSAM-2.5D}: Overcoming volumetric discontinuity and boundary
  ambiguity for 3d liver tumor segmentation}.
\bjtitle{Frontiers in Oncology}
\bvolume{16},
\bfpage{1850004}
(\byear{2026})
\doiurl{10.3389/fonc.2026.1850004}
\end{barticle}
\endbibitem

%%% 50
\bibitem[\protect\citeauthoryear{He et~al.}{2022}]{ref51_mae}
\begin{bchapter}
\bauthor{\bsnm{He}, \binits{K.}},
\bauthor{\bsnm{Chen}, \binits{X.}},
\bauthor{\bsnm{Xie}, \binits{S.}},
\bauthor{\bsnm{Li}, \binits{Y.}},
\bauthor{\bsnm{Doll{\'a}r}, \binits{P.}},
\bauthor{\bsnm{Girshick}, \binits{R.}}:
\bctitle{Masked autoencoders are scalable vision learners}.
In: \bbtitle{Proceedings of the IEEE/CVF Conference on Computer Vision and
  Pattern Recognition},
pp. \bfpage{16000}--\blpage{16009}
(\byear{2022})
\end{bchapter}
\endbibitem

%%% 51
\bibitem[\protect\citeauthoryear{Huix et~al.}{2024}]{huix2024naturaldomain}
\begin{bchapter}
\bauthor{\bsnm{Huix}, \binits{J.P.}},
\bauthor{\bsnm{Ganeshan}, \binits{A.R.}},
\bauthor{\bsnm{Haslum}, \binits{J.F.}},
\bauthor{\bsnm{S\"oderberg}, \binits{M.}},
\bauthor{\bsnm{Matsoukas}, \binits{C.}},
\bauthor{\bsnm{Smith}, \binits{K.}}:
\bctitle{Are natural domain foundation models useful for medical image
  classification?}
In: \bbtitle{Proceedings of the IEEE/CVF Winter Conference on Applications of
  Computer Vision},
pp. \bfpage{7634}--\blpage{7643}
(\byear{2024})
\end{bchapter}
\endbibitem

%%% 52
\bibitem[\protect\citeauthoryear{Houlsby et~al.}{2019}]{ref82_standard_adapter}
\begin{bchapter}
\bauthor{\bsnm{Houlsby}, \binits{N.}},
\bauthor{\bsnm{Giurgiu}, \binits{A.}},
\bauthor{\bsnm{Jastrzebski}, \binits{S.}},
\bauthor{\bsnm{Morrone}, \binits{B.}},
\bauthor{\bsnm{De~Laroussilhe}, \binits{Q.}},
\bauthor{\bsnm{Gesmundo}, \binits{A.}},
\bauthor{\bsnm{Attariyan}, \binits{M.}},
\bauthor{\bsnm{Gelly}, \binits{S.}}:
\bctitle{Parameter-efficient transfer learning for nlp}.
In: \bbtitle{International Conference on Machine Learning},
pp. \bfpage{2790}--\blpage{2799}
(\byear{2019}).
\bcomment{PMLR}
\end{bchapter}
\endbibitem

%%% 53
\bibitem[\protect\citeauthoryear{Han et~al.}{2024}]{ref95_peft_survey}
\begin{botherref}
\oauthor{\bsnm{Han}, \binits{Z.}},
\oauthor{\bsnm{Gao}, \binits{C.}},
\oauthor{\bsnm{Liu}, \binits{J.}},
\oauthor{\bsnm{Zhang}, \binits{J.}},
\oauthor{\bsnm{Zhang}, \binits{S.Q.}}:
Parameter-efficient fine-tuning for large models: A comprehensive survey.
arXiv preprint arXiv:2403.14608
(2024)
\end{botherref}
\endbibitem

%%% 54
\bibitem[\protect\citeauthoryear{He et~al.}{2025}]{ref121_vista3d}
\begin{bchapter}
\bauthor{\bsnm{He}, \binits{Y.}},
\bauthor{\bsnm{Guo}, \binits{P.}},
\bauthor{\bsnm{Tang}, \binits{Y.}},
\bauthor{\bsnm{Myronenko}, \binits{A.}},
\bauthor{\bsnm{Nath}, \binits{V.}},
\bauthor{\bsnm{Xu}, \binits{Z.}},
\bauthor{\bsnm{Yang}, \binits{D.}},
\bauthor{\bsnm{Zhao}, \binits{C.}},
\bauthor{\bsnm{Simon}, \binits{B.}},
\bauthor{\bsnm{Belue}, \binits{M.}},
\bauthor{\bsnm{Harmon}, \binits{S.}},
\bauthor{\bsnm{Turkbey}, \binits{B.}},
\bauthor{\bsnm{Xu}, \binits{D.}},
\bauthor{\bsnm{Li}, \binits{W.}}:
\bctitle{Vista3d: A unified segmentation foundation model for 3d medical
  imaging}.
In: \bbtitle{Proceedings of the Computer Vision and Pattern Recognition
  Conference (CVPR)},
pp. \bfpage{20863}--\blpage{20873}
(\byear{2025})
\end{bchapter}
\endbibitem

%%% 55
\bibitem[\protect\citeauthoryear{He
  et~al.}{2025}]{ref35_fm_advancing_healthcare}
\begin{barticle}
\bauthor{\bsnm{He}, \binits{Y.}},
\bauthor{\bsnm{Huang}, \binits{F.}},
\bauthor{\bsnm{Jiang}, \binits{X.}},
\bauthor{\bsnm{Nie}, \binits{Y.}},
\bauthor{\bsnm{Wang}, \binits{M.}},
\bauthor{\bsnm{Wang}, \binits{J.}},
\bauthor{\bsnm{Chen}, \binits{H.}}:
\batitle{Foundation model for advancing healthcare: Challenges, opportunities
  and future directions}.
\bjtitle{IEEE Reviews in Biomedical Engineering}
\bvolume{18},
\bfpage{172}--\blpage{191}
(\byear{2025})
\doiurl{10.1109/RBME.2024.3496744}
\end{barticle}
\endbibitem

%%% 56
\bibitem[\protect\citeauthoryear{Huang et~al.}{2024}]{ref145_mvfa}
\begin{bchapter}
\bauthor{\bsnm{Huang}, \binits{C.}},
\bauthor{\bsnm{Jiang}, \binits{A.}},
\bauthor{\bsnm{Feng}, \binits{J.}},
\bauthor{\bsnm{Zhang}, \binits{Y.}},
\bauthor{\bsnm{Wang}, \binits{X.}},
\bauthor{\bsnm{Wang}, \binits{Y.}}:
\bctitle{Adapting visual-language models for generalizable anomaly detection in
  medical images}.
In: \bbtitle{Proceedings of the IEEE/CVF Conference on Computer Vision and
  Pattern Recognition (CVPR)},
pp. \bfpage{11375}--\blpage{11385}
(\byear{2024})
\end{bchapter}
\endbibitem

%%% 57
\bibitem[\protect\citeauthoryear{Hu et~al.}{2025}]{hu2025better}
\begin{barticle}
\bauthor{\bsnm{Hu}, \binits{L.}},
\bauthor{\bsnm{Meng}, \binits{Z.}},
\bauthor{\bsnm{Tan}, \binits{C.}},
\bauthor{\bsnm{Zhou}, \binits{Y.}}:
\batitle{Better pseudo-labeling for semi-supervised domain generalization in
  medical magnetic resonance image segmentation}.
\bjtitle{International Journal of Computational Intelligence Systems}
\bvolume{18}(\bissue{1}),
\bfpage{71}
(\byear{2025})
\end{barticle}
\endbibitem

%%% 58
\bibitem[\protect\citeauthoryear{Huang et~al.}{2023}]{huang2023ssl_medical}
\begin{barticle}
\bauthor{\bsnm{Huang}, \binits{S.-C.}},
\bauthor{\bsnm{Pareek}, \binits{A.}},
\bauthor{\bsnm{Jensen}, \binits{M.}},
\bauthor{\bsnm{Lungren}, \binits{M.P.}},
\bauthor{\bsnm{Yeung}, \binits{S.}},
\bauthor{\bsnm{Chaudhari}, \binits{A.S.}}:
\batitle{Self-supervised learning for medical image classification: a
  systematic review and implementation guidelines}.
\bjtitle{NPJ Digital Medicine}
\bvolume{6}(\bissue{1}),
\bfpage{74}
(\byear{2023})
\end{barticle}
\endbibitem

%%% 59
\bibitem[\protect\citeauthoryear{Howard and Ruder}{2018}]{ref69_unilm_ft_text}
\begin{bchapter}
\bauthor{\bsnm{Howard}, \binits{J.}},
\bauthor{\bsnm{Ruder}, \binits{S.}}:
\bctitle{Universal language model fine-tuning for text classification}.
In: \bbtitle{Proceedings of the 56th Annual Meeting of the Association for
  Computational Linguistics (Volume 1: Long Papers)},
pp. \bfpage{328}--\blpage{339}
(\byear{2018})
\end{bchapter}
\endbibitem

%%% 60
\bibitem[\protect\citeauthoryear{Huang et~al.}{2021}]{ref125_gloria}
\begin{bchapter}
\bauthor{\bsnm{Huang}, \binits{S.-C.}},
\bauthor{\bsnm{Shen}, \binits{L.}},
\bauthor{\bsnm{Lungren}, \binits{M.P.}},
\bauthor{\bsnm{Yeung}, \binits{S.}}:
\bctitle{Gloria: A multimodal global-local representation learning framework
  for label-efficient medical image recognition}.
In: \bbtitle{Proceedings of the IEEE/CVF International Conference on Computer
  Vision},
pp. \bfpage{3942}--\blpage{3951}
(\byear{2021})
\end{bchapter}
\endbibitem

%%% 61
\bibitem[\protect\citeauthoryear{Hu et~al.}{2022}]{ref100_lora}
\begin{bchapter}
\bauthor{\bsnm{Hu}, \binits{E.J.}},
\bauthor{\bsnm{Shen}, \binits{Y.}},
\bauthor{\bsnm{Wallis}, \binits{P.}},
\bauthor{\bsnm{Allen-Zhu}, \binits{Z.}},
\bauthor{\bsnm{Li}, \binits{Y.}},
\bauthor{\bsnm{Wang}, \binits{S.}},
\bauthor{\bsnm{Wang}, \binits{L.}},
\bauthor{\bsnm{Chen}, \binits{W.}}:
\bctitle{Lora: Low-rank adaptation of large language models}.
In: \bbtitle{ICLR 2022}
(\byear{2022})
\end{bchapter}
\endbibitem

%%% 62
\bibitem[\protect\citeauthoryear{He et~al.}{2025}]{ref89_prompt_tuning}
\begin{barticle}
\bauthor{\bsnm{He}, \binits{A.}},
\bauthor{\bsnm{Wu}, \binits{Y.}},
\bauthor{\bsnm{Wang}, \binits{Z.}},
\bauthor{\bsnm{Li}, \binits{T.}},
\bauthor{\bsnm{Fu}, \binits{H.}}:
\batitle{Dvpt: Dynamic visual prompt tuning of large pre-trained models for
  medical image analysis}.
\bjtitle{Neural Networks}
\bvolume{185},
\bfpage{107168}
(\byear{2025})
\doiurl{10.1016/j.neunet.2025.107168}
\end{barticle}
\endbibitem

%%% 63
\bibitem[\protect\citeauthoryear{Huang et~al.}{2025}]{huang2025survey}
\begin{barticle}
\bauthor{\bsnm{Huang}, \binits{L.}},
\bauthor{\bsnm{Yu}, \binits{W.}},
\bauthor{\bsnm{Ma}, \binits{W.}},
\bauthor{\bsnm{Zhong}, \binits{W.}},
\bauthor{\bsnm{Feng}, \binits{Z.}},
\bauthor{\bsnm{Wang}, \binits{H.}},
\bauthor{\bsnm{Chen}, \binits{Q.}},
\bauthor{\bsnm{Peng}, \binits{W.}},
\bauthor{\bsnm{Feng}, \binits{X.}},
\bauthor{\bsnm{Qin}, \binits{B.}}, \betal:
\batitle{A survey on hallucination in large language models: Principles,
  taxonomy, challenges, and open questions}.
\bjtitle{ACM Transactions on Information Systems}
\bvolume{43}(\bissue{2}),
\bfpage{1}--\blpage{55}
(\byear{2025})
\end{barticle}
\endbibitem

%%% 64
\bibitem[\protect\citeauthoryear{{International Medical Device Regulators Forum
  (IMDRF)}}{2017}]{imdrf_samd_clinical_2017}
\begin{botherref}
\oauthor{\bsnm{{International Medical Device Regulators Forum (IMDRF)}}}:
Software as a medical device (samd): Clinical evaluation.
Technical Report IMDRF/SaMD WG/N41FINAL:2017,
IMDRF
(September 2017).
Final document, 21 September 2017.
\url{https://www.imdrf.org}
\end{botherref}
\endbibitem

%%% 65
\bibitem[\protect\citeauthoryear{{International Medical Device Regulators Forum
  (IMDRF)}}{2025}]{imdrf_gmlp_2025}
\begin{botherref}
\oauthor{\bsnm{{International Medical Device Regulators Forum (IMDRF)}}}:
Good machine learning practice for medical device development: Guiding
  principles.
Technical Report IMDRF/AIML WG/N88 FINAL:2025,
IMDRF
(January 2025).
Final document, published 29 January 2025; accessed 16 August 2026.
\url{https://www.imdrf.org/documents/good-machine-learning-practice-medical-device-development-guiding-principles}
\end{botherref}
\endbibitem

%%% 66
\bibitem[\protect\citeauthoryear{Irvin et~al.}{2019}]{irvin2019chexpert}
\begin{bchapter}
\bauthor{\bsnm{Irvin}, \binits{J.}},
\bauthor{\bsnm{Rajpurkar}, \binits{P.}},
\bauthor{\bsnm{Ko}, \binits{M.}},
\bauthor{\bsnm{Yu}, \binits{Y.}},
\bauthor{\bsnm{Ciurea-Ilcus}, \binits{S.}},
\bauthor{\bsnm{Chute}, \binits{C.}},
\bauthor{\bsnm{Marklund}, \binits{H.}},
\bauthor{\bsnm{Haghgoo}, \binits{B.}},
\bauthor{\bsnm{Ball}, \binits{R.}},
\bauthor{\bsnm{Shpanskaya}, \binits{K.}}, \betal:
\bctitle{Chexpert: A large chest radiograph dataset with uncertainty labels and
  expert comparison}.
In: \bbtitle{Proceedings of the AAAI Conference on Artificial Intelligence},
vol. \bseriesno{33},
pp. \bfpage{590}--\blpage{597}
(\byear{2019})
\end{bchapter}
\endbibitem

%%% 67
\bibitem[\protect\citeauthoryear{Jungo et~al.}{2018}]{jungo2018effect}
\begin{bchapter}
\bauthor{\bsnm{Jungo}, \binits{A.}},
\bauthor{\bsnm{Meier}, \binits{R.}},
\bauthor{\bsnm{Ermis}, \binits{E.}},
\bauthor{\bsnm{Blatti-Moreno}, \binits{M.}},
\bauthor{\bsnm{Herrmann}, \binits{E.}},
\bauthor{\bsnm{Wiest}, \binits{R.}},
\bauthor{\bsnm{Reyes}, \binits{M.}}:
\bctitle{On the effect of inter-observer variability for a reliable estimation
  of uncertainty of medical image segmentation}.
In: \bbtitle{International Conference on Medical Image Computing and
  Computer-assisted Intervention},
pp. \bfpage{682}--\blpage{690}
(\byear{2018}).
\bcomment{Springer}
\end{bchapter}
\endbibitem

%%% 68
\bibitem[\protect\citeauthoryear{Javed et~al.}{2024}]{javed2024cplip}
\begin{bchapter}
\bauthor{\bsnm{Javed}, \binits{S.}},
\bauthor{\bsnm{Mahmood}, \binits{A.}},
\bauthor{\bsnm{Ganapathi}, \binits{I.I.}},
\bauthor{\bsnm{Dharejo}, \binits{F.A.}},
\bauthor{\bsnm{Werghi}, \binits{N.}},
\bauthor{\bsnm{Bennamoun}, \binits{M.}}:
\bctitle{Cplip: Zero-shot learning for histopathology with comprehensive
  vision-language alignment}.
In: \bbtitle{Proceedings of the IEEE/CVF Conference on Computer Vision and
  Pattern Recognition},
pp. \bfpage{11450}--\blpage{11459}
(\byear{2024})
\end{bchapter}
\endbibitem

%%% 69
\bibitem[\protect\citeauthoryear{Jim{\'e}nez-S{\'a}nchez
  et~al.}{2024}]{jimenez2024copycats}
\begin{barticle}
\bauthor{\bsnm{Jim{\'e}nez-S{\'a}nchez}, \binits{A.}},
\bauthor{\bsnm{Avlona}, \binits{N.-R.}},
\bauthor{\bsnm{Juodelyte}, \binits{D.}},
\bauthor{\bsnm{Sourget}, \binits{T.}},
\bauthor{\bsnm{Vang-Larsen}, \binits{C.}},
\bauthor{\bsnm{Rogers}, \binits{A.}},
\bauthor{\bsnm{Zajac}, \binits{H.}},
\bauthor{\bsnm{Cheplygina}, \binits{V.}}:
\batitle{Copycats: the many lives of a publicly available medical imaging
  dataset}.
\bjtitle{Advances in Neural Information Processing Systems}
\bvolume{37},
\bfpage{113383}--\blpage{113404}
(\byear{2024})
\end{barticle}
\endbibitem

%%% 70
\bibitem[\protect\citeauthoryear{Jain et~al.}{2021}]{jain2021visualchexbert}
\begin{bchapter}
\bauthor{\bsnm{Jain}, \binits{S.}},
\bauthor{\bsnm{Smit}, \binits{A.}},
\bauthor{\bsnm{Truong}, \binits{S.Q.}},
\bauthor{\bsnm{Nguyen}, \binits{C.D.}},
\bauthor{\bsnm{Huynh}, \binits{M.-T.}},
\bauthor{\bsnm{Jain}, \binits{M.}},
\bauthor{\bsnm{Young}, \binits{V.A.}},
\bauthor{\bsnm{Ng}, \binits{A.Y.}},
\bauthor{\bsnm{Lungren}, \binits{M.P.}},
\bauthor{\bsnm{Rajpurkar}, \binits{P.}}:
\bctitle{Visualchexbert: addressing the discrepancy between radiology report
  labels and image labels}.
In: \bbtitle{Proceedings of the Conference on Health, Inference, and Learning},
pp. \bfpage{105}--\blpage{115}
(\byear{2021})
\end{bchapter}
\endbibitem

%%% 71
\bibitem[\protect\citeauthoryear{Jiao et~al.}{2025}]{jiao2025foundation}
\begin{botherref}
\oauthor{\bsnm{Jiao}, \binits{L.}},
\oauthor{\bsnm{Yang}, \binits{J.}},
\oauthor{\bsnm{Li}, \binits{R.}},
\oauthor{\bsnm{Liu}, \binits{F.}},
\oauthor{\bsnm{Liu}, \binits{X.}},
\oauthor{\bsnm{Chen}, \binits{P.}},
\oauthor{\bsnm{Guo}, \binits{Y.}},
\oauthor{\bsnm{Li}, \binits{L.}},
\oauthor{\bsnm{Shang}, \binits{R.}},
\oauthor{\bsnm{Ma}, \binits{W.}}:
Foundation model for medical imaging: A comprehensive review.
IEEE Transactions on Artificial Intelligence
(2025)
\end{botherref}
\endbibitem

%%% 72
\bibitem[\protect\citeauthoryear{Jiang et~al.}{2025}]{jiang2025diffugts}
\begin{bchapter}
\bauthor{\bsnm{Jiang}, \binits{Y.}},
\bauthor{\bsnm{Zhang}, \binits{P.}},
\bauthor{\bsnm{Yang}, \binits{D.}},
\bauthor{\bsnm{Tian}, \binits{Y.}},
\bauthor{\bsnm{Lin}, \binits{H.}},
\bauthor{\bsnm{Wang}, \binits{X.}}:
\bctitle{Advancing generalizable tumor segmentation with anomaly-aware
  open-vocabulary attention maps and frozen foundation diffusion models}.
In: \bbtitle{Proceedings of the IEEE/CVF Conference on Computer Vision and
  Pattern Recognition}
(\byear{2025}).
\doiurl{10.1109/CVPR52734.2025.02419}
\end{bchapter}
\endbibitem

%%% 73
\bibitem[\protect\citeauthoryear{Kore et~al.}{2024}]{kore2024empirical}
\begin{barticle}
\bauthor{\bsnm{Kore}, \binits{A.}},
\bauthor{\bsnm{Abbasi~Bavil}, \binits{E.}},
\bauthor{\bsnm{Subasri}, \binits{V.}},
\bauthor{\bsnm{Abdalla}, \binits{M.}},
\bauthor{\bsnm{Fine}, \binits{B.}},
\bauthor{\bsnm{Dolatabadi}, \binits{E.}},
\bauthor{\bsnm{Abdalla}, \binits{M.}}:
\batitle{Empirical data drift detection experiments on real-world medical
  imaging data}.
\bjtitle{Nature communications}
\bvolume{15}(\bissue{1}),
\bfpage{1887}
(\byear{2024})
\end{barticle}
\endbibitem

%%% 74
\bibitem[\protect\citeauthoryear{Kervadec et~al.}{2021}]{kervadec19a}
\begin{barticle}
\bauthor{\bsnm{Kervadec}, \binits{H.}},
\bauthor{\bsnm{Bouchtiba}, \binits{J.}},
\bauthor{\bsnm{Desrosiers}, \binits{C.}},
\bauthor{\bsnm{Granger}, \binits{E.}},
\bauthor{\bsnm{Dolz}, \binits{J.}},
\bauthor{\bsnm{Ayed}, \binits{I.B.}}:
\batitle{Boundary loss for highly unbalanced segmentation}.
\bjtitle{Medical image analysis}
\bvolume{67},
\bfpage{101851}
(\byear{2021})
\end{barticle}
\endbibitem

%%% 75
\bibitem[\protect\citeauthoryear{Kumari et~al.}{2025}]{kumari2025continual}
\begin{botherref}
\oauthor{\bsnm{Kumari}, \binits{P.}},
\oauthor{\bsnm{Chauhan}, \binits{J.}},
\oauthor{\bsnm{Bozorgpour}, \binits{A.}},
\oauthor{\bsnm{Huang}, \binits{B.}},
\oauthor{\bsnm{Azad}, \binits{R.}},
\oauthor{\bsnm{Merhof}, \binits{D.}}:
Continual learning in medical image analysis: A comprehensive review of recent
  advancements and future prospects.
Medical Image Analysis,
103730
(2025)
\end{botherref}
\endbibitem

%%% 76
\bibitem[\protect\citeauthoryear{K{\"o}men et~al.}{2026}]{komen2026pathorob}
\begin{barticle}
\bauthor{\bsnm{K{\"o}men}, \binits{J.}},
\bauthor{\bsnm{Jong}, \binits{E.D.}},
\bauthor{\bsnm{Hense}, \binits{J.}},
\bauthor{\bsnm{Marienwald}, \binits{H.}},
\bauthor{\bsnm{Dippel}, \binits{J.}},
\bauthor{\bsnm{Naumann}, \binits{P.}},
\bauthor{\bsnm{Marcus}, \binits{E.}},
\bauthor{\bsnm{Ruff}, \binits{L.}},
\bauthor{\bsnm{Alber}, \binits{M.}},
\bauthor{\bsnm{Teuwen}, \binits{J.}},
\bauthor{\bsnm{Klauschen}, \binits{F.}},
\bauthor{\bsnm{M{\"u}ller}, \binits{K.-R.}}:
\batitle{Towards robust foundation models for digital pathology}.
\bjtitle{Nature Communications}
\bvolume{17},
\bfpage{5218}
(\byear{2026})
\doiurl{10.1038/s41467-026-73923-2}
\end{barticle}
\endbibitem

%%% 77
\bibitem[\protect\citeauthoryear{Kim et~al.}{2024}]{kim2024monet}
\begin{barticle}
\bauthor{\bsnm{Kim}, \binits{C.}},
\bauthor{\bsnm{Gadgil}, \binits{S.U.}},
\bauthor{\bsnm{DeGrave}, \binits{A.J.}},
\bauthor{\bsnm{Omiye}, \binits{J.A.}},
\bauthor{\bsnm{Cai}, \binits{Z.R.}},
\bauthor{\bsnm{Daneshjou}, \binits{R.}},
\bauthor{\bsnm{Lee}, \binits{S.-I.}}:
\batitle{Transparent medical image ai via an image--text foundation model
  grounded in medical literature}.
\bjtitle{Nature medicine}
\bvolume{30}(\bissue{4}),
\bfpage{1154}--\blpage{1165}
(\byear{2024})
\end{barticle}
\endbibitem

%%% 78
\bibitem[\protect\citeauthoryear{Kim et~al.}{2025a}]{kim2025medivista}
\begin{barticle}
\bauthor{\bsnm{Kim}, \binits{S.}},
\bauthor{\bsnm{Jin}, \binits{P.}},
\bauthor{\bsnm{Chen}, \binits{C.}},
\bauthor{\bsnm{Kim}, \binits{K.}},
\bauthor{\bsnm{Lyu}, \binits{Z.}},
\bauthor{\bsnm{Ren}, \binits{H.}},
\bauthor{\bsnm{Kim}, \binits{S.}},
\bauthor{\bsnm{Liu}, \binits{Z.}},
\bauthor{\bsnm{Zhong}, \binits{A.}},
\bauthor{\bsnm{Liu}, \binits{T.}},
\bauthor{\bsnm{Li}, \binits{X.}},
\bauthor{\bsnm{Li}, \binits{Q.}}:
\batitle{{MediViSTA}: Medical video segmentation via temporal fusion {SAM}
  adaptation for echocardiography}.
\bjtitle{IEEE Journal of Biomedical and Health Informatics}
\bvolume{29}(\bissue{12}),
\bfpage{8954}--\blpage{8964}
(\byear{2025})
\doiurl{10.1109/JBHI.2025.3540306}
{\href{https://arxiv.org/abs/2309.13539}{{arXiv:2309.13539}}}
\end{barticle}
\endbibitem

%%% 79
\bibitem[\protect\citeauthoryear{Kim et~al.}{2025b}]{kim2025medical}
\begin{botherref}
\oauthor{\bsnm{Kim}, \binits{Y.}},
\oauthor{\bsnm{Jeong}, \binits{H.}},
\oauthor{\bsnm{Chen}, \binits{S.}},
\oauthor{\bsnm{Li}, \binits{S.S.}},
\oauthor{\bsnm{Park}, \binits{C.}},
\oauthor{\bsnm{Lu}, \binits{M.}},
\oauthor{\bsnm{Alhamoud}, \binits{K.}},
\oauthor{\bsnm{Mun}, \binits{J.}},
\oauthor{\bsnm{Grau}, \binits{C.}},
\oauthor{\bsnm{Jung}, \binits{M.}}, et al.:
Medical hallucinations in foundation models and their impact on healthcare.
arXiv preprint arXiv:2503.05777
(2025)
\end{botherref}
\endbibitem

%%% 80
\bibitem[\protect\citeauthoryear{Kelly et~al.}{2019}]{kelly2019key}
\begin{barticle}
\bauthor{\bsnm{Kelly}, \binits{C.J.}},
\bauthor{\bsnm{Karthikesalingam}, \binits{A.}},
\bauthor{\bsnm{Suleyman}, \binits{M.}},
\bauthor{\bsnm{Corrado}, \binits{G.}},
\bauthor{\bsnm{King}, \binits{D.}}:
\batitle{Key challenges for delivering clinical impact with artificial
  intelligence}.
\bjtitle{BMC medicine}
\bvolume{17}(\bissue{1}),
\bfpage{195}
(\byear{2019})
\end{barticle}
\endbibitem

%%% 81
\bibitem[\protect\citeauthoryear{Khan
  et~al.}{2025}]{ref33_comprehensive_survey_fm}
\begin{botherref}
\oauthor{\bsnm{Khan}, \binits{W.}},
\oauthor{\bsnm{Leem}, \binits{S.}},
\oauthor{\bsnm{See}, \binits{K.B.}},
\oauthor{\bsnm{Wong}, \binits{J.K.}},
\oauthor{\bsnm{Zhang}, \binits{S.}},
\oauthor{\bsnm{Fang}, \binits{R.}}:
A comprehensive survey of foundation models in medicine.
IEEE Reviews in Biomedical Engineering,
1--22
(2025)
\doiurl{10.1109/RBME.2025.3531360}
\end{botherref}
\endbibitem

%%% 82
\bibitem[\protect\citeauthoryear{Kirillov et~al.}{2023}]{ref21_sam}
\begin{bchapter}
\bauthor{\bsnm{Kirillov}, \binits{A.}},
\bauthor{\bsnm{Mintun}, \binits{E.}},
\bauthor{\bsnm{Ravi}, \binits{N.}},
\bauthor{\bsnm{Mao}, \binits{H.}},
\bauthor{\bsnm{Rolland}, \binits{C.}},
\bauthor{\bsnm{Gustafson}, \binits{L.}},
\bauthor{\bsnm{Xiao}, \binits{T.}},
\bauthor{\bsnm{Whitehead}, \binits{S.}},
\bauthor{\bsnm{Berg}, \binits{A.C.}},
\bauthor{\bsnm{Lo}, \binits{W.-Y.}},
\bauthor{\bsnm{Dollar}, \binits{P.}},
\bauthor{\bsnm{Girshick}, \binits{R.}}:
\bctitle{Segment anything}.
In: \bbtitle{Proceedings of the IEEE/CVF International Conference on Computer
  Vision (ICCV)},
pp. \bfpage{4015}--\blpage{4026}
(\byear{2023})
\end{bchapter}
\endbibitem

%%% 83
\bibitem[\protect\citeauthoryear{Kim et~al.}{2025}]{kim2025comedsam}
\begin{barticle}
\bauthor{\bsnm{Kim}, \binits{M.}},
\bauthor{\bsnm{Ryu}, \binits{K.}},
\bauthor{\bsnm{Han}, \binits{Y.}}:
\batitle{{CoMed-SAM}: Collaborative medical {SAM} for multi-modality image
  segmentation}.
\bjtitle{IEEE Access}
\bvolume{13},
\bfpage{184561}--\blpage{184573}
(\byear{2025})
\doiurl{10.1109/ACCESS.2025.3626037}
\end{barticle}
\endbibitem

%%% 84
\bibitem[\protect\citeauthoryear{Kumar et~al.}{2022}]{ref64_ft_distor_feat}
\begin{botherref}
\oauthor{\bsnm{Kumar}, \binits{A.}},
\oauthor{\bsnm{Raghunathan}, \binits{A.}},
\oauthor{\bsnm{Jones}, \binits{R.}},
\oauthor{\bsnm{Ma}, \binits{T.}},
\oauthor{\bsnm{Liang}, \binits{P.}}:
Fine-tuning can distort pretrained features and underperform
  out-of-distribution.
arXiv preprint arXiv:2202.10054
(2022)
\end{botherref}
\endbibitem

%%% 85
\bibitem[\protect\citeauthoryear{Karimi and
  Salcudean}{2019}]{karimi2019reducing}
\begin{barticle}
\bauthor{\bsnm{Karimi}, \binits{D.}},
\bauthor{\bsnm{Salcudean}, \binits{S.E.}}:
\batitle{Reducing the hausdorff distance in medical image segmentation with
  convolutional neural networks}.
\bjtitle{IEEE Transactions on medical imaging}
\bvolume{39}(\bissue{2}),
\bfpage{499}--\blpage{513}
(\byear{2019})
\end{barticle}
\endbibitem

%%% 86
\bibitem[\protect\citeauthoryear{Koetzier
  et~al.}{2024}]{koetzier2024generating}
\begin{barticle}
\bauthor{\bsnm{Koetzier}, \binits{L.R.}},
\bauthor{\bsnm{Wu}, \binits{J.}},
\bauthor{\bsnm{Mastrodicasa}, \binits{D.}},
\bauthor{\bsnm{Lutz}, \binits{A.}},
\bauthor{\bsnm{Chung}, \binits{M.}},
\bauthor{\bsnm{Koszek}, \binits{W.A.}},
\bauthor{\bsnm{Pratap}, \binits{J.}},
\bauthor{\bsnm{Chaudhari}, \binits{A.S.}},
\bauthor{\bsnm{Rajpurkar}, \binits{P.}},
\bauthor{\bsnm{Lungren}, \binits{M.P.}}, \betal:
\batitle{Generating synthetic data for medical imaging}.
\bjtitle{Radiology}
\bvolume{312}(\bissue{3}),
\bfpage{232471}
(\byear{2024})
\end{barticle}
\endbibitem

%%% 87
\bibitem[\protect\citeauthoryear{Kang et~al.}{2019}]{kang2019decoupling}
\begin{botherref}
\oauthor{\bsnm{Kang}, \binits{B.}},
\oauthor{\bsnm{Xie}, \binits{S.}},
\oauthor{\bsnm{Rohrbach}, \binits{M.}},
\oauthor{\bsnm{Yan}, \binits{Z.}},
\oauthor{\bsnm{Gordo}, \binits{A.}},
\oauthor{\bsnm{Feng}, \binits{J.}},
\oauthor{\bsnm{Kalantidis}, \binits{Y.}}:
Decoupling representation and classifier for long-tailed recognition.
arXiv preprint arXiv:1910.09217
(2019)
\end{botherref}
\endbibitem

%%% 88
\bibitem[\protect\citeauthoryear{Koutsoubis
  et~al.}{2024}]{koutsoubis2024privacy}
\begin{botherref}
\oauthor{\bsnm{Koutsoubis}, \binits{N.}},
\oauthor{\bsnm{Yilmaz}, \binits{Y.}},
\oauthor{\bsnm{Ramachandran}, \binits{R.P.}},
\oauthor{\bsnm{Schabath}, \binits{M.}},
\oauthor{\bsnm{Rasool}, \binits{G.}}:
Privacy preserving federated learning in medical imaging with uncertainty
  estimation.
arXiv preprint arXiv:2406.12815
(2024)
\end{botherref}
\endbibitem

%%% 89
\bibitem[\protect\citeauthoryear{Lu et~al.}{2024}]{ref111_villan_fm}
\begin{barticle}
\bauthor{\bsnm{Lu}, \binits{M.Y.}},
\bauthor{\bsnm{Chen}, \binits{B.}},
\bauthor{\bsnm{Williamson}, \binits{D.F.}},
\bauthor{\bsnm{Chen}, \binits{R.J.}},
\bauthor{\bsnm{Liang}, \binits{I.}},
\bauthor{\bsnm{Ding}, \binits{T.}},
\bauthor{\bsnm{Jaume}, \binits{G.}},
\bauthor{\bsnm{Odintsov}, \binits{I.}},
\bauthor{\bsnm{Le}, \binits{L.P.}},
\bauthor{\bsnm{Gerber}, \binits{G.}}, \betal:
\batitle{A visual-language foundation model for computational pathology}.
\bjtitle{Nature Medicine}
\bvolume{30}(\bissue{3}),
\bfpage{863}--\blpage{874}
(\byear{2024})
\doiurl{10.1038/s41591-024-02856-4}
\end{barticle}
\endbibitem

%%% 90
\bibitem[\protect\citeauthoryear{Liu et~al.}{2025}]{liu2025nasvlm}
\begin{barticle}
\bauthor{\bsnm{Liu}, \binits{X.}},
\bauthor{\bsnm{Gong}, \binits{W.}},
\bauthor{\bsnm{Chen}, \binits{X.}},
\bauthor{\bsnm{Li}, \binits{Z.}},
\bauthor{\bsnm{Liu}, \binits{Y.}},
\bauthor{\bsnm{Wang}, \binits{L.}},
\bauthor{\bsnm{Liu}, \binits{Q.}},
\bauthor{\bsnm{Sun}, \binits{X.}},
\bauthor{\bsnm{Liu}, \binits{X.}},
\bauthor{\bsnm{Chen}, \binits{X.}},
\bauthor{\bsnm{Shi}, \binits{Y.}},
\bauthor{\bsnm{Yu}, \binits{H.}}:
\batitle{Vision-language foundation model for generalizable nasal disease
  diagnosis using unlabeled endoscopic records}.
\bjtitle{Pattern Recognition}
\bvolume{165},
\bfpage{111646}
(\byear{2025})
\doiurl{10.1016/j.patcog.2025.111646}
\end{barticle}
\endbibitem

%%% 91
\bibitem[\protect\citeauthoryear{Lin et~al.}{2017}]{lin2017focal}
\begin{bchapter}
\bauthor{\bsnm{Lin}, \binits{T.-Y.}},
\bauthor{\bsnm{Goyal}, \binits{P.}},
\bauthor{\bsnm{Girshick}, \binits{R.}},
\bauthor{\bsnm{He}, \binits{K.}},
\bauthor{\bsnm{Dollar}, \binits{P.}}:
\bctitle{Focal loss for dense object detection}.
In: \bbtitle{Proceedings of the IEEE International Conference on Computer
  Vision (ICCV)}
(\byear{2017})
\end{bchapter}
\endbibitem

%%% 92
\bibitem[\protect\citeauthoryear{Luo et~al.}{2025}]{luo2025diagcot}
\begin{barticle}
\bauthor{\bsnm{Luo}, \binits{Y.}},
\bauthor{\bsnm{He}, \binits{W.}},
\bauthor{\bsnm{Cui}, \binits{Z.-X.}},
\bauthor{\bsnm{Liang}, \binits{D.}}:
\batitle{Teaching {AI} stepwise diagnostic reasoning with report-guided
  chain-of-thought learning}.
\bjtitle{arXiv preprint arXiv:2509.06409}
(\byear{2025})
\doiurl{10.48550/arXiv.2509.06409}
\end{barticle}
\endbibitem

%%% 93
\bibitem[\protect\citeauthoryear{Li et~al.}{2024}]{ref126_tfa-lt}
\begin{bchapter}
\bauthor{\bsnm{Li}, \binits{S.}},
\bauthor{\bsnm{Lin}, \binits{L.}},
\bauthor{\bsnm{Huang}, \binits{Y.}},
\bauthor{\bsnm{Cheng}, \binits{P.}},
\bauthor{\bsnm{Tang}, \binits{X.}}:
\bctitle{Text-guided foundation model adaptation for long-tailed medical image
  classification}.
In: \bbtitle{2024 IEEE International Symposium on Biomedical Imaging (ISBI)},
pp. \bfpage{1}--\blpage{5}
(\byear{2024}).
\bcomment{IEEE}
\end{bchapter}
\endbibitem

%%% 94
\bibitem[\protect\citeauthoryear{Li et~al.}{2026}]{li2026semimedsam}
\begin{barticle}
\bauthor{\bsnm{Li}, \binits{J.}},
\bauthor{\bsnm{Li}, \binits{Y.}},
\bauthor{\bsnm{Wu}, \binits{J.}},
\bauthor{\bsnm{Ji}, \binits{C.}},
\bauthor{\bsnm{Chen}, \binits{Z.}},
\bauthor{\bsnm{Lei}, \binits{W.}},
\bauthor{\bsnm{Wang}, \binits{R.}}:
\batitle{{Semi-MedSAM}: Adapting {SAM}-assisted semi-supervised multi-modality
  learning for medical endoscopic image segmentation}.
\bjtitle{Pattern Recognition}
\bvolume{176},
\bfpage{113206}
(\byear{2026})
\doiurl{10.1016/j.patcog.2026.113206}
\end{barticle}
\endbibitem

%%% 95
\bibitem[\protect\citeauthoryear{Longpre et~al.}{2024}]{longpre2024data}
\begin{bchapter}
\bauthor{\bsnm{Longpre}, \binits{S.}},
\bauthor{\bsnm{Mahari}, \binits{R.}},
\bauthor{\bsnm{Obeng-Marnu}, \binits{N.}},
\bauthor{\bsnm{Brannon}, \binits{W.}},
\bauthor{\bsnm{South}, \binits{T.}},
\bauthor{\bsnm{Gero}, \binits{K.I.}},
\bauthor{\bsnm{Pentland}, \binits{A.}},
\bauthor{\bsnm{Kabbara}, \binits{J.}}:
\bctitle{Position: Data authenticity, consent, \& provenance for {AI} are all
  broken: what will it take to fix them?}
In: \beditor{\bsnm{Salakhutdinov}, \binits{R.}},
\beditor{\bsnm{Kolter}, \binits{Z.}},
\beditor{\bsnm{Heller}, \binits{K.}},
\beditor{\bsnm{Weller}, \binits{A.}},
\beditor{\bsnm{Oliver}, \binits{N.}},
\beditor{\bsnm{Scarlett}, \binits{J.}},
\beditor{\bsnm{Berkenkamp}, \binits{F.}} (eds.)
\bbtitle{Proceedings of the 41st International Conference on Machine Learning}.
\bsertitle{Proceedings of Machine Learning Research},
vol. \bseriesno{235},
pp. \bfpage{32711}--\blpage{32725}.
\bpublisher{PMLR},
\blocation{Vienna, Austria}
(\byear{2024}).
\burl{https://proceedings.mlr.press/v235/longpre24b.html}
\end{bchapter}
\endbibitem

%%% 96
\bibitem[\protect\citeauthoryear{Liang et~al.}{2025}]{ref36_vfm_mia}
\begin{bchapter}
\bauthor{\bsnm{Liang}, \binits{P.}},
\bauthor{\bsnm{Pu}, \binits{B.}},
\bauthor{\bsnm{Huang}, \binits{H.}},
\bauthor{\bsnm{Li}, \binits{Y.}},
\bauthor{\bsnm{Wang}, \binits{H.}},
\bauthor{\bsnm{Ma}, \binits{W.}},
\bauthor{\bsnm{Chang}, \binits{Q.}}:
\bctitle{Vision foundation models in medical image analysis: Advances and
  challenges}.
In: \bbtitle{International Conference on Blockchain and Trustworthy Systems},
pp. \bfpage{170}--\blpage{181}
(\byear{2025}).
\bcomment{Springer}
\end{bchapter}
\endbibitem

%%% 97
\bibitem[\protect\citeauthoryear{Luo et~al.}{2026}]{luo2026kepil}
\begin{botherref}
\oauthor{\bsnm{Luo}, \binits{H.}},
\oauthor{\bsnm{Shu}, \binits{S.Z.}},
\oauthor{\bsnm{Zhou}, \binits{Z.}},
\oauthor{\bsnm{Berke}, \binits{R.}},
\oauthor{\bsnm{Reyes}, \binits{M.}}:
{KEPIL}: Knowledge-enhanced prompt-image learning for prompt-robust disease
  detection.
arXiv preprint arXiv:2605.09132
(2026)
\end{botherref}
\endbibitem

%%% 98
\bibitem[\protect\citeauthoryear{Li et~al.}{2023}]{li2023llava}
\begin{barticle}
\bauthor{\bsnm{Li}, \binits{C.}},
\bauthor{\bsnm{Wong}, \binits{C.}},
\bauthor{\bsnm{Zhang}, \binits{S.}},
\bauthor{\bsnm{Usuyama}, \binits{N.}},
\bauthor{\bsnm{Liu}, \binits{H.}},
\bauthor{\bsnm{Yang}, \binits{J.}},
\bauthor{\bsnm{Naumann}, \binits{T.}},
\bauthor{\bsnm{Poon}, \binits{H.}},
\bauthor{\bsnm{Gao}, \binits{J.}}:
\batitle{Llava-med: Training a large language-and-vision assistant for
  biomedicine in one day}.
\bjtitle{Advances in neural information processing systems}
\bvolume{36},
\bfpage{28541}--\blpage{28564}
(\byear{2023})
\end{barticle}
\endbibitem

%%% 99
\bibitem[\protect\citeauthoryear{Lei et~al.}{2025}]{ref122_medlsam}
\begin{barticle}
\bauthor{\bsnm{Lei}, \binits{W.}},
\bauthor{\bsnm{Xu}, \binits{W.}},
\bauthor{\bsnm{Li}, \binits{K.}},
\bauthor{\bsnm{Zhang}, \binits{X.}},
\bauthor{\bsnm{Zhang}, \binits{S.}}:
\batitle{Medlsam: Localize and segment anything model for 3d ct images}.
\bjtitle{Medical Image Analysis}
\bvolume{99},
\bfpage{103370}
(\byear{2025})
\doiurl{10.1016/j.media.2024.103370}
\end{barticle}
\endbibitem

%%% 100
\bibitem[\protect\citeauthoryear{Lian et~al.}{2025}]{lian2025alta}
\begin{barticle}
\bauthor{\bsnm{Lian}, \binits{C.}},
\bauthor{\bsnm{Zhou}, \binits{H.-Y.}},
\bauthor{\bsnm{Liang}, \binits{D.}},
\bauthor{\bsnm{Qin}, \binits{J.}},
\bauthor{\bsnm{Wang}, \binits{L.}}:
\batitle{Efficient medical vision-language alignment through adapting masked
  vision models}.
\bjtitle{IEEE Transactions on Medical Imaging}
(\byear{2025})
\doiurl{10.1109/TMI.2025.3575853}
\end{barticle}
\endbibitem

%%% 101
\bibitem[\protect\citeauthoryear{Moor et~al.}{2023}]{ref34_generalist_fm}
\begin{barticle}
\bauthor{\bsnm{Moor}, \binits{M.}},
\bauthor{\bsnm{Banerjee}, \binits{O.}},
\bauthor{\bsnm{Abad}, \binits{Z.S.H.}},
\bauthor{\bsnm{Krumholz}, \binits{H.M.}},
\bauthor{\bsnm{Leskovec}, \binits{J.}},
\bauthor{\bsnm{Topol}, \binits{E.J.}},
\bauthor{\bsnm{Rajpurkar}, \binits{P.}}:
\batitle{Foundation models for generalist medical artificial intelligence}.
\bjtitle{Nature}
\bvolume{616}(\bissue{7956}),
\bfpage{259}--\blpage{265}
(\byear{2023})
\end{barticle}
\endbibitem

%%% 102
\bibitem[\protect\citeauthoryear{Mahapatra et~al.}{2026}]{mahapatra2026valiant}
\begin{bchapter}
\bauthor{\bsnm{Mahapatra}, \binits{D.}},
\bauthor{\bsnm{Bozorgtabar}, \binits{B.}},
\bauthor{\bsnm{Roy}, \binits{S.}},
\bauthor{\bsnm{Razzak}, \binits{I.}},
\bauthor{\bsnm{Reyes}, \binits{M.}}:
\bctitle{{VALIANT}: Prompt instability for active learning in black-box medical
  imaging}.
In: \bbtitle{Proceedings of the AAAI Conference on Artificial Intelligence},
vol. \bseriesno{40},
pp. \bfpage{7901}--\blpage{7909}
(\byear{2026})
\end{bchapter}
\endbibitem

%%% 103
\bibitem[\protect\citeauthoryear{Ma et~al.}{2022}]{ma2022testtime}
\begin{bchapter}
\bauthor{\bsnm{Ma}, \binits{W.}},
\bauthor{\bsnm{Chen}, \binits{C.}},
\bauthor{\bsnm{Zheng}, \binits{S.}},
\bauthor{\bsnm{Qin}, \binits{J.}},
\bauthor{\bsnm{Zhang}, \binits{H.}},
\bauthor{\bsnm{Dou}, \binits{Q.}}:
\bctitle{Test-time adaptation with calibration of medical image classification
  nets for label distribution shift}.
In: \bbtitle{International Conference on Medical Image Computing and
  Computer-Assisted Intervention},
pp. \bfpage{313}--\blpage{323}
(\byear{2022}).
\bcomment{Springer}
\end{bchapter}
\endbibitem

%%% 104
\bibitem[\protect\citeauthoryear{Maier-Hein et~al.}{2018}]{maier2018rankings}
\begin{barticle}
\bauthor{\bsnm{Maier-Hein}, \binits{L.}},
\bauthor{\bsnm{Eisenmann}, \binits{M.}},
\bauthor{\bsnm{Reinke}, \binits{A.}},
\bauthor{\bsnm{Onogur}, \binits{S.}},
\bauthor{\bsnm{Stankovic}, \binits{M.}},
\bauthor{\bsnm{Scholz}, \binits{P.}},
\bauthor{\bsnm{Arbel}, \binits{T.}},
\bauthor{\bsnm{Bogunovic}, \binits{H.}},
\bauthor{\bsnm{Bradley}, \binits{A.P.}},
\bauthor{\bsnm{Carass}, \binits{A.}}, \betal:
\batitle{Why rankings of biomedical image analysis competitions should be
  interpreted with care}.
\bjtitle{Nature communications}
\bvolume{9}(\bissue{1}),
\bfpage{5217}
(\byear{2018})
\end{barticle}
\endbibitem

%%% 105
\bibitem[\protect\citeauthoryear{Ma et~al.}{2024}]{ref31_medsam}
\begin{barticle}
\bauthor{\bsnm{Ma}, \binits{J.}},
\bauthor{\bsnm{He}, \binits{Y.}},
\bauthor{\bsnm{Li}, \binits{F.}},
\bauthor{\bsnm{Han}, \binits{L.}},
\bauthor{\bsnm{You}, \binits{C.}},
\bauthor{\bsnm{Wang}, \binits{B.}}:
\batitle{Segment anything in medical images}.
\bjtitle{Nature Communications}
\bvolume{15}(\bissue{1}),
\bfpage{654}
(\byear{2024})
\end{barticle}
\endbibitem

%%% 106
\bibitem[\protect\citeauthoryear{Maier-Hein et~al.}{2024}]{ref47_metrics}
\begin{barticle}
\bauthor{\bsnm{Maier-Hein}, \binits{L.}},
\bauthor{\bsnm{Reinke}, \binits{A.}},
\bauthor{\bsnm{Godau}, \binits{P.}},
\bauthor{\bsnm{Tizabi}, \binits{M.D.}},
\bauthor{\bsnm{Buettner}, \binits{F.}},
\bauthor{\bsnm{Christodoulou}, \binits{E.}},
\bauthor{\bsnm{Glocker}, \binits{B.}},
\bauthor{\bsnm{Isensee}, \binits{F.}},
\bauthor{\bsnm{Kleesiek}, \binits{J.}},
\bauthor{\bsnm{Kozubek}, \binits{M.}}, \betal:
\batitle{Metrics reloaded: recommendations for image analysis validation}.
\bjtitle{Nature methods}
\bvolume{21}(\bissue{2}),
\bfpage{195}--\blpage{212}
(\byear{2024})
\end{barticle}
\endbibitem

%%% 107
\bibitem[\protect\citeauthoryear{Markus et~al.}{2021}]{markus2021role}
\begin{barticle}
\bauthor{\bsnm{Markus}, \binits{A.F.}},
\bauthor{\bsnm{Kors}, \binits{J.A.}},
\bauthor{\bsnm{Rijnbeek}, \binits{P.R.}}:
\batitle{The role of explainability in creating trustworthy artificial
  intelligence for health care: a comprehensive survey of the terminology,
  design choices, and evaluation strategies}.
\bjtitle{Journal of biomedical informatics}
\bvolume{113},
\bfpage{103655}
(\byear{2021})
\end{barticle}
\endbibitem

%%% 108
\bibitem[\protect\citeauthoryear{Ma et~al.}{2024}]{ma2024umamba}
\begin{botherref}
\oauthor{\bsnm{Ma}, \binits{J.}},
\oauthor{\bsnm{Li}, \binits{F.}},
\oauthor{\bsnm{Wang}, \binits{B.}}:
U-mamba: Enhancing long-range dependency for biomedical image segmentation.
arXiv preprint arXiv:2401.04722
(2024)
\end{botherref}
\endbibitem

%%% 109
\bibitem[\protect\citeauthoryear{Mennella et~al.}{2024}]{mennella2024ethical}
\begin{botherref}
\oauthor{\bsnm{Mennella}, \binits{C.}},
\oauthor{\bsnm{Maniscalco}, \binits{U.}},
\oauthor{\bsnm{De~Pietro}, \binits{G.}},
\oauthor{\bsnm{Esposito}, \binits{M.}}:
Ethical and regulatory challenges of ai technologies in healthcare: A narrative
  review.
Heliyon
\textbf{10}(4)
(2024)
\end{botherref}
\endbibitem

%%% 110
\bibitem[\protect\citeauthoryear{M{\"u}ller et~al.}{2022}]{muller2022towards}
\begin{barticle}
\bauthor{\bsnm{M{\"u}ller}, \binits{D.}},
\bauthor{\bsnm{Soto-Rey}, \binits{I.}},
\bauthor{\bsnm{Kramer}, \binits{F.}}:
\batitle{Towards a guideline for evaluation metrics in medical image
  segmentation}.
\bjtitle{BMC research notes}
\bvolume{15}(\bissue{1}),
\bfpage{210}
(\byear{2022})
\end{barticle}
\endbibitem

%%% 111
\bibitem[\protect\citeauthoryear{Messina et~al.}{2026}]{messina2026cure}
\begin{barticle}
\bauthor{\bsnm{Messina}, \binits{P.}},
\bauthor{\bsnm{Villa}, \binits{A.}},
\bauthor{\bsnm{Leon~Alcazar}, \binits{J.}},
\bauthor{\bsnm{Sanchez}, \binits{K.}},
\bauthor{\bsnm{Hinojosa}, \binits{C.}},
\bauthor{\bsnm{Parra}, \binits{D.}},
\bauthor{\bsnm{Soto}, \binits{A.}},
\bauthor{\bsnm{Ghanem}, \binits{B.}}:
\batitle{{CURE}: Curriculum-guided multi-task training for reliable anatomy
  grounded report generation}.
\bjtitle{arXiv preprint arXiv:2601.15408}
(\byear{2026})
\doiurl{10.48550/arXiv.2601.15408}
\end{barticle}
\endbibitem

%%% 112
\bibitem[\protect\citeauthoryear{Ma et~al.}{2025}]{ref157_medsam2}
\begin{botherref}
\oauthor{\bsnm{Ma}, \binits{J.}},
\oauthor{\bsnm{Yang}, \binits{Z.}},
\oauthor{\bsnm{Kim}, \binits{S.}},
\oauthor{\bsnm{Chen}, \binits{B.}},
\oauthor{\bsnm{Baharoon}, \binits{M.}},
\oauthor{\bsnm{Fallahpour}, \binits{A.}},
\oauthor{\bsnm{Asakereh}, \binits{R.}},
\oauthor{\bsnm{Lyu}, \binits{H.}},
\oauthor{\bsnm{Wang}, \binits{B.}}:
Medsam2: Segment anything in 3d medical images and videos.
arXiv preprint arXiv:2504.03600
(2025)
\end{botherref}
\endbibitem

%%% 113
\bibitem[\protect\citeauthoryear{Nakka}{2025}]{nakka2025mammosae}
\begin{bchapter}
\bauthor{\bsnm{Nakka}, \binits{K.K.}}:
\bctitle{Mammo-sae: interpreting breast cancer concept learning with sparse
  autoencoders}.
In: \bbtitle{Deep Breast Workshop on AI and Imaging for Diagnostic and
  Treatment Challenges in Breast Care},
pp. \bfpage{11}--\blpage{20}
(\byear{2025}).
\bcomment{Springer}
\end{bchapter}
\endbibitem

%%% 114
\bibitem[\protect\citeauthoryear{Nouyed et~al.}{2026}]{nouyed2026retinalfm}
\begin{barticle}
\bauthor{\bsnm{Nouyed}, \binits{M.I.}},
\bauthor{\bsnm{Al-Mamun}, \binits{M.}},
\bauthor{\bsnm{Adjeroh}, \binits{D.A.}},
\bauthor{\bsnm{Hu}, \binits{G.}}:
\batitle{Comparative analysis of general-purpose vs. domain-specific multimodal
  models for diabetic retinopathy classification}.
\bjtitle{Diagnostics}
\bvolume{16}(\bissue{10}),
\bfpage{1504}
(\byear{2026})
\doiurl{10.3390/diagnostics16101504}
\end{barticle}
\endbibitem

%%% 115
\bibitem[\protect\citeauthoryear{Nam and Park}{2026}]{nam2026selfrefining}
\begin{barticle}
\bauthor{\bsnm{Nam}, \binits{S.}},
\bauthor{\bsnm{Park}, \binits{S.H.}}:
\batitle{Self-refining segment anything model for nuclei segmentation as
  contrastive learning approach to label-efficient pathological imaging}.
\bjtitle{Diagnostics}
\bvolume{16}(\bissue{9}),
\bfpage{1370}
(\byear{2026})
\doiurl{10.3390/diagnostics16091370}
\end{barticle}
\endbibitem

%%% 116
\bibitem[\protect\citeauthoryear{Nguyen et~al.}{2026}]{nguyen2026clinically}
\begin{botherref}
\oauthor{\bsnm{Nguyen}, \binits{J.}},
\oauthor{\bsnm{Yu}, \binits{J.}},
\oauthor{\bsnm{Fang}, \binits{J.}},
\oauthor{\bsnm{Konz}, \binits{N.}},
\oauthor{\bsnm{Chen}, \binits{T.}},
\oauthor{\bsnm{Krishnan}, \binits{S.}},
\oauthor{\bsnm{Krishnan}, \binits{C.}},
\oauthor{\bsnm{Ding}, \binits{Y.}},
\oauthor{\bsnm{Wang}, \binits{H.}},
\oauthor{\bsnm{Shukla}, \binits{A.}}:
Clinically-informed modeling for pediatric brain tumor classification from
  whole-slide histopathology images.
arXiv preprint arXiv:2604.21060
(2026)
\end{botherref}
\endbibitem

%%% 117
\bibitem[\protect\citeauthoryear{Nath et~al.}{2020}]{nath2020diminishing}
\begin{barticle}
\bauthor{\bsnm{Nath}, \binits{V.}},
\bauthor{\bsnm{Yang}, \binits{D.}},
\bauthor{\bsnm{Landman}, \binits{B.A.}},
\bauthor{\bsnm{Xu}, \binits{D.}},
\bauthor{\bsnm{Roth}, \binits{H.R.}}:
\batitle{Diminishing uncertainty within the training pool: Active learning for
  medical image segmentation}.
\bjtitle{IEEE Transactions on Medical Imaging}
\bvolume{40}(\bissue{10}),
\bfpage{2534}--\blpage{2547}
(\byear{2020})
\end{barticle}
\endbibitem

%%% 118
\bibitem[\protect\citeauthoryear{Ostmeier et~al.}{2023}]{OSTMEIER2023102927}
\begin{barticle}
\bauthor{\bsnm{Ostmeier}, \binits{S.}},
\bauthor{\bsnm{Axelrod}, \binits{B.}},
\bauthor{\bsnm{Isensee}, \binits{F.}},
\bauthor{\bsnm{Bertels}, \binits{J.}},
\bauthor{\bsnm{Mlynash}, \binits{M.}},
\bauthor{\bsnm{Christensen}, \binits{S.}},
\bauthor{\bsnm{Lansberg}, \binits{M.G.}},
\bauthor{\bsnm{Albers}, \binits{G.W.}},
\bauthor{\bsnm{Sheth}, \binits{R.}},
\bauthor{\bsnm{Verhaaren}, \binits{B.F.J.}},
\bauthor{\bsnm{Mahammedi}, \binits{A.}},
\bauthor{\bsnm{Li}, \binits{L.-J.}},
\bauthor{\bsnm{Zaharchuk}, \binits{G.}},
\bauthor{\bsnm{Heit}, \binits{J.J.}}:
\batitle{Use-evaluator: Performance metrics for medical image segmentation
  models supervised by uncertain, small or empty reference annotations in
  neuroimaging}.
\bjtitle{Medical Image Analysis}
\bvolume{90},
\bfpage{102927}
(\byear{2023})
\doiurl{10.1016/j.media.2023.102927}
\end{barticle}
\endbibitem

%%% 119
\bibitem[\protect\citeauthoryear{Ong~Ly et~al.}{2024}]{ong2024shortcut}
\begin{barticle}
\bauthor{\bsnm{Ong~Ly}, \binits{C.}},
\bauthor{\bsnm{Unnikrishnan}, \binits{B.}},
\bauthor{\bsnm{Tadic}, \binits{T.}},
\bauthor{\bsnm{Patel}, \binits{T.}},
\bauthor{\bsnm{Duhamel}, \binits{J.}},
\bauthor{\bsnm{Kandel}, \binits{S.}},
\bauthor{\bsnm{Moayedi}, \binits{Y.}},
\bauthor{\bsnm{Brudno}, \binits{M.}},
\bauthor{\bsnm{Hope}, \binits{A.}},
\bauthor{\bsnm{Ross}, \binits{H.}}, \betal:
\batitle{Shortcut learning in medical ai hinders generalization: method for
  estimating ai model generalization without external data}.
\bjtitle{NPJ digital medicine}
\bvolume{7}(\bissue{1}),
\bfpage{124}
(\byear{2024})
\end{barticle}
\endbibitem

%%% 120
\bibitem[\protect\citeauthoryear{Pai et~al.}{2024}]{pai2024cancerfm}
\begin{barticle}
\bauthor{\bsnm{Pai}, \binits{S.}},
\bauthor{\bsnm{Bontempi}, \binits{D.}},
\bauthor{\bsnm{Hadzic}, \binits{I.}},
\bauthor{\bsnm{Prudente}, \binits{V.}},
\bauthor{\bsnm{Soka\v{c}}, \binits{M.}},
\bauthor{\bsnm{Chaunzwa}, \binits{T.L.}},
\bauthor{\bsnm{Bernatz}, \binits{S.}},
\bauthor{\bsnm{Hosny}, \binits{A.}},
\bauthor{\bsnm{Mak}, \binits{R.H.}},
\bauthor{\bsnm{Birkbak}, \binits{N.J.}},
\bauthor{\bsnm{Aerts}, \binits{H.J.W.L.}}:
\batitle{Foundation model for cancer imaging biomarkers}.
\bjtitle{Nature Machine Intelligence}
\bvolume{6}(\bissue{3}),
\bfpage{354}--\blpage{367}
(\byear{2024})
\doiurl{10.1038/s42256-024-00807-9}
\end{barticle}
\endbibitem

%%% 121
\bibitem[\protect\citeauthoryear{Paschali
  et~al.}{2025}]{paschali2025foundation}
\begin{barticle}
\bauthor{\bsnm{Paschali}, \binits{M.}},
\bauthor{\bsnm{Chen}, \binits{Z.}},
\bauthor{\bsnm{Blankemeier}, \binits{L.}},
\bauthor{\bsnm{Varma}, \binits{M.}},
\bauthor{\bsnm{Youssef}, \binits{A.}},
\bauthor{\bsnm{Bluethgen}, \binits{C.}},
\bauthor{\bsnm{Langlotz}, \binits{C.}},
\bauthor{\bsnm{Gatidis}, \binits{S.}},
\bauthor{\bsnm{Chaudhari}, \binits{A.}}:
\batitle{Foundation models in radiology: what, how, why, and why not}.
\bjtitle{Radiology}
\bvolume{314}(\bissue{2}),
\bfpage{240597}
(\byear{2025})
\end{barticle}
\endbibitem

%%% 122
\bibitem[\protect\citeauthoryear{Perkonigg et~al.}{2021}]{perkonigg2021dynamic}
\begin{barticle}
\bauthor{\bsnm{Perkonigg}, \binits{M.}},
\bauthor{\bsnm{Hofmanninger}, \binits{J.}},
\bauthor{\bsnm{Herold}, \binits{C.J.}},
\bauthor{\bsnm{Brink}, \binits{J.A.}},
\bauthor{\bsnm{Pianykh}, \binits{O.}},
\bauthor{\bsnm{Prosch}, \binits{H.}},
\bauthor{\bsnm{Langs}, \binits{G.}}:
\batitle{Dynamic memory to alleviate catastrophic forgetting in continual
  learning with medical imaging}.
\bjtitle{Nature communications}
\bvolume{12}(\bissue{1}),
\bfpage{5678}
(\byear{2021})
\end{barticle}
\endbibitem

%%% 123
\bibitem[\protect\citeauthoryear{Park et~al.}{2026}]{park2026glint}
\begin{botherref}
\oauthor{\bsnm{Park}, \binits{J.}},
\oauthor{\bsnm{Lee}, \binits{S.}},
\oauthor{\bsnm{Park}, \binits{J.}},
\oauthor{\bsnm{Yun}, \binits{H.}},
\oauthor{\bsnm{Kim}, \binits{H.}},
\oauthor{\bsnm{Jeong}, \binits{S.}},
\oauthor{\bsnm{Kang}, \binits{H.}},
\oauthor{\bsnm{Yoon}, \binits{B.}},
\oauthor{\bsnm{Choi}, \binits{K.}}:
{GLINT}: Sparsely gated vision-language alignment for fine-grained radiology
  representations.
arXiv preprint arXiv:2606.03180
(2026)
\end{botherref}
\endbibitem

%%% 124
\bibitem[\protect\citeauthoryear{Padilla
  et~al.}{2020}]{ref48_detection_metrics}
\begin{bchapter}
\bauthor{\bsnm{Padilla}, \binits{R.}},
\bauthor{\bsnm{Netto}, \binits{S.L.}},
\bauthor{\bsnm{Silva}, \binits{E.A.B.}}:
\bctitle{A survey on performance metrics for object-detection algorithms}.
In: \bbtitle{2020 International Conference on Systems, Signals and Image
  Processing (IWSSIP)},
pp. \bfpage{237}--\blpage{242}
(\byear{2020}).
\doiurl{10.1109/IWSSIP48289.2020.9145130}
\end{bchapter}
\endbibitem

%%% 125
\bibitem[\protect\citeauthoryear{Pfohl et~al.}{2022}]{pfohl2022comparison}
\begin{barticle}
\bauthor{\bsnm{Pfohl}, \binits{S.R.}},
\bauthor{\bsnm{Zhang}, \binits{H.}},
\bauthor{\bsnm{Xu}, \binits{Y.}},
\bauthor{\bsnm{Foryciarz}, \binits{A.}},
\bauthor{\bsnm{Ghassemi}, \binits{M.}},
\bauthor{\bsnm{Shah}, \binits{N.H.}}:
\batitle{A comparison of approaches to improve worst-case predictive model
  performance over patient subpopulations}.
\bjtitle{Scientific reports}
\bvolume{12}(\bissue{1}),
\bfpage{3254}
(\byear{2022})
\end{barticle}
\endbibitem

%%% 126
\bibitem[\protect\citeauthoryear{Qiu et~al.}{2025}]{qiu2025effortless}
\begin{barticle}
\bauthor{\bsnm{Qiu}, \binits{J.}},
\bauthor{\bsnm{Jain}, \binits{N.}},
\bauthor{\bsnm{Ammeling}, \binits{J.}},
\bauthor{\bsnm{Aubreville}, \binits{M.}},
\bauthor{\bsnm{Breininger}, \binits{K.}}:
\batitle{Effortless vision-language model specialization in histopathology
  without annotation}.
\bjtitle{arXiv preprint arXiv:2508.07835}
(\byear{2025})
\doiurl{10.48550/arXiv.2508.07835}
\end{barticle}
\endbibitem

%%% 127
\bibitem[\protect\citeauthoryear{Qin et~al.}{2023}]{qin2023medical_vlm}
\begin{bchapter}
\bauthor{\bsnm{Qin}, \binits{Z.}},
\bauthor{\bsnm{Yi}, \binits{H.}},
\bauthor{\bsnm{Lao}, \binits{Q.}},
\bauthor{\bsnm{Li}, \binits{K.}}:
\bctitle{Medical image understanding with pretrained vision language models: A
  comprehensive study}.
In: \bbtitle{The Eleventh International Conference on Learning Representations}
(\byear{2023})
\end{bchapter}
\endbibitem

%%% 128
\bibitem[\protect\citeauthoryear{Qi et~al.}{2025}]{qi2025mediaug}
\begin{bchapter}
\bauthor{\bsnm{Qi}, \binits{X.}},
\bauthor{\bsnm{Zhang}, \binits{Z.}},
\bauthor{\bsnm{Gang}, \binits{C.}},
\bauthor{\bsnm{Zhang}, \binits{H.}},
\bauthor{\bsnm{Zhang}, \binits{L.}},
\bauthor{\bsnm{Zhang}, \binits{Z.}},
\bauthor{\bsnm{Zhao}, \binits{Y.}}:
\bctitle{Mediaug: Exploring visual augmentation in medical imaging}.
In: \bbtitle{Annual Conference on Medical Image Understanding and Analysis},
pp. \bfpage{218}--\blpage{232}
(\byear{2025}).
\bcomment{Springer}
\end{bchapter}
\endbibitem

%%% 129
\bibitem[\protect\citeauthoryear{Rolnick
  et~al.}{2019}]{ref148_experience_replay}
\begin{botherref}
\oauthor{\bsnm{Rolnick}, \binits{D.}},
\oauthor{\bsnm{Ahuja}, \binits{A.}},
\oauthor{\bsnm{Schwarz}, \binits{J.}},
\oauthor{\bsnm{Lillicrap}, \binits{T.}},
\oauthor{\bsnm{Wayne}, \binits{G.}}:
Experience replay for continual learning.
Advances in neural information processing systems
\textbf{32}
(2019)
\end{botherref}
\endbibitem

%%% 130
\bibitem[\protect\citeauthoryear{Raja~Babu et~al.}{2024}]{raja2024synthetic}
\begin{bchapter}
\bauthor{\bsnm{Raja~Babu}, \binits{K.A.}},
\bauthor{\bsnm{Sathish}, \binits{R.}},
\bauthor{\bsnm{Pattanaik}, \binits{M.}},
\bauthor{\bsnm{Venkataramani}, \binits{R.}}:
\bctitle{Synthetic simplicity: Unveiling bias in medical data augmentation}.
In: \bbtitle{MICCAI Workshop on Data Engineering in Medical Imaging},
pp. \bfpage{64}--\blpage{72}
(\byear{2024}).
\bcomment{Springer}
\end{bchapter}
\endbibitem

%%% 131
\bibitem[\protect\citeauthoryear{Rieke et~al.}{2020}]{rieke2020future}
\begin{barticle}
\bauthor{\bsnm{Rieke}, \binits{N.}},
\bauthor{\bsnm{Hancox}, \binits{J.}},
\bauthor{\bsnm{Li}, \binits{W.}},
\bauthor{\bsnm{Milletari}, \binits{F.}},
\bauthor{\bsnm{Roth}, \binits{H.R.}},
\bauthor{\bsnm{Albarqouni}, \binits{S.}},
\bauthor{\bsnm{Bakas}, \binits{S.}},
\bauthor{\bsnm{Galtier}, \binits{M.N.}},
\bauthor{\bsnm{Landman}, \binits{B.A.}},
\bauthor{\bsnm{Maier-Hein}, \binits{K.}}, \betal:
\batitle{The future of digital health with federated learning}.
\bjtitle{NPJ digital medicine}
\bvolume{3}(\bissue{1}),
\bfpage{119}
(\byear{2020})
\end{barticle}
\endbibitem

%%% 132
\bibitem[\protect\citeauthoryear{Ryu et~al.}{2025}]{ryu2025vision}
\begin{barticle}
\bauthor{\bsnm{Ryu}, \binits{J.S.}},
\bauthor{\bsnm{Kang}, \binits{H.}},
\bauthor{\bsnm{Chu}, \binits{Y.}},
\bauthor{\bsnm{Yang}, \binits{S.}}:
\batitle{Vision-language foundation models for medical imaging: a review of
  current practices and innovations}.
\bjtitle{Biomedical Engineering Letters}
\bvolume{15}(\bissue{5}),
\bfpage{809}--\blpage{830}
(\byear{2025})
\end{barticle}
\endbibitem

%%% 133
\bibitem[\protect\citeauthoryear{Radford et~al.}{2021}]{ref22_clip}
\begin{bchapter}
\bauthor{\bsnm{Radford}, \binits{A.}},
\bauthor{\bsnm{Kim}, \binits{J.W.}},
\bauthor{\bsnm{Hallacy}, \binits{C.}},
\bauthor{\bsnm{Ramesh}, \binits{A.}},
\bauthor{\bsnm{Goh}, \binits{G.}},
\bauthor{\bsnm{Agarwal}, \binits{S.}},
\bauthor{\bsnm{Sastry}, \binits{G.}},
\bauthor{\bsnm{Askell}, \binits{A.}},
\bauthor{\bsnm{Mishkin}, \binits{P.}},
\bauthor{\bsnm{Clark}, \binits{J.}}, \betal:
\bctitle{Learning transferable visual models from natural language
  supervision}.
In: \bbtitle{International Conference on Machine Learning},
pp. \bfpage{8748}--\blpage{8763}
(\byear{2021}).
\bcomment{PmLR}
\end{bchapter}
\endbibitem

%%% 134
\bibitem[\protect\citeauthoryear{Rajendran
  et~al.}{2025}]{rajendran2025foundation}
\begin{botherref}
\oauthor{\bsnm{Rajendran}, \binits{P.}},
\oauthor{\bsnm{Safari}, \binits{M.}},
\oauthor{\bsnm{He}, \binits{W.}},
\oauthor{\bsnm{Hu}, \binits{M.}},
\oauthor{\bsnm{Wang}, \binits{S.}},
\oauthor{\bsnm{Zhou}, \binits{J.}},
\oauthor{\bsnm{Yang}, \binits{X.}}:
Foundation models in medical image analysis: A systematic review and
  meta-analysis.
arXiv preprint arXiv:2510.16973
(2025)
\end{botherref}
\endbibitem

%%% 135
\bibitem[\protect\citeauthoryear{Raji et~al.}{2020}]{raji2020closing}
\begin{bchapter}
\bauthor{\bsnm{Raji}, \binits{I.D.}},
\bauthor{\bsnm{Smart}, \binits{A.}},
\bauthor{\bsnm{White}, \binits{R.N.}},
\bauthor{\bsnm{Mitchell}, \binits{M.}},
\bauthor{\bsnm{Gebru}, \binits{T.}},
\bauthor{\bsnm{Hutchinson}, \binits{B.}},
\bauthor{\bsnm{Smith-Loud}, \binits{J.}},
\bauthor{\bsnm{Theron}, \binits{D.}},
\bauthor{\bsnm{Barnes}, \binits{P.}}:
\bctitle{Closing the ai accountability gap: Defining an end-to-end framework
  for internal algorithmic auditing}.
In: \bbtitle{Proceedings of the 2020 Conference on Fairness, Accountability,
  and Transparency},
pp. \bfpage{33}--\blpage{44}
(\byear{2020})
\end{bchapter}
\endbibitem

%%% 136
\bibitem[\protect\citeauthoryear{Raghu et~al.}{2019}]{ref65_transfusion}
\begin{botherref}
\oauthor{\bsnm{Raghu}, \binits{M.}},
\oauthor{\bsnm{Zhang}, \binits{C.}},
\oauthor{\bsnm{Kleinberg}, \binits{J.}},
\oauthor{\bsnm{Bengio}, \binits{S.}}:
Transfusion: Understanding transfer learning for medical imaging.
Advances in neural information processing systems
\textbf{32}
(2019)
\end{botherref}
\endbibitem

%%% 137
\bibitem[\protect\citeauthoryear{Sellergren
  et~al.}{2025}]{sellergren2025medgemma}
\begin{barticle}
\bauthor{\bsnm{Sellergren}, \binits{A.}},
\bauthor{\bsnm{Kazemzadeh}, \binits{S.}},
\bauthor{\bsnm{Jaroensri}, \binits{T.}},
\bauthor{\bsnm{Kiraly}, \binits{A.}},
\bauthor{\bsnm{Traverse}, \binits{M.}},
\bauthor{\bsnm{Kohlberger}, \binits{T.}},
\bauthor{\bsnm{Xu}, \binits{S.}},
\bauthor{\bsnm{Jamil}, \binits{F.}},
\bauthor{\bsnm{Hughes}, \binits{C.}},
\bauthor{\bsnm{Lau}, \binits{C.}}, \betal:
\batitle{{MedGemma} technical report}.
\bjtitle{arXiv preprint arXiv:2507.05201}
(\byear{2025})
\doiurl{10.48550/arXiv.2507.05201}
\end{barticle}
\endbibitem

%%% 138
\bibitem[\protect\citeauthoryear{Shi et~al.}{2024}]{shi2024shapemamba}
\begin{bchapter}
\bauthor{\bsnm{Shi}, \binits{R.}},
\bauthor{\bsnm{Pang}, \binits{Q.}},
\bauthor{\bsnm{Ma}, \binits{L.}},
\bauthor{\bsnm{Duan}, \binits{L.}},
\bauthor{\bsnm{Huang}, \binits{T.}},
\bauthor{\bsnm{Jiang}, \binits{T.}}:
\bctitle{Shapemamba-em: Fine-tuning foundation model with local shape
  descriptors and mamba blocks for 3d em image segmentation}.
In: \bbtitle{International Conference on Medical Image Computing and
  Computer-assisted Intervention},
pp. \bfpage{731}--\blpage{741}
(\byear{2024}).
\bcomment{Springer}
\end{bchapter}
\endbibitem

%%% 139
\bibitem[\protect\citeauthoryear{Silva-Rodr\'{\i}guez
  et~al.}{2025}]{silvarodriguez2025fseft}
\begin{barticle}
\bauthor{\bsnm{Silva-Rodr\'{\i}guez}, \binits{J.}},
\bauthor{\bsnm{Dolz}, \binits{J.}},
\bauthor{\bsnm{Ben~Ayed}, \binits{I.}}:
\batitle{Towards foundation models and few-shot parameter-efficient fine-tuning
  for volumetric organ segmentation}.
\bjtitle{Medical Image Analysis}
\bvolume{103},
\bfpage{103596}
(\byear{2025})
\doiurl{10.1016/j.media.2025.103596}
\end{barticle}
\endbibitem

%%% 140
\bibitem[\protect\citeauthoryear{Setio et~al.}{2017}]{setio2017validation}
\begin{barticle}
\bauthor{\bsnm{Setio}, \binits{A.A.A.}},
\bauthor{\bsnm{Traverso}, \binits{A.}},
\bauthor{\bsnm{De~Bel}, \binits{T.}},
\bauthor{\bsnm{Berens}, \binits{M.S.}},
\bauthor{\bsnm{Van Den~Bogaard}, \binits{C.}},
\bauthor{\bsnm{Cerello}, \binits{P.}},
\bauthor{\bsnm{Chen}, \binits{H.}},
\bauthor{\bsnm{Dou}, \binits{Q.}},
\bauthor{\bsnm{Fantacci}, \binits{M.E.}},
\bauthor{\bsnm{Geurts}, \binits{B.}}, \betal:
\batitle{Validation, comparison, and combination of algorithms for automatic
  detection of pulmonary nodules in computed tomography images: the luna16
  challenge}.
\bjtitle{Medical image analysis}
\bvolume{42},
\bfpage{1}--\blpage{13}
(\byear{2017})
\end{barticle}
\endbibitem

%%% 141
\bibitem[\protect\citeauthoryear{Tu et~al.}{2024}]{tu2024towards}
\begin{barticle}
\bauthor{\bsnm{Tu}, \binits{T.}},
\bauthor{\bsnm{Azizi}, \binits{S.}},
\bauthor{\bsnm{Driess}, \binits{D.}},
\bauthor{\bsnm{Schaekermann}, \binits{M.}},
\bauthor{\bsnm{Amin}, \binits{M.}},
\bauthor{\bsnm{Chang}, \binits{P.-C.}},
\bauthor{\bsnm{Carroll}, \binits{A.}},
\bauthor{\bsnm{Lau}, \binits{C.}},
\bauthor{\bsnm{Tanno}, \binits{R.}},
\bauthor{\bsnm{Ktena}, \binits{I.}}, \betal:
\batitle{Towards generalist biomedical ai}.
\bjtitle{Nejm Ai}
\bvolume{1}(\bissue{3}),
\bfpage{2300138}
(\byear{2024})
\end{barticle}
\endbibitem

%%% 142
\bibitem[\protect\citeauthoryear{Tampu et~al.}{2022}]{tampu2022inflation}
\begin{barticle}
\bauthor{\bsnm{Tampu}, \binits{I.E.}},
\bauthor{\bsnm{Eklund}, \binits{A.}},
\bauthor{\bsnm{Haj-Hosseini}, \binits{N.}}:
\batitle{Inflation of test accuracy due to data leakage in deep learning-based
  classification of oct images}.
\bjtitle{Scientific Data}
\bvolume{9}(\bissue{1}),
\bfpage{580}
(\byear{2022})
\end{barticle}
\endbibitem

%%% 143
\bibitem[\protect\citeauthoryear{{Therapeutic Goods
  Administration}}{2025}]{tga2025aievidence}
\begin{botherref}
\oauthor{\bsnm{{Therapeutic Goods Administration}}}:
Evidence Requirements for Software Using {AI}.
Last updated September 4, 2025. Accessed August 26, 2026
(2025).
\url{https://www.tga.gov.au/products/medical-devices/software-and-artificial-intelligence-ai/manufacturing/artificial-intelligence-ai-and-medical-device-software-regulation/evidence-requirements-software-using-ai}
\end{botherref}
\endbibitem

%%% 144
\bibitem[\protect\citeauthoryear{{Therapeutic Goods
  Administration}}{2026}]{tga2026software}
\begin{botherref}
\oauthor{\bsnm{{Therapeutic Goods Administration}}}:
Understanding How We Regulate Software-Based Medical Devices.
Published February 24, 2026. Accessed August 26, 2026
(2026).
\url{https://www.tga.gov.au/resources/guidance/understanding-how-we-regulate-software-based-medical-devices}
\end{botherref}
\endbibitem

%%% 145
\bibitem[\protect\citeauthoryear{Tejani et~al.}{2024}]{tejani2024claim}
\begin{barticle}
\bauthor{\bsnm{Tejani}, \binits{A.S.}},
\bauthor{\bsnm{Klontzas}, \binits{M.E.}},
\bauthor{\bsnm{Gatti}, \binits{A.A.}},
\bauthor{\bsnm{Mongan}, \binits{J.T.}},
\bauthor{\bsnm{Moy}, \binits{L.}},
\bauthor{\bsnm{Park}, \binits{S.H.}},
\bauthor{\bsnm{Kahn}, \binits{C.E.} \bsuffix{Jr}},
\bauthor{\bsnm{{CLAIM 2024 Update Panel}}}:
\batitle{Checklist for artificial intelligence in medical imaging ({CLAIM}):
  2024 update}.
\bjtitle{Radiology: Artificial Intelligence}
\bvolume{6}(\bissue{4}),
\bfpage{240300}
(\byear{2024})
\end{barticle}
\endbibitem

%%% 146
\bibitem[\protect\citeauthoryear{Taleb
  et~al.}{2020}]{ref1_3d_ssl_methods_medical}
\begin{barticle}
\bauthor{\bsnm{Taleb}, \binits{A.}},
\bauthor{\bsnm{Loetzsch}, \binits{W.}},
\bauthor{\bsnm{Danz}, \binits{N.}},
\bauthor{\bsnm{Severin}, \binits{J.}},
\bauthor{\bsnm{Gaertner}, \binits{T.}},
\bauthor{\bsnm{Bergner}, \binits{B.}},
\bauthor{\bsnm{Lippert}, \binits{C.}}:
\batitle{3d self-supervised methods for medical imaging}.
\bjtitle{Advances in neural information processing systems}
\bvolume{33},
\bfpage{18158}--\blpage{18172}
(\byear{2020})
\end{barticle}
\endbibitem

%%% 147
\bibitem[\protect\citeauthoryear{{U.S. Food and Drug
  Administration}}{2025}]{fda2025pccp}
\begin{botherref}
\oauthor{\bsnm{{U.S. Food and Drug Administration}}}:
Marketing Submission Recommendations for a Predetermined Change Control Plan
  for Artificial Intelligence-Enabled Device Software Functions.
Final Guidance for Industry and Food and Drug Administration Staff, issued
  August 18, 2025
(2025).
\url{https://www.fda.gov/media/166704/download}
\end{botherref}
\endbibitem

%%% 148
\bibitem[\protect\citeauthoryear{Vasey et~al.}{2022}]{ref_decide_ai2022}
\begin{botherref}
\oauthor{\bsnm{Vasey}, \binits{B.}},
\oauthor{\bsnm{Nagendran}, \binits{M.}},
\oauthor{\bsnm{Campbell}, \binits{B.}},
\oauthor{\bsnm{Clifton}, \binits{D.A.}},
\oauthor{\bsnm{Collins}, \binits{G.S.}},
\oauthor{\bsnm{Denaxas}, \binits{S.}},
\oauthor{\bsnm{Denniston}, \binits{A.K.}},
\oauthor{\bsnm{Faes}, \binits{L.}},
\oauthor{\bsnm{Geerts}, \binits{B.}},
\oauthor{\bsnm{Ibrahim}, \binits{M.}}, et al.:
Reporting guideline for the early stage clinical evaluation of decision support
  systems driven by artificial intelligence: Decide-ai.
bmj
\textbf{377}
(2022)
\end{botherref}
\endbibitem

%%% 149
\bibitem[\protect\citeauthoryear{Vardi et~al.}{2026}]{vardi2026discrepancy}
\begin{botherref}
\oauthor{\bsnm{Vardi}, \binits{B.}},
\oauthor{\bsnm{Schonberger}, \binits{D.}},
\oauthor{\bsnm{Friedmann}, \binits{Y.}},
\oauthor{\bsnm{Yakhini}, \binits{Z.}},
\oauthor{\bsnm{Barshack}, \binits{I.}},
\oauthor{\bsnm{Loebel}, \binits{A.}},
\oauthor{\bsnm{Shamir}, \binits{A.}}:
Discrepancy minimization improves cross-hospital robustness in digital
  pathology.
arXiv preprint arXiv:2605.25175
(2026)
\end{botherref}
\endbibitem

%%% 150
\bibitem[\protect\citeauthoryear{Wang et~al.}{2026}]{wang2026anopilad}
\begin{bchapter}
\bauthor{\bsnm{Wang}, \binits{J.}},
\bauthor{\bsnm{Byeon}, \binits{K.}},
\bauthor{\bsnm{Song}, \binits{J.}},
\bauthor{\bsnm{Nguyen}, \binits{A.}},
\bauthor{\bsnm{Ahn}, \binits{S.}},
\bauthor{\bsnm{Lee}, \binits{S.H.}},
\bauthor{\bsnm{Kwak}, \binits{J.T.}}:
\bctitle{Pathology-informed latent diffusion model for anomaly detection in
  lymph node metastasis}.
In: \bbtitle{Lecture Notes in Computer Science}
(\byear{2026}).
\doiurl{10.1007/978-3-032-04937-7_43}
\end{bchapter}
\endbibitem

%%% 151
\bibitem[\protect\citeauthoryear{Wan et~al.}{2026}]{wan2026dcf}
\begin{barticle}
\bauthor{\bsnm{Wan}, \binits{Y.}},
\bauthor{\bsnm{Chen}, \binits{Z.}},
\bauthor{\bsnm{Xu}, \binits{Y.}},
\bauthor{\bsnm{Xu}, \binits{Y.}},
\bauthor{\bsnm{Li}, \binits{M.}},
\bauthor{\bsnm{Wang}, \binits{Y.}}:
\batitle{Breaking error coupling via divergent--convergent coordination for
  semi-supervised medical image segmentation}.
\bjtitle{Medical Image Analysis}
\bvolume{114},
\bfpage{104236}
(\byear{2026})
\doiurl{10.1016/j.media.2026.104236}
\end{barticle}
\endbibitem

%%% 152
\bibitem[\protect\citeauthoryear{Wang et~al.}{2022}]{9392296}
\begin{barticle}
\bauthor{\bsnm{Wang}, \binits{X.}},
\bauthor{\bsnm{Chen}, \binits{Y.}},
\bauthor{\bsnm{Zhu}, \binits{W.}}:
\batitle{A survey on curriculum learning}.
\bjtitle{IEEE Transactions on Pattern Analysis and Machine Intelligence}
\bvolume{44}(\bissue{9}),
\bfpage{4555}--\blpage{4576}
(\byear{2022})
\doiurl{10.1109/TPAMI.2021.3069908}
\end{barticle}
\endbibitem

%%% 153
\bibitem[\protect\citeauthoryear{Wang et~al.}{2025}]{ref119_sammed3d}
\begin{bchapter}
\bauthor{\bsnm{Wang}, \binits{H.}},
\bauthor{\bsnm{Guo}, \binits{S.}},
\bauthor{\bsnm{Ye}, \binits{J.}},
\bauthor{\bsnm{Deng}, \binits{Z.}},
\bauthor{\bsnm{Cheng}, \binits{J.}},
\bauthor{\bsnm{Li}, \binits{T.}},
\bauthor{\bsnm{Chen}, \binits{J.}},
\bauthor{\bsnm{Su}, \binits{Y.}},
\bauthor{\bsnm{Huang}, \binits{Z.}},
\bauthor{\bsnm{Shen}, \binits{Y.}}, \betal:
\bctitle{Sam-med3d: towards general-purpose segmentation models for volumetric
  medical images}.
In: \bbtitle{European Conference on Computer Vision},
pp. \bfpage{51}--\blpage{67}
(\byear{2025}).
\bcomment{Springer}
\end{bchapter}
\endbibitem

%%% 154
\bibitem[\protect\citeauthoryear{Wortsman et~al.}{2022}]{wortsman2022robust}
\begin{bchapter}
\bauthor{\bsnm{Wortsman}, \binits{M.}},
\bauthor{\bsnm{Ilharco}, \binits{G.}},
\bauthor{\bsnm{Kim}, \binits{J.W.}},
\bauthor{\bsnm{Li}, \binits{M.}},
\bauthor{\bsnm{Kornblith}, \binits{S.}},
\bauthor{\bsnm{Roelofs}, \binits{R.}},
\bauthor{\bsnm{Lopes}, \binits{R.G.}},
\bauthor{\bsnm{Hajishirzi}, \binits{H.}},
\bauthor{\bsnm{Farhadi}, \binits{A.}},
\bauthor{\bsnm{Namkoong}, \binits{H.}}, \betal:
\bctitle{Robust fine-tuning of zero-shot models}.
In: \bbtitle{Proceedings of the IEEE/CVF Conference on Computer Vision and
  Pattern Recognition},
pp. \bfpage{7959}--\blpage{7971}
(\byear{2022})
\end{bchapter}
\endbibitem

%%% 155
\bibitem[\protect\citeauthoryear{Wang et~al.}{2024}]{wang2024comprehensive}
\begin{barticle}
\bauthor{\bsnm{Wang}, \binits{H.}},
\bauthor{\bsnm{Jin}, \binits{Q.}},
\bauthor{\bsnm{Li}, \binits{S.}},
\bauthor{\bsnm{Liu}, \binits{S.}},
\bauthor{\bsnm{Wang}, \binits{M.}},
\bauthor{\bsnm{Song}, \binits{Z.}}:
\batitle{A comprehensive survey on deep active learning in medical image
  analysis}.
\bjtitle{Medical Image Analysis}
\bvolume{95},
\bfpage{103201}
(\byear{2024})
\end{barticle}
\endbibitem

%%% 156
\bibitem[\protect\citeauthoryear{Wu et~al.}{2026}]{wu2026promptgroup}
\begin{botherref}
\oauthor{\bsnm{Wu}, \binits{Y.}},
\oauthor{\bsnm{Liang}, \binits{Z.}},
\oauthor{\bsnm{Zeng}, \binits{W.}},
\oauthor{\bsnm{Xie}, \binits{X.}},
\oauthor{\bsnm{Yu}, \binits{J.}}:
Prompt group-aware training for robust text-guided nuclei segmentation.
arXiv preprint arXiv:2603.06384
(2026)
\end{botherref}
\endbibitem

%%% 157
\bibitem[\protect\citeauthoryear{Wang and Mao}{2026}]{wang2026promptmed}
\begin{barticle}
\bauthor{\bsnm{Wang}, \binits{S.}},
\bauthor{\bsnm{Mao}, \binits{R.}}:
\batitle{{PromptMed}: Prompt-driven semi-supervised medical image
  classification with class-balanced consistency and contrastive learning}.
\bjtitle{Expert Systems with Applications}
\bvolume{310},
\bfpage{131345}
(\byear{2026})
\doiurl{10.1016/j.eswa.2026.131345}
\end{barticle}
\endbibitem

%%% 158
\bibitem[\protect\citeauthoryear{Wang et~al.}{2017}]{Wang_2017_CVPR}
\begin{bchapter}
\bauthor{\bsnm{Wang}, \binits{X.}},
\bauthor{\bsnm{Peng}, \binits{Y.}},
\bauthor{\bsnm{Lu}, \binits{L.}},
\bauthor{\bsnm{Lu}, \binits{Z.}},
\bauthor{\bsnm{Bagheri}, \binits{M.}},
\bauthor{\bsnm{Summers}, \binits{R.M.}}:
\bctitle{Chestx-ray8: Hospital-scale chest x-ray database and benchmarks on
  weakly-supervised classification and localization of common thorax diseases}.
In: \bbtitle{Proceedings of the IEEE Conference on Computer Vision and Pattern
  Recognition (CVPR)}
(\byear{2017})
\end{bchapter}
\endbibitem

%%% 159
\bibitem[\protect\citeauthoryear{Wang et~al.}{2026}]{wang2026astra}
\begin{botherref}
\oauthor{\bsnm{Wang}, \binits{T.}},
\oauthor{\bsnm{Su}, \binits{Z.}},
\oauthor{\bsnm{Akbar}, \binits{A.R.}},
\oauthor{\bsnm{Sajjad}, \binits{U.}},
\oauthor{\bsnm{Gokhale}, \binits{L.}},
\oauthor{\bsnm{Rabolli}, \binits{C.}},
\oauthor{\bsnm{Chen}, \binits{W.}},
\oauthor{\bsnm{Parwani}, \binits{A.}},
\oauthor{\bsnm{Niazi}, \binits{M.K.K.}}:
Unified multi-foundation-model slide representation for pan-cancer recognition
  and text-guided tumor localization.
arXiv preprint arXiv:2604.22846
(2026)
\end{botherref}
\endbibitem

%%% 160
\bibitem[\protect\citeauthoryear{Wang et~al.}{2022}]{ref32_medclip}
\begin{bchapter}
\bauthor{\bsnm{Wang}, \binits{Z.}},
\bauthor{\bsnm{Wu}, \binits{Z.}},
\bauthor{\bsnm{Agarwal}, \binits{D.}},
\bauthor{\bsnm{Sun}, \binits{J.}}:
\bctitle{{M}ed{CLIP}: Contrastive learning from unpaired medical images and
  text}.
In: \beditor{\bsnm{Goldberg}, \binits{Y.}},
\beditor{\bsnm{Kozareva}, \binits{Z.}},
\beditor{\bsnm{Zhang}, \binits{Y.}} (eds.)
\bbtitle{Proceedings of the 2022 Conference on Empirical Methods in Natural
  Language Processing},
pp. \bfpage{3876}--\blpage{3887}.
\bpublisher{Association for Computational Linguistics},
\blocation{Abu Dhabi, United Arab Emirates}
(\byear{2022}).
\doiurl{10.18653/v1/2022.emnlp-main.256} .
\burl{https://aclanthology.org/2022.emnlp-main.256/}
\end{bchapter}
\endbibitem

%%% 161
\bibitem[\protect\citeauthoryear{Wu et~al.}{2026}]{wu2026deep}
\begin{botherref}
\oauthor{\bsnm{Wu}, \binits{R.}},
\oauthor{\bsnm{Wang}, \binits{H.}},
\oauthor{\bsnm{Chen}, \binits{H.-T.}},
\oauthor{\bsnm{Carneiro}, \binits{G.}}:
Deep multimodal learning with missing modality: A survey.
Transactions on Machine Learning Research
(2026).
Survey Certification
\end{botherref}
\endbibitem

%%% 162
\bibitem[\protect\citeauthoryear{Wu et~al.}{2025}]{ref53_medsa}
\begin{barticle}
\bauthor{\bsnm{Wu}, \binits{J.}},
\bauthor{\bsnm{Wang}, \binits{Z.}},
\bauthor{\bsnm{Hong}, \binits{M.}},
\bauthor{\bsnm{Ji}, \binits{W.}},
\bauthor{\bsnm{Fu}, \binits{H.}},
\bauthor{\bsnm{Xu}, \binits{Y.}},
\bauthor{\bsnm{Xu}, \binits{M.}},
\bauthor{\bsnm{Jin}, \binits{Y.}}:
\batitle{Medical sam adapter: Adapting segment anything model for medical image
  segmentation}.
\bjtitle{Medical Image Analysis}
\bvolume{102},
\bfpage{103547}
(\byear{2025})
\doiurl{10.1016/j.media.2025.103547}
\end{barticle}
\endbibitem

%%% 163
\bibitem[\protect\citeauthoryear{Wu et~al.}{2025}]{wu2025sampopath}
\begin{botherref}
\oauthor{\bsnm{Wu}, \binits{Y.}},
\oauthor{\bsnm{Zeng}, \binits{W.}},
\oauthor{\bsnm{Song}, \binits{P.}},
\oauthor{\bsnm{Zhao}, \binits{C.}},
\oauthor{\bsnm{Xie}, \binits{X.}},
\oauthor{\bsnm{Wu}, \binits{G.}},
\oauthor{\bsnm{Yu}, \binits{J.}}:
{SAMPO-Path}: Segmentation intent-aligned preference optimization for pathology
  foundation model segmentation.
arXiv preprint arXiv:2508.02464
(2025)
\end{botherref}
\endbibitem

%%% 164
\bibitem[\protect\citeauthoryear{Wang
  et~al.}{2024}]{wang2024continual_learning_survey}
\begin{barticle}
\bauthor{\bsnm{Wang}, \binits{L.}},
\bauthor{\bsnm{Zhang}, \binits{X.}},
\bauthor{\bsnm{Su}, \binits{H.}},
\bauthor{\bsnm{Zhu}, \binits{J.}}:
\batitle{A comprehensive survey of continual learning: Theory, method and
  application}.
\bjtitle{IEEE transactions on pattern analysis and machine intelligence}
\bvolume{46}(\bissue{8}),
\bfpage{5362}--\blpage{5383}
(\byear{2024})
\end{barticle}
\endbibitem

%%% 165
\bibitem[\protect\citeauthoryear{Xia et~al.}{2026}]{xia2026rfd}
\begin{botherref}
\oauthor{\bsnm{Xia}, \binits{J.}},
\oauthor{\bsnm{Shen}, \binits{Z.}},
\oauthor{\bsnm{Xia}, \binits{W.}},
\oauthor{\bsnm{Pu}, \binits{Y.}},
\oauthor{\bsnm{Zou}, \binits{X.}},
\oauthor{\bsnm{Liu}, \binits{W.}},
\oauthor{\bsnm{Ding}, \binits{D.}},
\oauthor{\bsnm{Wang}, \binits{C.}},
\oauthor{\bsnm{Xue}, \binits{G.}},
\oauthor{\bsnm{Dong}, \binits{S.}}, et al.:
Rfd: A reducing feature discrepancy method for unsupervised cross-modality sam
  adaptation.
Computerized Medical Imaging and Graphics,
102781
(2026)
\end{botherref}
\endbibitem

%%% 166
\bibitem[\protect\citeauthoryear{Xu et~al.}{2024}]{xu2024provgigapath}
\begin{barticle}
\bauthor{\bsnm{Xu}, \binits{H.}},
\bauthor{\bsnm{Usuyama}, \binits{N.}},
\bauthor{\bsnm{Bagga}, \binits{J.}},
\bauthor{\bsnm{Zhang}, \binits{S.}},
\bauthor{\bsnm{Rao}, \binits{R.}},
\bauthor{\bsnm{Naumann}, \binits{T.}},
\bauthor{\bsnm{Wong}, \binits{C.}},
\bauthor{\bsnm{Gero}, \binits{Z.}},
\bauthor{\bsnm{Gonz{\'a}lez}, \binits{J.}},
\bauthor{\bsnm{Gu}, \binits{Y.}},
\bauthor{\bsnm{Xu}, \binits{Y.}},
\bauthor{\bsnm{Wei}, \binits{M.}},
\bauthor{\bsnm{Wang}, \binits{W.}},
\bauthor{\bsnm{Ma}, \binits{S.}},
\bauthor{\bsnm{Wei}, \binits{F.}},
\bauthor{\bsnm{Yang}, \binits{J.}},
\bauthor{\bsnm{Li}, \binits{C.}},
\bauthor{\bsnm{Gao}, \binits{J.}}, \betal:
\batitle{A whole-slide foundation model for digital pathology from real-world
  data}.
\bjtitle{Nature}
\bvolume{630}(\bissue{8015}),
\bfpage{181}--\blpage{188}
(\byear{2024})
\doiurl{10.1038/s41586-024-07441-w}
\end{barticle}
\endbibitem

%%% 167
\bibitem[\protect\citeauthoryear{Xu et~al.}{2026}]{ref68_peft_review}
\begin{botherref}
\oauthor{\bsnm{Xu}, \binits{L.}},
\oauthor{\bsnm{Xie}, \binits{H.}},
\oauthor{\bsnm{Qin}, \binits{S.J.}},
\oauthor{\bsnm{Tao}, \binits{X.}},
\oauthor{\bsnm{Wang}, \binits{F.L.}}:
Parameter-efficient fine-tuning methods for pretrained language models: A
  critical review and assessment.
IEEE Transactions on Pattern Analysis and Machine Intelligence
(2026)
\end{botherref}
\endbibitem

%%% 168
\bibitem[\protect\citeauthoryear{Xu et~al.}{2022}]{xu2022adversarial}
\begin{bchapter}
\bauthor{\bsnm{Xu}, \binits{Y.}},
\bauthor{\bsnm{Xie}, \binits{S.}},
\bauthor{\bsnm{Reynolds}, \binits{M.}},
\bauthor{\bsnm{Ragoza}, \binits{M.}},
\bauthor{\bsnm{Gong}, \binits{M.}},
\bauthor{\bsnm{Batmanghelich}, \binits{K.}}:
\bctitle{Adversarial consistency for single domain generalization in medical
  image segmentation}.
In: \bbtitle{International Conference on Medical Image Computing and
  Computer-Assisted Intervention},
pp. \bfpage{671}--\blpage{681}
(\byear{2022}).
\bcomment{Springer}
\end{bchapter}
\endbibitem

%%% 169
\bibitem[\protect\citeauthoryear{Xie et~al.}{2022}]{ref114_simmim}
\begin{bchapter}
\bauthor{\bsnm{Xie}, \binits{Z.}},
\bauthor{\bsnm{Zhang}, \binits{Z.}},
\bauthor{\bsnm{Cao}, \binits{Y.}},
\bauthor{\bsnm{Lin}, \binits{Y.}},
\bauthor{\bsnm{Bao}, \binits{J.}},
\bauthor{\bsnm{Yao}, \binits{Z.}},
\bauthor{\bsnm{Dai}, \binits{Q.}},
\bauthor{\bsnm{Hu}, \binits{H.}}:
\bctitle{Simmim: A simple framework for masked image modeling}.
In: \bbtitle{Proceedings of the IEEE/CVF Conference on Computer Vision and
  Pattern Recognition (CVPR)},
pp. \bfpage{9653}--\blpage{9663}
(\byear{2022})
\end{bchapter}
\endbibitem

%%% 170
\bibitem[\protect\citeauthoryear{Yagis et~al.}{2021}]{yagis2021data_leakage}
\begin{barticle}
\bauthor{\bsnm{Yagis}, \binits{E.}},
\bauthor{\bsnm{Atnafu}, \binits{S.W.}},
\bauthor{\bsnm{Herrera}, \binits{A.}},
\bauthor{\bsnm{Marzi}, \binits{C.}},
\bauthor{\bsnm{Scheda}, \binits{R.}},
\bauthor{\bsnm{Giannelli}, \binits{M.}},
\bauthor{\bsnm{Tessa}, \binits{C.}},
\bauthor{\bsnm{Citi}, \binits{L.}},
\bauthor{\bsnm{Diciotti}, \binits{S.}}:
\batitle{Effect of data leakage in brain mri classification using 2d
  convolutional neural networks}.
\bjtitle{Scientific reports}
\bvolume{11}(\bissue{1}),
\bfpage{22544}
(\byear{2021})
\end{barticle}
\endbibitem

%%% 171
\bibitem[\protect\citeauthoryear{Yosinski
  et~al.}{2014}]{ref76_how_transferable_dl}
\begin{botherref}
\oauthor{\bsnm{Yosinski}, \binits{J.}},
\oauthor{\bsnm{Clune}, \binits{J.}},
\oauthor{\bsnm{Bengio}, \binits{Y.}},
\oauthor{\bsnm{Lipson}, \binits{H.}}:
How transferable are features in deep neural networks?
Advances in neural information processing systems
\textbf{27}
(2014)
\end{botherref}
\endbibitem

%%% 172
\bibitem[\protect\citeauthoryear{Yu et~al.}{2020}]{yu2020gradient}
\begin{barticle}
\bauthor{\bsnm{Yu}, \binits{T.}},
\bauthor{\bsnm{Kumar}, \binits{S.}},
\bauthor{\bsnm{Gupta}, \binits{A.}},
\bauthor{\bsnm{Levine}, \binits{S.}},
\bauthor{\bsnm{Hausman}, \binits{K.}},
\bauthor{\bsnm{Finn}, \binits{C.}}:
\batitle{Gradient surgery for multi-task learning}.
\bjtitle{Advances in neural information processing systems}
\bvolume{33},
\bfpage{5824}--\blpage{5836}
(\byear{2020})
\end{barticle}
\endbibitem

%%% 173
\bibitem[\protect\citeauthoryear{Yu et~al.}{2024}]{yu2024cufit}
\begin{bchapter}
\bauthor{\bsnm{Yu}, \binits{Y.}},
\bauthor{\bsnm{Ko}, \binits{M.}},
\bauthor{\bsnm{Shin}, \binits{S.}},
\bauthor{\bsnm{Kim}, \binits{K.}},
\bauthor{\bsnm{Lee}, \binits{K.}}:
\bctitle{Curriculum fine-tuning of vision foundation model for medical image
  classification under label noise}.
In: \bbtitle{Advances in Neural Information Processing Systems},
vol. \bseriesno{37}
(\byear{2024})
\end{bchapter}
\endbibitem

%%% 174
\bibitem[\protect\citeauthoryear{Yu et~al.}{2022}]{yu2022external}
\begin{barticle}
\bauthor{\bsnm{Yu}, \binits{A.C.}},
\bauthor{\bsnm{Mohajer}, \binits{B.}},
\bauthor{\bsnm{Eng}, \binits{J.}}:
\batitle{External validation of deep learning algorithms for radiologic
  diagnosis: a systematic review}.
\bjtitle{Radiology: Artificial Intelligence}
\bvolume{4}(\bissue{3}),
\bfpage{210064}
(\byear{2022})
\end{barticle}
\endbibitem

%%% 175
\bibitem[\protect\citeauthoryear{Yan et~al.}{2025}]{yan2025samed2}
\begin{bchapter}
\bauthor{\bsnm{Yan}, \binits{Z.}},
\bauthor{\bsnm{Song}, \binits{S.}},
\bauthor{\bsnm{Song}, \binits{D.}},
\bauthor{\bsnm{Li}, \binits{Y.}},
\bauthor{\bsnm{Zhou}, \binits{R.}},
\bauthor{\bsnm{Sun}, \binits{W.}},
\bauthor{\bsnm{Chen}, \binits{Z.}},
\bauthor{\bsnm{Kim}, \binits{S.}},
\bauthor{\bsnm{Ren}, \binits{H.}},
\bauthor{\bsnm{Liu}, \binits{T.}}, \betal:
\bctitle{Samed-2: Selective memory enhanced medical segment anything model}.
In: \bbtitle{International Conference on Medical Image Computing and
  Computer-Assisted Intervention},
pp. \bfpage{540}--\blpage{550}
(\byear{2025}).
\bcomment{Springer}
\end{bchapter}
\endbibitem

%%% 176
\bibitem[\protect\citeauthoryear{Ye et~al.}{2025}]{ye2025medseqft}
\begin{barticle}
\bauthor{\bsnm{Ye}, \binits{Y.}},
\bauthor{\bsnm{Wu}, \binits{Y.}},
\bauthor{\bsnm{Luo}, \binits{X.}},
\bauthor{\bsnm{Zhang}, \binits{H.}},
\bauthor{\bsnm{Chen}, \binits{Z.}},
\bauthor{\bsnm{Dang}, \binits{T.}},
\bauthor{\bsnm{Zhang}, \binits{Y.}},
\bauthor{\bsnm{Xia}, \binits{Y.}}:
\batitle{{MedSeqFT}: Sequential fine-tuning foundation models for 3d medical
  image segmentation}.
\bjtitle{arXiv preprint arXiv:2509.06096}
(\byear{2025})
\doiurl{10.48550/arXiv.2509.06096}
\end{barticle}
\endbibitem

%%% 177
\bibitem[\protect\citeauthoryear{Yan et~al.}{2018}]{yan2018deeplesion}
\begin{barticle}
\bauthor{\bsnm{Yan}, \binits{K.}},
\bauthor{\bsnm{Wang}, \binits{X.}},
\bauthor{\bsnm{Lu}, \binits{L.}},
\bauthor{\bsnm{Summers}, \binits{R.M.}}:
\batitle{Deeplesion: automated mining of large-scale lesion annotations and
  universal lesion detection with deep learning}.
\bjtitle{Journal of medical imaging}
\bvolume{5}(\bissue{3}),
\bfpage{036501}--\blpage{036501}
(\byear{2018})
\end{barticle}
\endbibitem

%%% 178
\bibitem[\protect\citeauthoryear{Yan et~al.}{2025}]{yan2025multimodal}
\begin{barticle}
\bauthor{\bsnm{Yan}, \binits{S.}},
\bauthor{\bsnm{Yu}, \binits{Z.}},
\bauthor{\bsnm{Primiero}, \binits{C.}},
\bauthor{\bsnm{Vico-Alonso}, \binits{C.}},
\bauthor{\bsnm{Wang}, \binits{Z.}},
\bauthor{\bsnm{Yang}, \binits{L.}},
\bauthor{\bsnm{Tschandl}, \binits{P.}},
\bauthor{\bsnm{Hu}, \binits{M.}},
\bauthor{\bsnm{Ju}, \binits{L.}},
\bauthor{\bsnm{Tan}, \binits{G.}}, \betal:
\batitle{A multimodal vision foundation model for clinical dermatology}.
\bjtitle{Nature Medicine}
\bvolume{31}(\bissue{8}),
\bfpage{2691}--\blpage{2702}
(\byear{2025})
\end{barticle}
\endbibitem

%%% 179
\bibitem[\protect\citeauthoryear{Ye et~al.}{2024}]{ye2024spurious}
\begin{botherref}
\oauthor{\bsnm{Ye}, \binits{W.}},
\oauthor{\bsnm{Zheng}, \binits{G.}},
\oauthor{\bsnm{Cao}, \binits{X.}},
\oauthor{\bsnm{Ma}, \binits{Y.}},
\oauthor{\bsnm{Zhang}, \binits{A.}}:
Spurious correlations in machine learning: A survey.
arXiv preprint arXiv:2402.12715
(2024)
\end{botherref}
\endbibitem

%%% 180
\bibitem[\protect\citeauthoryear{Zech et~al.}{2018}]{zech2018variable}
\begin{barticle}
\bauthor{\bsnm{Zech}, \binits{J.R.}},
\bauthor{\bsnm{Badgeley}, \binits{M.A.}},
\bauthor{\bsnm{Liu}, \binits{M.}},
\bauthor{\bsnm{Costa}, \binits{A.B.}},
\bauthor{\bsnm{Titano}, \binits{J.J.}},
\bauthor{\bsnm{Oermann}, \binits{E.K.}}:
\batitle{Variable generalization performance of a deep learning model to detect
  pneumonia in chest radiographs: a cross-sectional study}.
\bjtitle{PLoS medicine}
\bvolume{15}(\bissue{11}),
\bfpage{1002683}
(\byear{2018})
\end{barticle}
\endbibitem

%%% 181
\bibitem[\protect\citeauthoryear{Zhou et~al.}{2023}]{zhou2023foundation}
\begin{barticle}
\bauthor{\bsnm{Zhou}, \binits{Y.}},
\bauthor{\bsnm{Chia}, \binits{M.A.}},
\bauthor{\bsnm{Wagner}, \binits{S.K.}},
\bauthor{\bsnm{Ayhan}, \binits{M.S.}},
\bauthor{\bsnm{Williamson}, \binits{D.J.}},
\bauthor{\bsnm{Struyven}, \binits{R.R.}},
\bauthor{\bsnm{Liu}, \binits{T.}},
\bauthor{\bsnm{Xu}, \binits{M.}},
\bauthor{\bsnm{Lozano}, \binits{M.G.}},
\bauthor{\bsnm{Woodward-Court}, \binits{P.}}, \betal:
\batitle{A foundation model for generalizable disease detection from retinal
  images}.
\bjtitle{Nature}
\bvolume{622}(\bissue{7981}),
\bfpage{156}--\blpage{163}
(\byear{2023})
\end{barticle}
\endbibitem

%%% 182
\bibitem[\protect\citeauthoryear{Zhao et~al.}{2019}]{zhao2019learning}
\begin{bchapter}
\bauthor{\bsnm{Zhao}, \binits{H.}},
\bauthor{\bsnm{Des~Combes}, \binits{R.T.}},
\bauthor{\bsnm{Zhang}, \binits{K.}},
\bauthor{\bsnm{Gordon}, \binits{G.}}:
\bctitle{On learning invariant representations for domain adaptation}.
In: \bbtitle{International Conference on Machine Learning},
pp. \bfpage{7523}--\blpage{7532}
(\byear{2019}).
\bcomment{PMLR}
\end{bchapter}
\endbibitem

%%% 183
\bibitem[\protect\citeauthoryear{Zaken et~al.}{2022}]{ref97_bitfit}
\begin{bchapter}
\bauthor{\bsnm{Zaken}, \binits{E.B.}},
\bauthor{\bsnm{Goldberg}, \binits{Y.}},
\bauthor{\bsnm{Ravfogel}, \binits{S.}}:
\bctitle{Bitfit: Simple parameter-efficient fine-tuning for transformer-based
  masked language-models}.
In: \bbtitle{Proceedings of the 60th Annual Meeting of the Association for
  Computational Linguistics (Volume 2: Short Papers)},
pp. \bfpage{1}--\blpage{9}
(\byear{2022})
\end{bchapter}
\endbibitem

%%% 184
\bibitem[\protect\citeauthoryear{Zhang et~al.}{2026}]{zhang2026data}
\begin{barticle}
\bauthor{\bsnm{Zhang}, \binits{Y.}},
\bauthor{\bsnm{Gao}, \binits{J.}},
\bauthor{\bsnm{Tan}, \binits{Z.}},
\bauthor{\bsnm{Zhou}, \binits{L.}},
\bauthor{\bsnm{Ding}, \binits{K.}},
\bauthor{\bsnm{Zhou}, \binits{M.}},
\bauthor{\bsnm{Zhang}, \binits{S.}},
\bauthor{\bsnm{Wang}, \binits{D.}}:
\batitle{Data-centric foundation models in computational healthcare: A survey}.
\bjtitle{ACM Computing Surveys}
\bvolume{58}(\bissue{11}),
\bfpage{1}--\blpage{35}
(\byear{2026})
\end{barticle}
\endbibitem

%%% 185
\bibitem[\protect\citeauthoryear{Zhao et~al.}{2025}]{ref123_BiomedParse}
\begin{barticle}
\bauthor{\bsnm{Zhao}, \binits{T.}},
\bauthor{\bsnm{Gu}, \binits{Y.}},
\bauthor{\bsnm{Yang}, \binits{J.}},
\bauthor{\bsnm{Usuyama}, \binits{N.}},
\bauthor{\bsnm{Lee}, \binits{H.H.}},
\bauthor{\bsnm{Kiblawi}, \binits{S.}},
\bauthor{\bsnm{Naumann}, \binits{T.}},
\bauthor{\bsnm{Gao}, \binits{J.}},
\bauthor{\bsnm{Crabtree}, \binits{A.}},
\bauthor{\bsnm{Abel}, \binits{J.}}, \betal:
\batitle{A foundation model for joint segmentation, detection and recognition
  of biomedical objects across nine modalities}.
\bjtitle{Nature methods}
\bvolume{22}(\bissue{1}),
\bfpage{166}--\blpage{176}
(\byear{2025})
\end{barticle}
\endbibitem

%%% 186
\bibitem[\protect\citeauthoryear{Zhang}{2026}]{zhang2026hospitalbias}
\begin{bchapter}
\bauthor{\bsnm{Zhang}, \binits{M.}}:
\bctitle{Hospital-specific bias in patch-based pathology models}.
In: \bbtitle{2026 IEEE International Symposium on Biomedical Imaging (ISBI)},
pp. \bfpage{1}--\blpage{4}
(\byear{2026})
\end{bchapter}
\endbibitem

%%% 187
\bibitem[\protect\citeauthoryear{Zhang et~al.}{2022}]{ref_112_convirt}
\begin{bchapter}
\bauthor{\bsnm{Zhang}, \binits{Y.}},
\bauthor{\bsnm{Jiang}, \binits{H.}},
\bauthor{\bsnm{Miura}, \binits{Y.}},
\bauthor{\bsnm{Manning}, \binits{C.D.}},
\bauthor{\bsnm{Langlotz}, \binits{C.P.}}:
\bctitle{Contrastive learning of medical visual representations from paired
  images and text}.
In: \bbtitle{Machine Learning for Healthcare Conference},
pp. \bfpage{2}--\blpage{25}
(\byear{2022}).
\bcomment{PMLR}
\end{bchapter}
\endbibitem

%%% 188
\bibitem[\protect\citeauthoryear{Zhang et~al.}{2023}]{zhang2023deep}
\begin{barticle}
\bauthor{\bsnm{Zhang}, \binits{Y.}},
\bauthor{\bsnm{Kang}, \binits{B.}},
\bauthor{\bsnm{Hooi}, \binits{B.}},
\bauthor{\bsnm{Yan}, \binits{S.}},
\bauthor{\bsnm{Feng}, \binits{J.}}:
\batitle{Deep long-tailed learning: A survey}.
\bjtitle{IEEE transactions on pattern analysis and machine intelligence}
\bvolume{45}(\bissue{9}),
\bfpage{10795}--\blpage{10816}
(\byear{2023})
\end{barticle}
\endbibitem

%%% 189
\bibitem[\protect\citeauthoryear{Zhang and Liu}{2023}]{ref54_SAMed}
\begin{botherref}
\oauthor{\bsnm{Zhang}, \binits{K.}},
\oauthor{\bsnm{Liu}, \binits{D.}}:
Customized segment anything model for medical image segmentation.
arXiv preprint arXiv:2304.13785
(2023)
\end{botherref}
\endbibitem

%%% 190
\bibitem[\protect\citeauthoryear{Zhou et~al.}{2023}]{ref26_ssl_mae_mia}
\begin{bchapter}
\bauthor{\bsnm{Zhou}, \binits{L.}},
\bauthor{\bsnm{Liu}, \binits{H.}},
\bauthor{\bsnm{Bae}, \binits{J.}},
\bauthor{\bsnm{He}, \binits{J.}},
\bauthor{\bsnm{Samaras}, \binits{D.}},
\bauthor{\bsnm{Prasanna}, \binits{P.}}:
\bctitle{Self pre-training with masked autoencoders for medical image
  classification and segmentation}.
In: \bbtitle{2023 IEEE 20th International Symposium on Biomedical Imaging
  (ISBI)},
pp. \bfpage{1}--\blpage{6}
(\byear{2023}).
\bcomment{IEEE}
\end{bchapter}
\endbibitem

%%% 191
\bibitem[\protect\citeauthoryear{Zhang and
  Metaxas}{2024}]{ref12_challenges_per_fm}
\begin{barticle}
\bauthor{\bsnm{Zhang}, \binits{S.}},
\bauthor{\bsnm{Metaxas}, \binits{D.}}:
\batitle{On the challenges and perspectives of foundation models for medical
  image analysis}.
\bjtitle{Medical Image Analysis}
\bvolume{91},
\bfpage{102996}
(\byear{2024})
\doiurl{10.1016/j.media.2023.102996}
\end{barticle}
\endbibitem

%%% 192
\bibitem[\protect\citeauthoryear{Zhou et~al.}{2022}]{zhou2022generalizable}
\begin{bchapter}
\bauthor{\bsnm{Zhou}, \binits{Z.}},
\bauthor{\bsnm{Qi}, \binits{L.}},
\bauthor{\bsnm{Shi}, \binits{Y.}}:
\bctitle{Generalizable medical image segmentation via random amplitude mixup
  and domain-specific image restoration}.
In: \bbtitle{European Conference on Computer Vision},
pp. \bfpage{420}--\blpage{436}
(\byear{2022}).
\bcomment{Springer}
\end{bchapter}
\endbibitem

%%% 193
\bibitem[\protect\citeauthoryear{Zhu et~al.}{2024}]{ref136_melo}
\begin{bchapter}
\bauthor{\bsnm{Zhu}, \binits{Y.}},
\bauthor{\bsnm{Shen}, \binits{Z.}},
\bauthor{\bsnm{Zhao}, \binits{Z.}},
\bauthor{\bsnm{Wang}, \binits{S.}},
\bauthor{\bsnm{Wang}, \binits{X.}},
\bauthor{\bsnm{Zhao}, \binits{X.}},
\bauthor{\bsnm{Shen}, \binits{D.}},
\bauthor{\bsnm{Wang}, \binits{Q.}}:
\bctitle{Melo: Low-rank adaptation is better than fine-tuning for medical image
  diagnosis}.
In: \bbtitle{2024 IEEE International Symposium on Biomedical Imaging (ISBI)},
pp. \bfpage{1}--\blpage{5}
(\byear{2024}).
\doiurl{10.1109/ISBI56570.2024.10635615}
\end{bchapter}
\endbibitem

%%% 194
\bibitem[\protect\citeauthoryear{Zhang et~al.}{2024}]{ref146_mediclip}
\begin{bchapter}
\bauthor{\bsnm{Zhang}, \binits{X.}},
\bauthor{\bsnm{Xu}, \binits{M.}},
\bauthor{\bsnm{Qiu}, \binits{D.}},
\bauthor{\bsnm{Yan}, \binits{R.}},
\bauthor{\bsnm{Lang}, \binits{N.}},
\bauthor{\bsnm{Zhou}, \binits{X.}}:
\bctitle{Mediclip: Adapting clip for few-shot medical image anomaly detection}.
In: \bbtitle{International Conference on Medical Image Computing and
  Computer-Assisted Intervention},
pp. \bfpage{458}--\blpage{468}
(\byear{2024}).
\bcomment{Springer}
\end{bchapter}
\endbibitem

%%% 195
\bibitem[\protect\citeauthoryear{Zhou et~al.}{2026}]{zhou2026stamp}
\begin{botherref}
\oauthor{\bsnm{Zhou}, \binits{F.}},
\oauthor{\bsnm{Xu}, \binits{Y.}},
\oauthor{\bsnm{Zhang}, \binits{Z.}},
\oauthor{\bsnm{Wang}, \binits{Y.}},
\oauthor{\bsnm{Guo}, \binits{Z.}},
\oauthor{\bsnm{Liang}, \binits{L.}},
\oauthor{\bsnm{Ma}, \binits{J.}},
\oauthor{\bsnm{Jin}, \binits{C.}},
\oauthor{\bsnm{Liu}, \binits{Z.}},
\oauthor{\bsnm{Zhou}, \binits{H.}},
\oauthor{\bsnm{Wang}, \binits{H.}},
\oauthor{\bsnm{Cai}, \binits{D.}},
\oauthor{\bsnm{Zhao}, \binits{C.}},
\oauthor{\bsnm{Wang}, \binits{X.}},
\oauthor{\bsnm{Yang}, \binits{C.}},
\oauthor{\bsnm{Wang}, \binits{Y.}},
\oauthor{\bsnm{Li}, \binits{W.}},
\oauthor{\bsnm{Gao}, \binits{F.}},
\oauthor{\bsnm{Wang}, \binits{Z.}},
\oauthor{\bsnm{Li}, \binits{Z.}},
\oauthor{\bsnm{Zhang}, \binits{X.}},
\oauthor{\bsnm{Liang}, \binits{L.}},
\oauthor{\bsnm{Chen}, \binits{H.}}:
Spatial transcriptomics-guided alignment enhances molecular profiling in
  pathology foundation model.
arXiv preprint arXiv:2606.03644
(2026)
\end{botherref}
\endbibitem

%%% 196
\bibitem[\protect\citeauthoryear{Zhang
  et~al.}{2026}]{zhang2026smolvlmsynthetic}
\begin{barticle}
\bauthor{\bsnm{Zhang}, \binits{S.}},
\bauthor{\bsnm{Yilmaz}, \binits{A.}},
\bauthor{\bsnm{Gencoglan}, \binits{G.}},
\bauthor{\bsnm{Temelkuran}, \binits{B.}}:
\batitle{Fine-tuning a small vision language model using synthetic data for
  explaining bacterial skin disease images}.
\bjtitle{Diagnostics}
\bvolume{16}(\bissue{4}),
\bfpage{603}
(\byear{2026})
\doiurl{10.3390/diagnostics16040603}
\end{barticle}
\endbibitem

%%% 197
\bibitem[\protect\citeauthoryear{Zhang et~al.}{2025}]{ref86_brain_adapter}
\begin{bchapter}
\bauthor{\bsnm{Zhang}, \binits{J.}},
\bauthor{\bsnm{Yu}, \binits{X.}},
\bauthor{\bsnm{Lyu}, \binits{Y.}},
\bauthor{\bsnm{Zhang}, \binits{L.}},
\bauthor{\bsnm{Chen}, \binits{T.}},
\bauthor{\bsnm{Cao}, \binits{C.}},
\bauthor{\bsnm{Zhuang}, \binits{Y.}},
\bauthor{\bsnm{Chen}, \binits{M.}},
\bauthor{\bsnm{Liu}, \binits{T.}},
\bauthor{\bsnm{Zhu}, \binits{D.}}:
\bctitle{Brain-adapter: enhancing neurological disorder analysis with
  adapter-tuning multimodal large language models}.
In: \bbtitle{2025 IEEE 22nd International Symposium on Biomedical Imaging
  (ISBI)},
pp. \bfpage{1}--\blpage{5}
(\byear{2025}).
\bcomment{IEEE}
\end{bchapter}
\endbibitem

\end{thebibliography}

\end{document}